\documentclass{isprs} 
\usepackage{subcaption}
\usepackage{setspace}
\usepackage{geometry} 
\usepackage{epstopdf}
\usepackage[labelsep=period]{caption}  
\usepackage[british]{babel} 
\usepackage[hang]{footmisc}
\usepackage{hyperref}
\usepackage{pgfplotstable,booktabs}
\usepackage[utf8]{inputenc} % allow utf-8 input
\usepackage[T1]{fontenc}    % use 8-bit T1 fonts
\usepackage{hyperref}       % hyperlinks
\usepackage{url}            % simple URL typesetting
\usepackage{booktabs}       % professional-quality tables
\usepackage{amsfonts}       % blackboard math symbols
\usepackage{nicefrac}       % compact symbols for 1/2, etc.
\usepackage{microtype}      % microtypography
\usepackage{lipsum}
\usepackage{graphicx}
\usepackage[normalem]{ulem}
\useunder{\uline}{\ul}{}
\usepackage{longtable}
\usepackage{colortbl}
\usepackage{hyperref}
\usepackage{tikz}
\usepackage{tabu} % if you want
\usepackage{mathtools}
\usepackage{placeins}
\usepackage{isotope}
\usepackage{makecell}
\usepackage{graphicx}
\usepackage{lscape}
\usepackage{blindtext}
\usepackage{textgreek}
\usepackage{svg}
\usepackage{indentfirst}
\usepackage{amssymb}
\usepackage{multirow}
\usepackage{float}
\begin{document}

\title{A Benchmark Study of Neural Network Compression Methods for Hyperspectral Image Classification}
% Leave empty (\version{}) when submitting the full paper
\version{}

\author{Sai Shi}

\address{Department of Computer and Information Science, Temple University, Philadelphia, USA
}

\icwg{}   %This field is optional.

\abstract{
Deep neural networks have achieved strong performance in image classification tasks due to their ability to learn complex patterns from high-dimensional data. However, their large computational and memory requirements often limit deployment on resource-constrained platforms such as remote sensing devices and edge systems. Network compression techniques have therefore been proposed to reduce model size and computational cost while maintaining predictive performance. In this study, we conduct a systematic evaluation of neural network compression methods for a remote sensing application, namely hyperspectral land cover classification. Specifically, we examine three widely used compression strategies for convolutional neural networks: pruning, quantization, and knowledge distillation. Experiments are conducted on two benchmark hyperspectral datasets, considering classification accuracy, memory consumption, and inference efficiency. Our results demonstrate that compressed models can significantly reduce model size and computational cost while maintaining competitive classification performance. These findings provide insights into the trade-offs between compression ratio, efficiency, and accuracy, and highlight the potential of compression techniques for enabling efficient deep learning deployment in remote sensing applications.}

% The source codes of the methods in this paper are available from: \url{https://github.com/sai-shi/hyperspectral_network_compression}. 
% }

% \keywords{Network Compression, Hyperspectral Image Classification, Pruning, Quantization, Knowledge Distillation}

\maketitle

\section{Introduction}\label{INTRODUCTION}

Remote sensing refers to the acquisition of information about objects or phenomena without direct physical contact, through sensors that measure reflected or emitted electromagnetic energy, typically deployed on satellite or airborne platforms \cite{toth2016remote}. Its fundamental objective is to infer physical, chemical, or biological properties of targets that are inaccessible, large-scale, or impractical to observe through in situ measurements. Remote sensing has been widely applied across diverse domains \cite{navalgund2007remote}, including agriculture \cite{shanmugapriya2019applications}, ecology \cite{pettorelli2014satellite}, geology \cite{ramakrishnan2015hyperspectral}, oceanography \cite{devi2015applications}, and meteorology \cite{thies2011satellite}. Representative applications encompass land-cover classification, land-use mapping, weather forecasting, natural hazard monitoring, and natural resource exploration.

Land cover is the physical material at the surface of Earth, including grass, trees, water, etc. Land cover classification, or land cover mapping, aims to detect the pixel-level land-cover category given spectral and spatial information \cite{gomez2016optical,foody2002status}. Various types of remote sensing images have been used for land cover classification, which can be characterised by their spectral, spatial, radiometric, and temporal resolutions. In particular, Hyperspectral Imaging (HSI) \cite{khan2018modern,lu2014medical} is a method of capturing and processing information from a broad range of electromagnetic wavelengths, typically outside the visible spectrum. From the same region on Earth's surface, hundreds of narrow spectral bands are provided by hyperspectral imaging sensors \cite{Vali2020}. Each pixel in a hyperspectral image (HSI) can be thought of as a high-dimensional vector whose entries correspond to the wavelength's spectral reflectance. Hyperspectral imaging can exploit the spatial connections among the different spectra in an area, permitting more intricate phantom spatial models for more exact division and order of the picture \cite{HeLin2018}. It serves as a powerful tool for remote sensing applications aimed at mapping and monitoring land cover and land use patterns at various spatial and temporal scales.

There are numerous large-scale hyperspectral imaging datasets that are publicly available. For example, HySpecNet-11k \cite{fuchs2023} is a large-scale hyperspectral benchmark dataset made up of 11,483 nonoverlapping image patches acquired by the EnMAP satellite. Each patch is a portion of 128 × 128 pixels with 224 spectral bands and with a ground sample distance of 30 m. Another large-scale hyperspectral imaging dataset is WHU-OHS \cite{LI2022103022}, which consists of about 90 million manually labeled samples of 7795 Orbita hyperspectral satellite (OHS) image patches (sized 512 × 512) from 40 Chinese locations. This dataset ranges from the visible to near-infrared range, with an average spectral resolution of 15 nm. Additional work \cite{Chakrabarti2011} created a database by collecting fifty hyperspectral images of indoor and outdoor scenes, featuring a diversity of objects, materials and scale. 

Additionally, a number of small-scale datasets for hyperspectral imaging have been made accessible to researchers, such as Indian Pines, University of Pavia, Salinas, Houston, Botswana, and Kennedy Space Center (KSC). Typically, each dataset consists of a single static image, with large number of spectrals. Though each dataset may be collected around a specific region using specific remote sensing device, they share similar properties with each other. The datasets can be downloaded using this link \footnote{\url{https://www.ehu.eus/ccwintco/index.php/Hyperspectral_Remote_Sensing_Scenes}}. 

% Table \ref{tab:datasets} displays the description of each dataset.
% \begin{table*}[!h]
% \centering
% \caption{Main public labeled datasets in hyperspectral imaging (Labels: number of non-zero class pixels)}
% \label{tab:datasets}
%     \resizebox{0.7\textwidth}{!}{
%     \large
%     \begin{tabular}{*{8}{c}}
%         \toprule
%         \hline
%        Dataset & Pixels &  Bands & Labels & Range($\mu$m) & GSD(m) & Classes & Sensor\\
%         \midrule
%         \hline
%         India Pines & 145x145 & 224 & 10,249 & 0.4-2.5 & 20 & 16 & AVIRIS\\
%         Pavia Centre & 1096x1096 & 102 & 7,456 & 0.43-0.85 & 1.3 & 9 & ROSIS\\
%         Pavia University & 610x610 & 103 & 42,776 & 0.43-0.85 & 1.3 & 9 & ROSIS\\
%         Salinas & 512x217 & 224 & 54,129 & 0.4-2.5 & 3.7 & 16 & AVIRIS\\
%         Houston & 349x1905 & 144 & 15,029 & 0.38-1.05 & 2.5 & 15 & LiDAR\\
%         KSC & 314,368 & 176 & 5,211 & 0.4-2.5 & 18 & 13 & AVIRIS\\
%         Botswana & 377,856 & 145 & 3,248 & 0.4-2.5 & 30 & 14 & EO-1\\
%         DFC 2018 & 2383x601 & 48 & 547,807 & 0.38-1.05 & 1 & 20 & LiDAR\\
        
%         \hline
%         \bottomrule
%     \end{tabular}
%     }
%     \vspace{-1em}
% \end{table*}

Hyperspectral image classification \cite{haidarh2025exploring,ahmad2025comprehensive} involves analyzing images captured across many narrow spectral bands (often hundreds) in pixel-level to identify and categorize materials or objects on the Earth's surface. This task leverages the unique spectral signatures of different materials, allowing for detailed analysis and differentiation that is not possible with standard RGB images or multispectral images. To better formulate the problem, this process can be defined as $f: X \rightarrow Y$, with input space $X \subseteq \mathbb{N} ^ {W \times H \times K}$ are respectively the width, height, and number of spectral bands for each input image, where the output space for pixel-level hyperspectral image classification is represented as $Y \subseteq \mathbb{C}$, where $\mathbb{C}=\{\Omega_0, \Omega_1, ..., \Omega_k\} $ is the set of possible land cover categories.

Several techniques have been applied to hyperspectral image classification, including traditional statistical algorithms and machine learning approaches, such as random forest and support vector machines \cite{ullah2024conventional}. Recently, deep learning has been increasingly popular in remote sensing tasks such as land cover classification, image fusion, and scene classification \cite{Zhu2017,paoletti2019deep,MA2019166}. Compared with manually designed feature-based methods, deep learning architecture can learn more abstract and discriminative semantic features. Specifically, Convolutional Neural Networks (CNNs) have been a popular choice for hyperspectral image classification \cite{bhosle2019} due to the similarities between remote sensing and computer vision, where CNNs have found the most success \cite{Voulodimos2018,KATTENBORN202124,Liuxinni2019}. While the advantage of deep learning is in achieving state-of-the-art accuracy on many tasks, the main weakness of deep learning is that it results in large neural network models that are computationally expensive and slow. Instead of making decisions at the workshop, most remote sensing devices prefer to gather real-time data and analyze them on an aircraft or satellite. In this context, due to the hardware constraint of CPUs and GPUs embedded in remote sensing devices, there is limited computing power that can be applied directly to deep learning models on these machines, which makes it challenging to make use of them \cite{Zhangshuo2020}. 

In response, network compression techniques \cite{choudhary2020comprehensive,deng2020model} enable the deployment of efficient models that can run effectively on resource-constrained devices without significantly compromising predictive performance. The objective of neural network compression is to transform a pretrained neural network into a comparably accurate but more compact model that requires less memory, fewer floating-point operations, and reduced energy consumption, thereby improving inference speed and lowering hardware requirements. This is typically achieved through techniques such as parameter pruning \cite{cheng2024survey}, weight quantization \cite{rokh2023comprehensive}, knowledge distillation \cite{gou2021knowledge}, low-rank factorization \cite{kishore2017literature}, and architecture redesign. By reducing model complexity while preserving accuracy, compression methods facilitate real-time processing, edge deployment, and large-scale inference, making them particularly well suited for remote sensing applications where computational resources, storage capacity, and power availability may be limited.

We start by discussing the data sets used for benchmarking various neural network compression algorithms in Section 2. In Section 3, we review several major paradigms for neural network compression. Section 4 shows the benchmark experimental results, and we discussed the results and limitations of this work in section 5.

\section{Datasets}\label{HSI}

In this section, we will discuss about the datasets being used for this work, including the characteristics of the raw data and the approach of preprocessing hyperspectral images. Since the main properties for hyperspectral datasets are similar, despite some minor difference in the image scales, we focus on two most widely-used datasets: Indian Pines \cite{PURR1947} and University of Pavia \cite{huang2009comparative}.  

\subsection{Raw Data}

In this work, we focus on evaluating hyperspectral image classification on these two small-scale datasets due to three primary reasons: a) we focus on resource-constrained environments (e.g., devices with limited computational power or bandwidth, such as drones or edge devices), where having access to large datasets might be infeasible; b) compression techniques should be evaluated based on the model's efficiency gains rather than just on large-scale data performance. If the compressed model works well on a small dataset, the approach is likely valid for larger datasets too; c) research on small datasets is an efficient way to iterate on new ideas and test model behavior before scaling up to larger datasets, saving time and resources.

\subsubsection{Indian Pines}

Indian Pine is a dataset captured using the AVIRIS sensor \cite{GREEN1998227}. With a ground sampling distance (GSD) of 20 m/px, the scene covers agricultural regions in North-Western Indiana, USA, producing an image with 224 spectral bands that is 145x145 pixels. The span of wavelengths covered by the sensor is $0.4-2.5 \mu m$. The majority of the image's pixels depict fields with a variety of crops, while the remaining ones show forests and dense vegetation. There are sixteen labeled groups (corn, grass, soybean, woods, etc.), some of which are quite uncommon (less than 100 samples for oats or alfalfa). Water absorption bands (104-108, 150-163, and 220) are often removed before processing. Despite its small size, it is one of the community's main reference datasets. When evaluating classification algorithms, rare classes are typically ignored. 

\subsubsection{University of Pavia}

The ROSIS sensor \cite{kunkel1988rosis} collected the scene over the University of Pavia campus in northern Italy, producing an image of size 610x610 pixels. Some of the samples in the image contain no information and have to be discarded before the analysis, which yields an image of size 610x340. Nine classes that are part of an urban setting are present in the dataset, with a total of 42,776 labeled samples, including the asphalt, meadows, gravel, trees, metal sheet, bare soil, bitumen, brick, and shadow. The geometric resolution is 1.3 meters. After removing the noisy bands, the scene under consideration has 103 spectral bands that span the spectral range from 0.43 to 0.86 m.

\subsection{Data Preprocessing}

The data preprocessing of hyperspectral image consists of three main steps: cleaning, transformation, and splitting. The purpose is to ensure the data format is compatible with CNN inputs while maintaining the spatial and temporal structures in the data.

\begin{figure*}[!h]
\centering
\begin{subfigure}{\linewidth}
    \centering
    \includegraphics[width=0.75\textwidth]{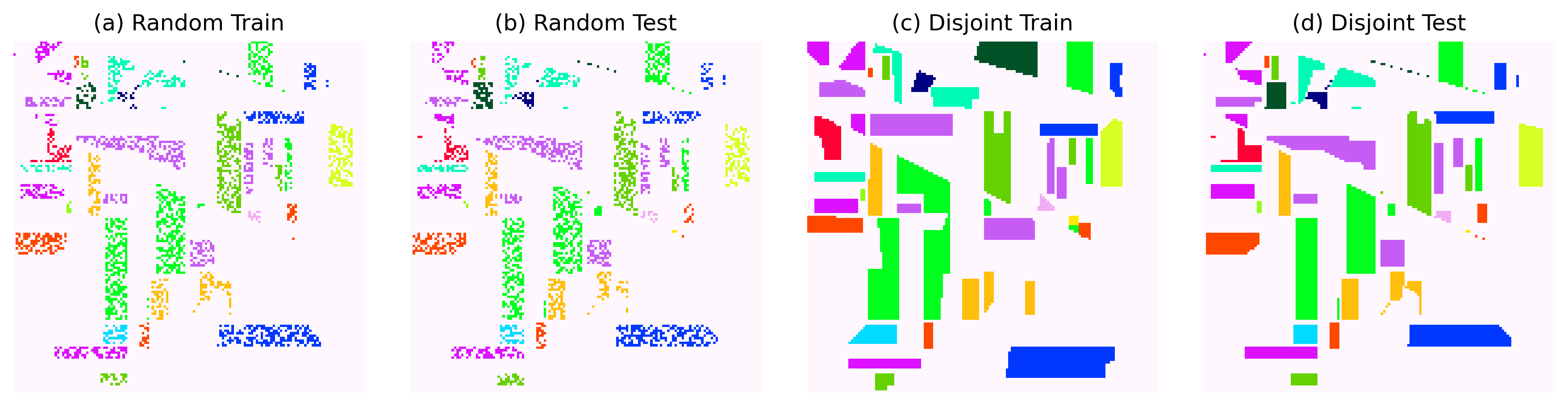}
    \caption{IP dataset}\label{fig:ip_split}
\end{subfigure}
\bigskip
\begin{subfigure}{\linewidth}
  \centering
  \includegraphics[width=0.75\textwidth]{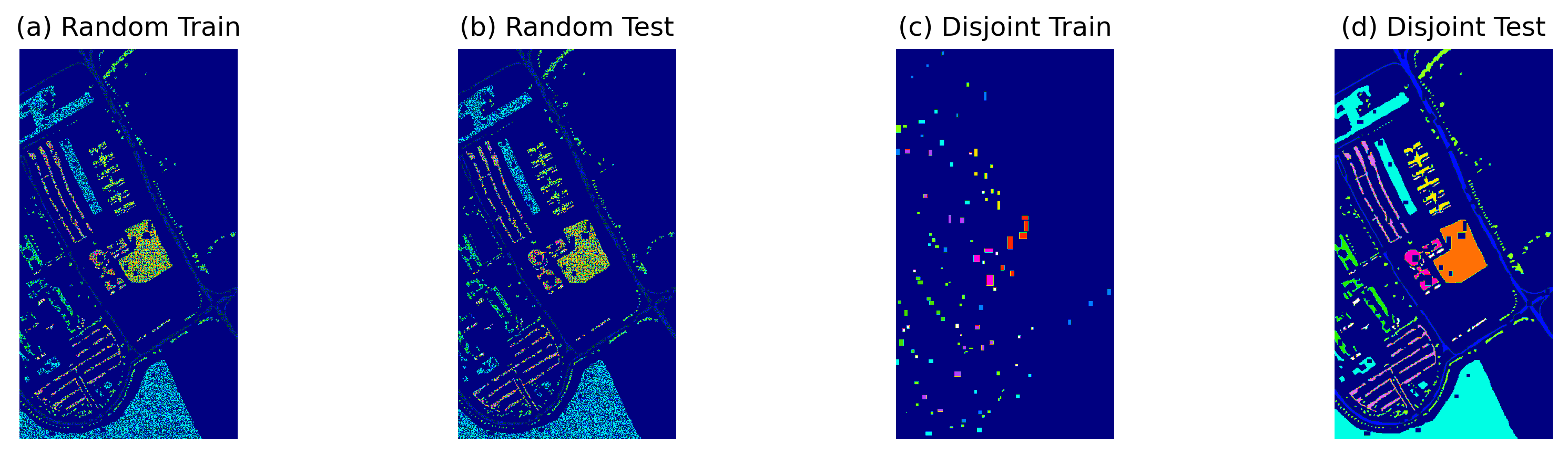}
  \caption{UP dataset}\label{fig:up_split}
\end{subfigure} 
\caption{Random selection vs. spatially disjoint samples on the IP and UP dataset, considering the same number of samples per class}
\label{fig:data_split}
\end{figure*}

\paragraph{\textbf{Data cleaning:}} Most of the raw data of hyperspectral imaging usually has 'unclassified' or 'undefined' pixels, whose labels are marked as zeros. The reasons for undefined class could be: 1)the data is missing due to environmental constraints, such as cloud shadow, darkness, etc.; 2)there are insufficient information about which class the pixel belongs to; 3)the pixel class cannot be recognized by the the deployed sensors. To clean the raw data, we filtered out these 'unclassified' pixels before training the model, which followed the standard practice in recent work \cite{Lishutao2019,yu2017convolutional}. We removed the bands covering the region of water absorption. As a result, the number of bands in the Indian Pines dataset is reduced from 224 to 200 after band selection.

\paragraph{\textbf{Data transformation:}} Normalization techniques have been applied to the benchmark dataset in this work to transform features into a similar scale. Specifically, we utilized standard normalization on the features by removing the mean and scaling to unit variance. In remote sensing datasets, the number of channels/features is often large, which may hinder the machine learning model performance due to irrelevant information or overfitting. In response, the dimentionality reduction technique is often applied in hyperspectral images to reduce the number of features without sacrificing too much information. Principle Component Analysis (PCA) and auto-encoders are two typical dimentionality reduction methods in remote sensing applications \cite{rodarmel2002principal,tao2015unsupervised}. In this work, we applied PCA to the raw images to reduce the number of channels, since it is easier to implement, computationally efficient, and more interpretable than neural network based techniques. 

\paragraph{\textbf{Data splitting:}} Many existing remote sensing works divide the datasets in a train and test splits by randomly sampling over the whole image. However, in recent work, people have realized that randomly sampling the training samples over the whole image is not a realistic use case. It is a poor indicator of generalization power since the method ignores that neighboring pixels will be highly correlated, which indicates the test set will be very close to the train set. In the case of CNN, the receptive field of the network might include test samples in the training set. Hence, the IEEE Geoscience and Remote Sensing Society (GRSS) developed a more realistic remote sensing benchmark: Data and Algorithm Standard Evaluation (DASE) \footnote{\url{http://dase.grss-ieee.org}}. This benchmark designed well-defined train/test splits to better evaluate model performance where the samples are extracted from significantly disjoint parts of the image. This allows the evaluation to measure how the model generalizes to new geo-entities. In our experiments, we performed both random data splitting and disjoint splitting according to the DASE benchmark. The Indian Pines image and University of Pavia image after two types of splitting are shown in Fig.\ref{fig:data_split}. The train/test split ratios for Indian Pines and University of Pavia are 55\%/45\% and 7\%/93\%, respectively, which is the same as the disjoint split ratio for the purpose of fair comparison between the two different split.

\section{Approaches}
\subsection{Convolutional Neural Network}\label{sec:CNN}

Convolutional Neural Networks (CNN) have been initially designed for image classification tasks \cite{alzubaidi2021review,li2021survey,gu2018recent} but have later found many applications in object detection \cite{dhillon2020convolutional}, image segmentation \cite{sultana2020evolution}, human action recognition \cite{ji20123d}, and natural language processing \cite{wang2018application}. The architecture of CNNs is designed to take advantage of the 2D structure of an input, which can be an image or any 2d-pattern signal. This is achieved with local connections and tied weights followed by some form of pooling in order to capture invariant features. CNNs also have the advantage of being simpler to train and requiring fewer parameters than feedforward neural networks with the same number of hidden units. In this work, we will focus on the application of CNNs in hyperspectral image classification.

\begin{figure*}[!h]
\centering
\includegraphics[width=0.9\textwidth]{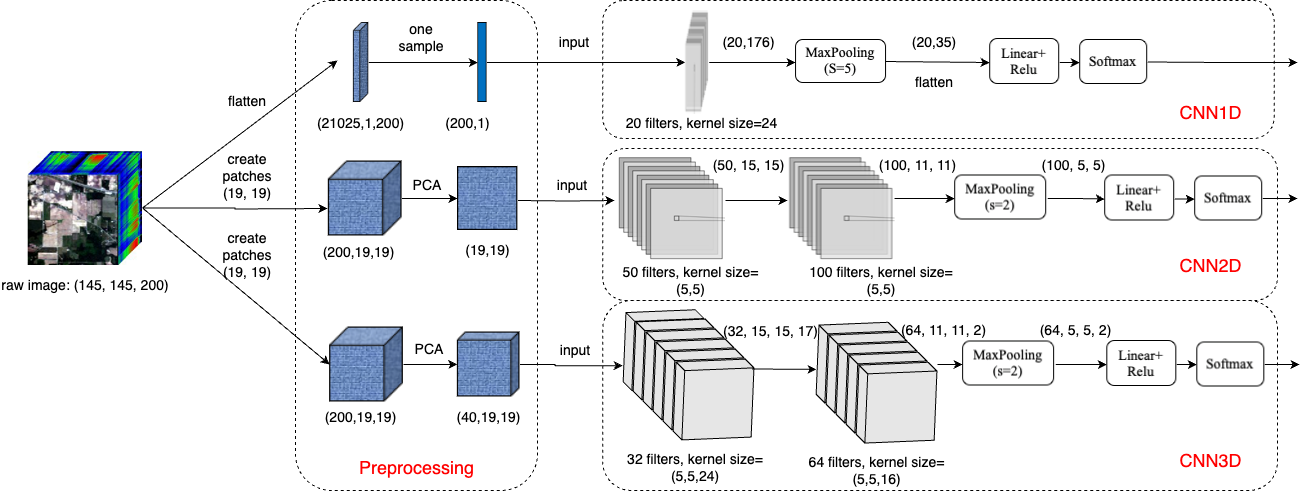}
\caption{\label{fig:dimension} The implementation details of spectral model (CNN1D) , spatial model (CNN2D), and spatial-spectral model (CNN3D)}
\end{figure*}

For the baseline evaluation and comparison, we followed the same CNN structure used in this work \cite{paoletti2019deep}, where a dense CNN model has been designed and evaluated on both Indian Pines and University of Pavia dataset. Specifically, there are two Convolutional (Conv) layers and two Fully-Connected (FC) layers. The filter sizes of two Conv layers are 50 and 100, respectively, with a kernel size of 5x5. Rectified Linear Unit (ReLU) has been used as the activation function and batch normalization has been applied after each Conv layer. A max pooling layer with pooling size 2x2 is attached to the second convolutional layer to create downsampled feature maps. The output size of the first FC layers is 100. After hyperparameter tuning, a learning rate of 0.001 is used for Indian Pines and 0.0008 is used for University of Pavia.   

Hyperspectral images contain rich spatial and spectral information, which makes it unique among other types of image data. In the current literature \cite{Lishutao2019}, three CNN-based feature extraction methods can be found for Hyperspectral Image classification, depending on whether they perform spectral, spatial, or spectral-spatial feature extraction. 

\paragraph{\textbf{Spectral models:}} Regarding spectral models, researchers consider the spectral pixels $x_i \in \mathbb{N}^{n_{channels}}$ as the input data, where $n_{channels}$ can either be the number of original bands or a reasonable number of spectral channels extracted using Principle Component Analysis (PCA) or other dimensionality reduction methods. In this work, we apply PCA to reduce the number of channels of raw image into 40 for both Indian Pines and University of Pavia dataset, following the work in \cite{paoletti2019deep}. After dimentionality reduction, classic statistical machine learning models such as Random Forest, Support Vector Machines, and CNN1D are deployed in our work for this method. 

\paragraph{\textbf{Spatial models:}} In spatial models, only spatial information is obtained from the hyperspectral image \cite{eches2011enhancing}. The first step is to reduce the spectral dimension using dimensionality reduction method. The second step is cropping the spectral-reduced image into several patches with size of $d \times d$, covering pixel-centered neighbors. The last step is to extract the spatial information in all patches using CNN2D. In our work, we utilize PCA to reduce the number of channels into one, then crop the original image into pixel-centered patches with size $19 \times 19$. 

\paragraph{\textbf{Spatial-spectral models:}}Utilizing spatial-spectral models entails leveraging both spectral and spatial information within the image to enhance classification accuracy \cite{HeLin2018,fauvel2012advances}. This approach integrates spectral and spatial analyses to offer more detailed and precise insights into the objects present in the image. CNN2D and CNN3D models are both utilized for this technique, with CNN3D models employing 3-D filters to extract high-level spectral-spatial features. 

The implementation details of our spectral model (CNN1D), spatial model (CNN2D), and spatial-spectral model (CNN3D) for processing Indian Pines image are shown in Fig. \ref{fig:dimension}. The output dimension after each layer can be found in the figure. Note that the input of CNN1D is a one dimension vector, the input of CNN2D is a three-dimension tensor, and the input of CNN3D is a four-dimensional tensor.

\subsection{Network Pruning}

\begin{figure*}[!h]
\centering
\includegraphics[width=0.6\textwidth]{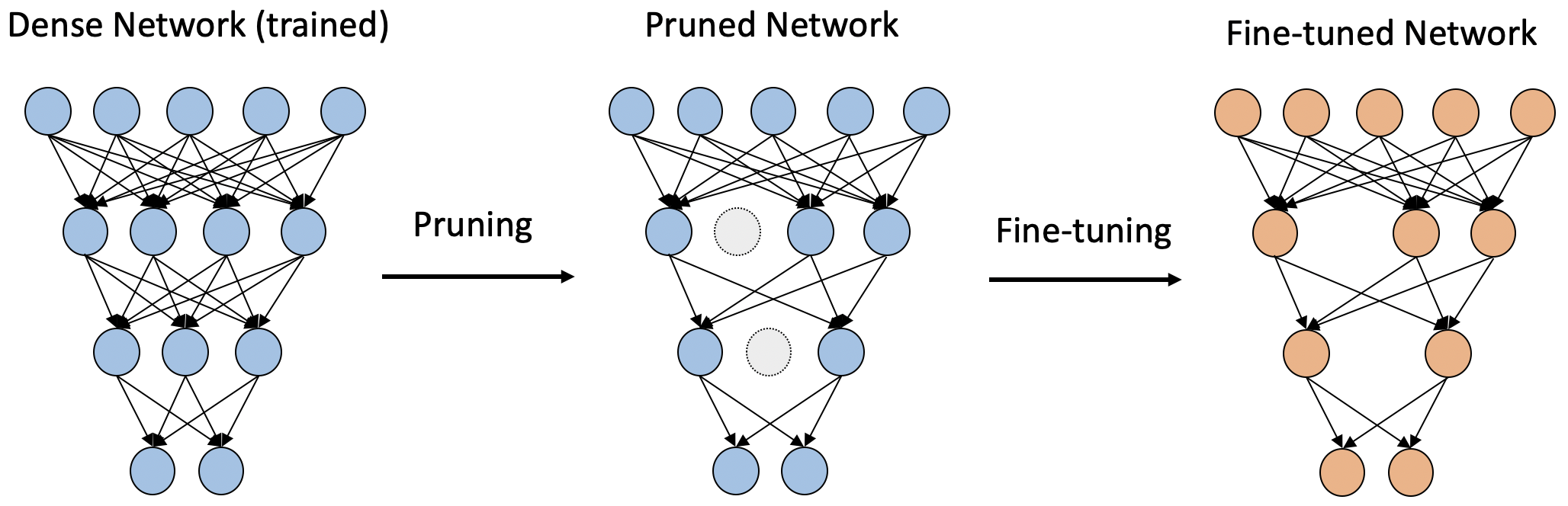}
\caption{\label{fig:pruning pipeline} A typical three-stage network pruning pipeline}
\end{figure*}

Dense, randomly initialized, feed-forward networks have been discovered to contain sub-networks that, when trained independently, can achieve test accuracy comparable to the original network in a similar amount of iterations \cite{Frankle2019}. It has also been found that network pruning can be helpful as an architecture search paradigm since the pruned architecture itself is more important to the efficiency in the final model rather than the learned weights \cite{Liu2018RethinkingTV}. Hence, network pruning is an important technique for both memory size and bandwidth reduction. It allows neural networks to be deployed in constrained environments like embedded systems. Pruning removes redundant weights or neurons that do not significantly contribute to the accuracy.

There are different options on how to prune the networks, depending on the specific granularity. The structure of sparsity is important to both the prediction accuracy and efficiency of hardware architecture \cite{Mao2017}. Both the Conv and FC layers can be pruned. For Conv layers, pruning can occur on a fine-grained level, which refers to pruning individual weights within the model. However, this pruning often leads to unstructured model, where parallel computing no longer works to accelerate the model training and inference \cite{Parashar2017scnn}. Hence, structured pruning techniques have been proposed such as vector-level, kernel-level, and filter-level pruning \cite{he2023structured}. Vector-level pruning involves pruning groups of weights together in predefined patterns, such as vectors or rows of weights within layers. Kernel-level pruning targets entire kernels (sets of weights associated with a particular feature or neuron connection) within a layer. Filter-level pruning (also known as channel pruning) removes entire filters in convolutional layers, which means removing a full output channel. The granularity decreases as the pruning becomes more structured. In this work, we focus on the structured pruning methods since they are more beneficial for acceleration due to regularity. 

Fig. \ref{fig:pruning pipeline} shows a typical three-stage network pruning pipeline, which consists of three stages \cite{Han2015}: 1) train a large, over-parameterized model (many large pre-trained models are publicly available online, such as VGG, ResNet, etc.), 2) prune the trained large model according to a certain criterion, and 3) fine-tune the pruned model to regain the lost performance. Based on our study in the literature survey, fine-tuning is often a necessary step to achieve better model performance. As a result, different fine-tuning strategies have been utilized on network pruning in this work.

\subsubsection{Fine-tuning strategy}

\begin{figure*}[!h]
\centering
\includegraphics[width=0.7\textwidth]{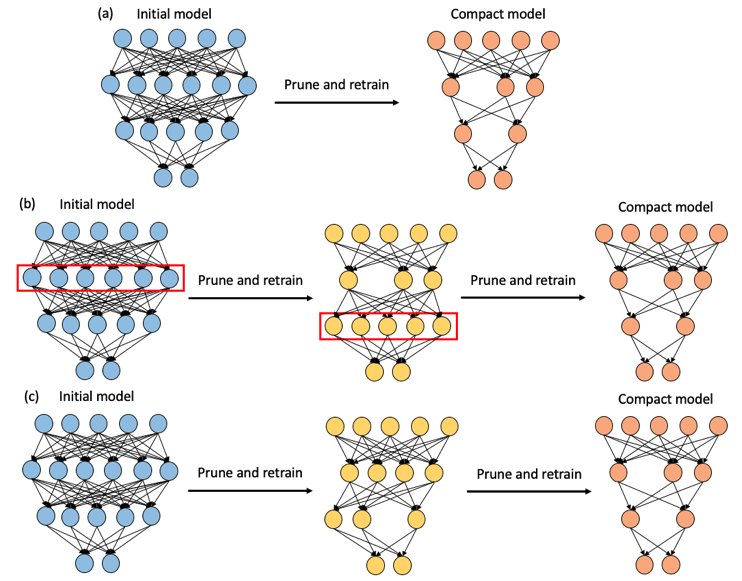}
\caption{\label{fig:finetune} Three types of fine-tuning schemes: (a) One-shot pruning: prune and retrain the model only once (b) Iterative pruning: prune layer by layer, and retrain the model before pruning the next layer (c) Multi-pass pruning: prune and retrain the whole network, then repeat this process until satisfied}
\end{figure*}

There are three fine-tuning strategies applied in network pruning methods: one-shot pruning, iterative pruning, and multi-pass pruning. Fig. \ref{fig:finetune} shows the mechanism for each fine-tuning strategy. Different fine-tuning schemes often produce a different compact final model. 

\paragraph{\textbf{One-shot pruning:}} This classic pruning method prunes the whole pre-trained network only once based on a pre-defined pruning ratio. Then the network is fine-tuned using the same training data based on some specific pruning criterion. The fine-tuning process only occurs once after the network pruning. 

\paragraph{\textbf{Iterative pruning:}} This method prunes filters layer by layer and then retrains the whole network iteratively. In other words, it retrains the model before pruning the next layer. Recent work \cite{li2017pruning} reveals that this fine-tuning strategy shows better results when some filters from the sensitive layers are pruned away, or large portions of the networks are pruned away. However, the iterative process requires many more epochs especially for deep networks. 

\paragraph{\textbf{Multi-pass pruning:}} After pruning the whole network and obtaining a narrow network based on a pre-defined pruning ratio, this method applies the whole process again to learn an even more compact model. We can repeat the entire process multiple times, and each time the network is further pruned based on a pre-defined pruning ratio. This strategy is more efficient than iterative pruning for very deep networks since it is independent of the number of layers. 

\subsubsection{Benchmark methods}

Filter-level pruning removes entire filters (channels) from convolutional layers, resulting in a structured reduction in the model's architecture. This type of pruning aligns well with modern hardware architectures (GPUs, TPUs), which are optimized for dense tensor computations. Hence, in this work we only focus on filter-level pruning methods. Four classic network pruning methods have been deployed as benchmark methods in our experiments. 

\paragraph{\textbf{L1-norm-based filter pruning:}}L1-norm-based filter pruning \cite{li2017pruning} is one of the typical network pruning techniques used in deep learning. It involves removing the entire filter of a convolutional layer based on the L1-norm of the layer's weights. Specifically, in L1-norm-based channel pruning, the convolutional layer's weights are first sorted in descending order based on their absolute values. The channels with the smallest sum of absolute weights are then pruned. The kernels in the subsequent convolutional layers corresponding to the pruned feature maps have also been removed. 

\paragraph{\textbf{ThiNet:}}ThiNet \cite{LuoWL17} is another filter-level pruning method. Unlike traditional pruning methods that often use weight magnitudes, ThiNet employs a data-driven approach to select filters based on their contribution to the next layer’s output. Specifically, this method measures the reconstruction error between the original and pruned feature maps of the next layer, then iteratively prunes the filters that contribute the least to this reconstruction, aiming to minimize the impact on network performance. The advantage of ThiNet is that it is more accurate than magnitude-based pruning, as it directly evaluates the importance of filters based on actual input data.

\paragraph{\textbf{Network slimming:}}Unlike other pruning methods, which directly prune weights or filters, Network slimming \cite{Liu8237560} associates a scaling factor reused from Batch Normalization (BN) layers with each filter in convolutional layers to guide the pruning process. It is a common pratice to use batch normalization (BN) layers in CNNs, where each channel has an associated scaling factor in the batch normalization layer. Network Slimming leverages these scaling factors by applying L1 regularization to them during training, encouraging certain channels to have small or near-zero scaling factors. Pruning is adopted on less important filters by setting the BN scaling parameters to zero. The advantage of network slimming is that it provides an inherent measure of each channel's importance to the model. This method does not depend on a specific architecture, making it versatile for different CNNs. 

\paragraph{\textbf{Soft Filter Pruning (SFP):}}Unlike traditional pruning methods, which permanently remove filters or weights from the network, SFP \cite{He2018SoftFP} temporarily removes filters with low importance scores, but retains the option to "regrow" them during later fine-tuning steps if necessary. To do this, SFP dynamically adjusts the network during the fine-tuning process. If a pruned filter shows potential for importance in subsequent training, it can be "regrown" by updating its weights in later iterations, which allows the method searches with a larger optimization space. The ability to regrow previously pruned filters provides a level of flexibility that most hard pruning methods lack. Filters that were initially pruned can be reactivated if they prove useful during fine-tuning, making the network more adaptable to maintain accuracy. 

Even though network pruning has shown effectiveness in reducing the network memory and computational cost, there are a few known limitations: 1) Before obtaining an optimized pruned network, we still need to train with the full-size network. There are lots of computations during the model developing phase, including tuning the hyperparameters, searching the optimal network architectures, etc.  2) Extreme levels of sparsity using pruning often leads to instability in training \cite{alford2019training}. 3) The community suffers from a lack of standardized benchmarks and metrics, which makes it difficult to compare them each other \cite{blalock2020state}. 

\subsubsection{Network Pruning in remote sensing}

\begin{table*}[!ht]
\centering
\footnotesize
\caption{Existing work of neural network compression in remote sensing applications}
\label{tab:remote sensing}
\begin{tabular}{|c|c|c|c|c|c|}
\hline
\textbf{Approach} &
  \textbf{Paper} &
  \textbf{Network} &
  \textbf{\begin{tabular}[c]{@{}c@{}}Size \\ Reduction\end{tabular}} &
  \textbf{Datasets} &
  \textbf{Application} \\ \hline
 &
  \begin{tabular}[c]{@{}c@{}}\cite{Xiaokai2024}\end{tabular} &
  \cellcolor[HTML]{FFFFFF}{\color[HTML]{333333} \begin{tabular}[c]{@{}c@{}}MobileNet,\\ ResNet\end{tabular}} &
  \cellcolor[HTML]{FFFFFF}{\color[HTML]{333333} 80\%} &
  \cellcolor[HTML]{FFFFFF}{\color[HTML]{333333} \begin{tabular}[c]{@{}c@{}}OPTIMAL-31\end{tabular}} &
  \cellcolor[HTML]{FFFFFF}{\color[HTML]{333333} \begin{tabular}[c]{@{}c@{}}Scene \\    Classification\end{tabular}} \\ \cline{2-6} 
 &
  \begin{tabular}[c]{@{}c@{}}\cite{Chengcheng2023}\end{tabular} &
  \cellcolor[HTML]{FFFFFF}\begin{tabular}[c]{@{}c@{}}2D-CNN\end{tabular} &
  \cellcolor[HTML]{FFFFFF}97\% &
  \cellcolor[HTML]{FFFFFF}\begin{tabular}[c]{@{}c@{}}Indian Pines, \\ Salinas, Xiongan\end{tabular} &
  \cellcolor[HTML]{FFFFFF}\begin{tabular}[c]{@{}c@{}}Hyperspectral\\     Classification\end{tabular} \\ \cline{2-6} 
 &
  \cellcolor[HTML]{FFFFFF}{\color[HTML]{333333} \begin{tabular}[c]{@{}c@{}}\cite{Yiheng2023}\end{tabular}} &
  \cellcolor[HTML]{FFFFFF}{\color[HTML]{333333} VGG, ResNet} &
  \cellcolor[HTML]{FFFFFF}{\color[HTML]{333333} 70\%} &
  \cellcolor[HTML]{FFFFFF}\begin{tabular}[c]{@{}c@{}}AID, NWPU45, \\ PatternNet, \\ WHU19 \end{tabular} &
  \cellcolor[HTML]{FFFFFF}{\color[HTML]{333333} \begin{tabular}[c]{@{}c@{}}Scene\\ Classification \end{tabular}} \\ \cline{2-6} 
 &
  \begin{tabular}[c]{@{}c@{}}\cite{Xiaokai2023}\end{tabular} &
  \cellcolor[HTML]{FFFFFF}2D-CNN &
  \cellcolor[HTML]{FFFFFF}80\% &
  \cellcolor[HTML]{FFFFFF}\begin{tabular}[c]{@{}c@{}}Bijie \end{tabular} &
  \cellcolor[HTML]{FFFFFF}\begin{tabular}[c]{@{}c@{}}Landslides \\Recognition\end{tabular} \\ \cline{2-6} 
\multirow{-5}{*}{\textbf{Pruning}} &

  \begin{tabular}[c]{@{}c@{}}\cite{Xianpeng2022}\end{tabular} &
  \cellcolor[HTML]{FFFFFF}AlexNet, VGG &
  \cellcolor[HTML]{FFFFFF}- &
  \cellcolor[HTML]{FFFFFF}\begin{tabular}[c]{@{}c@{}}UC Merced, \\ NWPU-RESISC45 \end{tabular} &
  \cellcolor[HTML]{FFFFFF}\begin{tabular}[c]{@{}c@{}}Scene\\     Classification\end{tabular} \\ \hline
 &
  \begin{tabular}[c]{@{}c@{}}\cite{Palle2024}\end{tabular} &
  \cellcolor[HTML]{FFFFFF}3D-CNN &
  \cellcolor[HTML]{FFFFFF}- &
  \cellcolor[HTML]{FFFFFF}\begin{tabular}[c]{@{}c@{}} Indian Pines, \\ Pavia University \end{tabular} &
  \cellcolor[HTML]{FFFFFF}\begin{tabular}[c]{@{}c@{}}Hyperspectral\\     Classification\end{tabular} \\ \cline{2-6} 
\multirow{-2}{*}{\textbf{Quantization}} &
\begin{tabular}[c]{@{}c@{}} \cite{Weiwei2023}\end{tabular} &
  \cellcolor[HTML]{FFFFFF}1D-CNN &
  \cellcolor[HTML]{FFFFFF}63\% &
  \cellcolor[HTML]{FFFFFF}\begin{tabular}[c]{@{}c@{}}Pavia University, \\ Salinas, \\ Houston  \end{tabular} &
  \cellcolor[HTML]{FFFFFF}\begin{tabular}[c]{@{}c@{}}Hyperspectral\\   Classification\end{tabular} \\ \cline{2-6} &
  \begin{tabular}[c]{@{}c@{}} \cite{Mei2022}\end{tabular} &
  \cellcolor[HTML]{FFFFFF}2D-CNN &
  \cellcolor[HTML]{FFFFFF}13.6\% &
  \cellcolor[HTML]{FFFFFF}\begin{tabular}[c]{@{}c@{}}Pavia University, \\ Salinas\end{tabular} &
  \cellcolor[HTML]{FFFFFF}\begin{tabular}[c]{@{}c@{}}Hyperspectral\\Classification\end{tabular} \\ \hline
 &
  \cellcolor[HTML]{FFFFFF}{\color[HTML]{333333} \begin{tabular}[c]{@{}c@{}} \cite{Yangyang2024} \end{tabular}} &
  \cellcolor[HTML]{FFFFFF}{\color[HTML]{333333} U-Net} &
  \cellcolor[HTML]{FFFFFF}{\color[HTML]{333333} 43.6\%} &
  \cellcolor[HTML]{FFFFFF}{\color[HTML]{333333} \begin{tabular}[c]{@{}c@{}}Vaihingen, \\ Potsdam\end{tabular}} &
  \cellcolor[HTML]{FFFFFF}{\color[HTML]{333333} \begin{tabular}[c]{@{}c@{}}Semantic\\     Segmentation\end{tabular}} \\ \cline{2-6} 
 &
  \cellcolor[HTML]{FFFFFF}{\color[HTML]{333333} \begin{tabular}[c]{@{}c@{}}\cite{Chi2022}     \end{tabular}} &
  \cellcolor[HTML]{FFFFFF}{\color[HTML]{333333} \begin{tabular}[c]{@{}c@{}}2DCNN,\\ 3DCNN\end{tabular}} &
  \cellcolor[HTML]{FFFFFF}{\color[HTML]{333333} -} &
  \cellcolor[HTML]{FFFFFF}{\color[HTML]{333333} \begin{tabular}[c]{@{}c@{}}India Pines, \\ University of Pavia, \\ KSC, Botswana\end{tabular}} &
  \cellcolor[HTML]{FFFFFF}{\color[HTML]{333333} \begin{tabular}[c]{@{}c@{}}Hyperspectral\\  Classification\end{tabular}} \\ \cline{2-6} 
 &
  \cellcolor[HTML]{FFFFFF}{\color[HTML]{333333} \begin{tabular}[c]{@{}c@{}}\cite{Xu2022}    \end{tabular}} &
  \cellcolor[HTML]{FFFFFF}{\color[HTML]{333333} ResNet} &
  \cellcolor[HTML]{FFFFFF}{\color[HTML]{333333} 84.6\%} &
  \cellcolor[HTML]{FFFFFF}{\color[HTML]{333333} \begin{tabular}[c]{@{}c@{}}University of Pavia, \\ Salinas, Houston\end{tabular}} &
  \cellcolor[HTML]{FFFFFF}{\color[HTML]{333333} \begin{tabular}[c]{@{}c@{}}Hyperspectral\\ Classification\end{tabular}} \\ \cline{2-6} 
 &
  \cellcolor[HTML]{FFFFFF}{\color[HTML]{333333} \begin{tabular}[c]{@{}c@{}}\cite{Yangyang2024}    \end{tabular}} &
  \cellcolor[HTML]{FFFFFF}{\color[HTML]{333333} 2D-CNN} &
  \cellcolor[HTML]{FFFFFF}{\color[HTML]{333333} 99.5\%} &
  \cellcolor[HTML]{FFFFFF}{\color[HTML]{333333} MSTAR} &
  \cellcolor[HTML]{FFFFFF}{\color[HTML]{333333} \begin{tabular}[c]{@{}c@{}}Hyperspectral\\Face Recognition\end{tabular}} \\ \cline{2-6} 
\multirow{-5}{*}{\textbf{\begin{tabular}[c]{@{}c@{}}Knowledge \\ Distillation\end{tabular}}} &
  \cellcolor[HTML]{FFFFFF}{\color[HTML]{333333} \begin{tabular}[c]{@{}c@{}}\cite{Ronghua2022}     \end{tabular}} &
  \cellcolor[HTML]{FFFFFF}{\color[HTML]{333333} 2D-CNN} &
  \cellcolor[HTML]{FFFFFF}{\color[HTML]{333333} -} &
  \cellcolor[HTML]{FFFFFF}{\color[HTML]{333333} \begin{tabular}[c]{@{}c@{}}Indian Pines, \\ Pavia University, \\ Kennedy Space Center, \\ Pavia
  Center\end{tabular}} &
  \cellcolor[HTML]{FFFFFF}{\color[HTML]{333333} \begin{tabular}[c]{@{}c@{}}Hyperspectral\\     Classification\end{tabular}} \\ \hline
\end{tabular}
\end{table*}

Pruning techniques can be especially useful in remote sensing applications, where memory constraints frequently result in restricted processing capability. By reducing overfitting, pruning can also assist in enhancing the generalization capabilities of neural networks. Some of the existing work has been shown in Table \ref{tab:remote sensing}.

Several recent studies have explored neural network pruning techniques for remote sensing tasks. Xiaokai et al. \cite{Xiaokai2024} propose a block-level pruning method based on semantic similarity analysis, where localization maps are used to measure the similarity between feature maps of the original and pruned networks, and blocks that cause minimal semantic information loss are retained; experiments on the OPTIMAL-31 dataset show that the model size can be reduced by up to 80\% without significant accuracy degradation. Chengcheng et al. \cite{Chengcheng2023} introduce LiteCCLKNet, a lightweight CNN for hyperspectral image (HSI) classification that replaces conventional 2D convolutions with 1D and $1\times1$ convolutions and incorporates a criss-cross large kernel (CCLK) module with self-attention to capture long-range dependencies; the model achieves competitive performance on Indian Pines, Salinas, and Xiongan datasets while reducing parameters by up to 97\% compared with VGG. Yiheng et al. \cite{Yiheng2023} develop an energy-based filter pruning approach in which eigenvalues of weight tensors are used to compute the energy of convolutional layers, and filters with low energy are removed, based on the observation that layers with sharper parameter distributions contribute less to representation capacity; evaluations on AID, WHU19, and NWPU45 demonstrate up to 70\% computational cost reduction on VGG-16 and ResNet-50 with minimal accuracy loss. Xiaokai et al. \cite{Xiaokai2023} further propose a filter-level pruning strategy for landslide recognition that leverages evolutionary computation with continual masking to optimize pruning decisions, achieving up to 80\% reduction in computational cost on the Bijie landslide dataset without significant performance degradation. Finally, Xianpeng et al. \cite{Xianpeng2022} introduce an interpretable CNN-based pruning method that incorporates an additional loss term to encourage each convolutional filter to represent meaningful object parts, allowing non-interpretable filters to be pruned; experiments on AlexNet and VGG using the UC Merced and NWPU-RESISC45 datasets demonstrate effective model compression while maintaining classification performance.

In this study, we implemented and evaluated the benchmark network pruning methods due to their robustness, wide adoption, and proven effectiveness across various domains. These established methods provide a reproducible and comparable baseline, essential for consistent benchmarking in hyperspectral image classification. Additionally, while the above recent pruning techniques offer some advancements, their application to hyperspectral data remains largely unexplored. They often combine network pruning with other model compression or machine learning algorithms, which introduce significant complexity that may not align with the practical constraints of this study. Future work can build upon these results by incorporating recent, remote sensing-specific pruning methods to further explore their impact on hyperspectral image classification.

\subsection{Quantization}

Historically, the weights of most neural networks use single-precision floating-point format (FP32), while its representation sometimes has greater precision than needed. Compressing CNNs by reducing precision values has been previously studied \cite{Balzer1991}. The objective of quantization techniques is to perform computations and store tensors at lower bit-widths than floating-point precision (e.g., integer approximation) in order to accelerate the inference phase without a significant drop in performance. Quantization has been shown to reduce overfitting, as well as reducing memory and computational cost. 

To better formulate quantization techniques, assume the resulting quantized values are uniformly spaced and we want to map a floating point value $x \in [\alpha, \beta]$ to a b-bit integer $x_{q} \in [\alpha_{q}, \beta_{q}]$:

\begin{equation} 
\label{eqn: uniform quantization}
    x_{q} = round(\frac{x}{S}) - Z
\end{equation}

where $S$ is a real valued scaling factor, and $Z$ is an integer zero point. Given the clipping range (a typical choice is to use the min/max of the numbers), $S$ can be computed as:

\begin{equation} 
\label{eqn: S value}
    S = \frac{\beta_{q} - \alpha_{q}}{2^b - 1}
\end{equation}

Similarly,a dequantization operation is to recover the floating-point value $x$ from the quantized values $x_{q}$:

\begin{equation} 
\label{eqn: dequantization}
    \tilde{x} = S(x_{q} + Z)
\end{equation}

\subsubsection{Quantization Modes} 
There are three types of quantization modes, which is shown in Table \ref{tab:qmodes}.

\begin{table*}[!ht]
\centering
\begin{tabular}{|
>{\columncolor[HTML]{FFFFFF}}c |
>{\columncolor[HTML]{FFFFFF}}c |
>{\columncolor[HTML]{FFFFFF}}c |
>{\columncolor[HTML]{FFFFFF}}c |}
\hline
{\color[HTML]{000000} Quantization Modes} &
  {\color[HTML]{000000} Data Requirements} &
  {\color[HTML]{000000} Inference Latency} &
  {\color[HTML]{000000} Inference Accuracy Loss} \\ \hline
{\color[HTML]{000000} Dynamic Quantization}        & {\color[HTML]{000000} No Data}        & {\color[HTML]{000000} Fast}    & {\color[HTML]{000000} Smallest} \\ \hline
{\color[HTML]{000000} Static Quantization}         & {\color[HTML]{000000} Unlabeled data} & {\color[HTML]{000000} Fastest} & {\color[HTML]{000000} Small}    \\ \hline
{\color[HTML]{000000} Quantization Aware Training} & {\color[HTML]{000000} Labeled data}   & {\color[HTML]{000000} Fastest} & {\color[HTML]{000000} Smallest} \\ \hline
\end{tabular}
\caption{Different quantization modes}
\label{tab:qmodes}
\end{table*}

\paragraph{\textbf{Dynamic quantization:}}Prior to the inference runtime, weights are quantized into integers in dynamic quantization \cite{Sun9711308}. However, since networks do not already know the scales and zero points for the output, we must first obtain the floating point output tensors, identify the tensor's $(alpha, beta)$, compute the scale and zero point, and then dynamically convert the tensor to an integer during runtime. The benefit of dynamic quantization is that no calibration data are required before making an inference. However, this technique has a runtime overhead problem with computing scales and zero points.

\paragraph{\textbf{Static quantization:}}Static quantization precomputes scales and zero points \cite{Das2018}. The neural network is executed on some representative unlabeled data to obtain the distribution statistics for all weights. This information will then be used to compute the scales and zero points. One advantage of static quantization is that it does not increase overhead. However, if the data is not representative, the scales and zero points produced during inference might not accurately represent the actual situation. 

\paragraph{\textbf{Quantization Aware Training (QAT):}}In QAT, we take the quantization effect into consideration during training \cite{Jacob2018quantization}. The locations where quantization takes place will be covered with "fake" quantization modules. These modules have two purposes: 1) practice clamping and rounding; 2) monitor the scales and the zero points of the weights while working out. Using the data stored in the "fake" quantization modules, the floating point model might be instantly converted to a quantized integer model after training is complete.

\subsubsection{Quantization in remote sensing}

In order to create more effective and portable models for hyperspectral image classification, network quantization can be utilized in this application. Reducing the model's memory and compute needs can considerably shorten processing times and lower the hardware cost, which can be especially helpful for real-time applications where hardware resource is limited. Table \ref{tab:remote sensing} provides some examples of recent remote sensing research using network quantization.

Recent studies have also explored quantization techniques to reduce the computational and memory requirements of deep learning models for hyperspectral image (HSI) classification. Palle et al. \cite{Palle2024} apply dynamic quantization to the Hybrid Spectral CNN (HybridSN) by replacing real-valued weights with binary weights, significantly reducing memory usage while maintaining model efficiency. To mitigate quantization noise introduced by binarization, the absolute mean of the original weights is used as a scaling factor, and both activations and weights are quantized by rounding values to the nearest multiple of this scale. Experiments on two datasets demonstrate a 2.5$\times$ inference speedup with only a 0.5\% accuracy reduction compared to the full-precision model. Weiwei et al. \cite{Weiwei2023} propose a curriculum learning-based progressive binarization strategy for CNN-based HSI classification, where a 1D-CNN architecture with eight convolutional layers progressively binarizes all layers except the first. An adaptive gradient scaling module is introduced to stabilize training and improve optimization of the binary network, achieving up to 63\% memory reduction without significant performance loss on the Pavia University, Salinas, and Houston datasets. Mei et al. \cite{Mei2022} develop a step activation quantization method that constrains CNN inputs to low-bit integer representations using nonlinear uniform quantization, enabling floating-point operations to be replaced with integer operations during inference. Their approach significantly accelerates inference, achieving approximately 10$\times$ speedup and reducing memory consumption by 13.6$\times$ compared with the original real-valued network.

\subsection{Knowledge Distillation}

Recently, another model compression technique that has drawn more attention from the research community is knowledge distillation \cite{moslemi2024survey,Beyer2022,gou2021knowledge,wang2021knowledge}. Under practical constraints, it is used to transfer information from a big, complicated model to a smaller, simpler model without significant performance loss. In order to produce predictions on a set of training data, a large model known as the teacher model must be pretrained. The goal of the student model's optimization during training is to learn from the teacher model's knowledge representations and match its predictions. With minimal computing resources needed to run, this method enables the student model to perform comparably to the teacher model, making it more appropriate to deploy devices with constrained resources, such as mobile phones or embedded systems. 

\subsubsection{Knowledge types in KD}

Depending on the knowledge types being distilled, knowledge distillation methods can be categorized into three groups: response-based knowledge distillation, feature-based knowledge distillation, and relation-based knowledge distillation \cite{yang2023categories}. 
%The schematic illustration of three types of knowledge source in a deep teacher network is shown in Fig. \ref{fig:knowledge}. 

\paragraph{\textbf{Response-based knowledge:}}In response-based knowledge distillation, the student model is trained to mimic the responses of the teacher model rather than trying to directly match its output probabilities or logits. Specifically, during training, the student model is presented with the same input as the teacher model and is trained to produce the same output as the teacher model. The most popular response-based knowledge for image classification is known as soft targets \cite{Hinton2015DistillingTK}. Specifically, soft targets are the probabilities that the input belongs to the classes and can be estimated by a softmax function. We can further represent the distillation loss by computing the divergence loss (e.g., Kullback-Leibler divergence) of logits of teacher and student, $z_t$ and $z_s$. This loss can be used to guide the learning of a student network from the teacher network.

Response-based knowledge can be used for different types of model predictions. For example, the response in the object detection task may provide both the logits and the offset of a bounding box \cite{Chen2017LearningEO}. It has been further explored to address the information of ground-truth labels as the conditional targets \cite{Meng2019ConditionalTL}. Although response-based knowledge distillation is straightforward to implement, it typically only utilizes the final layer's output and ignores the teacher model's critical intermediate-level guidance, which is especially important when employing very deep neural networks for representation learning. Furthermore, it is worth noting that response-based knowledge distillation only applies to supervised learning settings.

\paragraph{\textbf{Feature-based knowledge:}}Deep neural networks excel at acquiring various levels of feature representation with increasing levels of abstraction. The output from both the final and intermediate layers can guide the training of the student model \cite{yang2024vitkd,li2022knowledge,heo2019knowledge}. Instead of solely relying on the final layer's output, feature-based knowledge distillation aims to replicate the features or activations obtained in the intermediate levels of the teacher model \cite{sun2020contrastive,aguilar2020knowledge}. Consequently, the student model can utilize the knowledge gained by the teacher model to enhance its performance. This approach is particularly advantageous for deep neural networks, as the intermediate representations contain valuable information about the input data.

Feature-based knowledge distillation can offer valuable guidance for the student model's learning process, but effectively identifying the appropriate hint layers from the teacher model and the corresponding guided layers from the student model requires further investigation. Moreover, given the potential size differences in the hint and guided layers, it is crucial to explore methods for appropriately aligning the feature representations of the teacher and student networks.

\paragraph{\textbf{Relation-based knowledge:}}Knowledge that represents the relationship between feature maps can also be used to guide the learning process of a student model. Some of the examples are correlations between feature maps, graphs, similarity matrices, feature embeddings, data samples, or probabilistic distributions based on feature representations \cite{ramakrishnan2020relationship,dong2020distilling,yang2022cross}. This method is known as relation-based knowledge distillation \cite{park2019relational}. By explicitly modeling the relationships between the outputs of the teacher and student models, relation-based knowledge distillation can help the student model generalize better to new examples. However, this method is more complex than traditional knowledge distillation methods, making it more challenging to implement and optimize. Furthermore, how to model the related information from feature maps or data samples as knowledge still deserves further study.

\subsubsection{Distillation modes in KD}

\begin{figure*}[!h]
\centering
\includegraphics[width=0.65\textwidth]{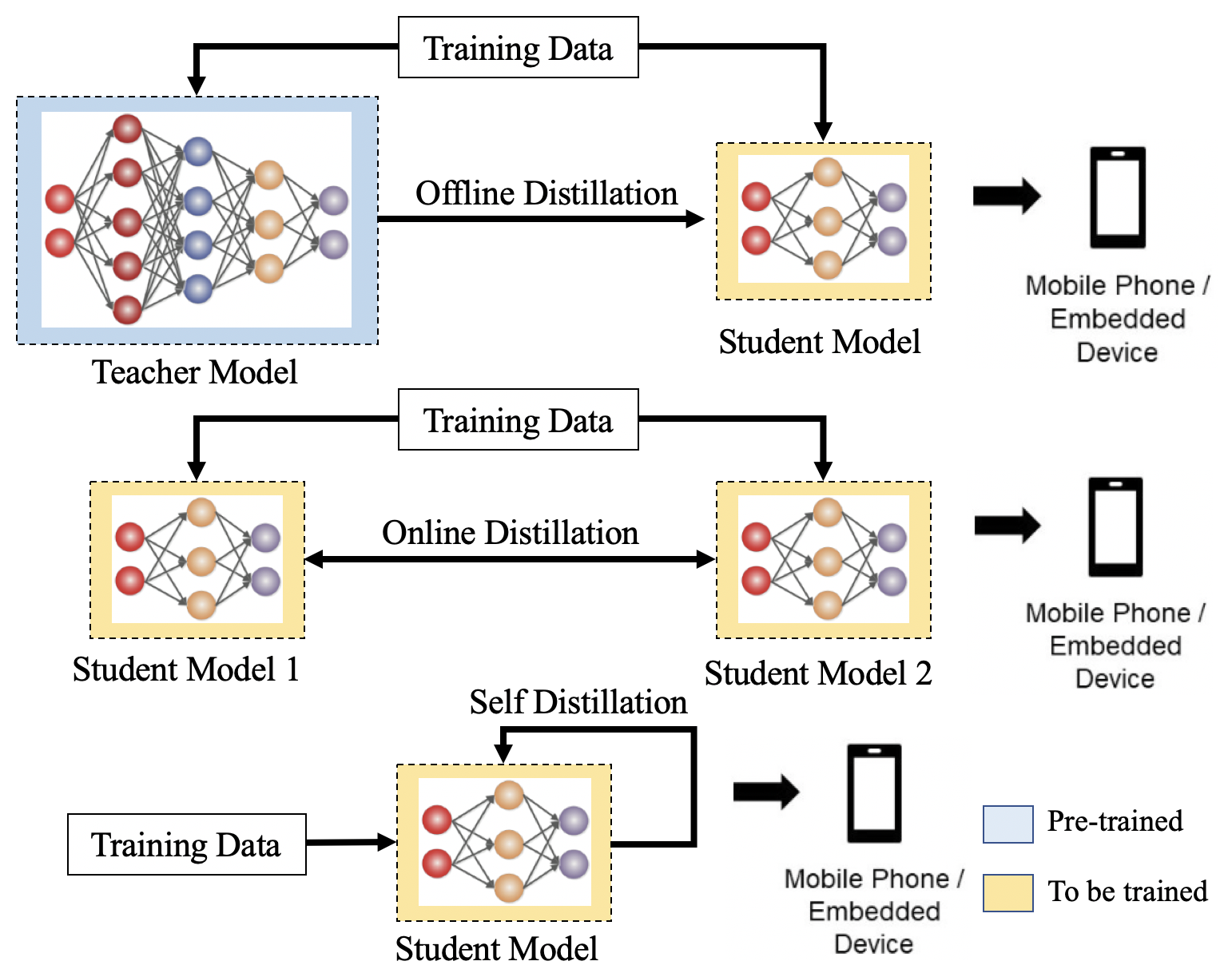}
\caption{\label{fig:modes} Different distillations modes. Blue: networks are learned before distillation; Yellow: networks are learned during distillation}
\end{figure*}

According to whether the teacher model is updated simultaneously with the student model or not, there are three options for the distillation modes (i.e., training scheme): offline distillation, online distillation, and self-distillation, as shown in Fig. \ref{fig:modes}. In online KD approaches, there is no pre-trained network and all models are updated simultaneously in a single end-to-end training process. Online KD can be operationalized using parallel computing thus making it a highly efficient method. In self-distillation, the same network is used for both the teacher and student models. It can be considered as a special case of online KD method. For example, knowledge from deeper layers of a network can be used to train the shallow layers. They are different from offline KD methods since no pre-trained teacher model is required, which is usually a large deep neural network and consumes huge memory. 

\subsubsection{Benchmark Methods}

We implemented and evaluated six offline distillation methods, four online distillation methods, and four self distillation methods on both Indian Pines and Pavia University datasets. The detail for all methods has been discussed below. 

\paragraph{\textbf{Soft Targets:}}The most popular response-based knowledge for image classification is known as soft targets \cite{Hinton2015DistillingTK}. Specifically, soft targets are the probabilities that the input belongs to the classes and can be estimated by a softmax function as shown in Equation \ref{eqn: soft targets}, where $z_i$ is the logit for the $i$-th class, and a temperature factor T is introduced to control the importance of each soft target. As stated in the paper, soft targets contain the informative dark knowledge from the teacher model. We can further represent the distillation loss by computing the divergence loss (e.g., Kullback-Leibler divergence) of logits of teacher and student, $z_t$ and $z_s$. This loss can be used to guide the learning of a student network from the teacher network. 

\begin{equation} 
\label{eqn: soft targets}
    p(z_i, T) = \frac{e^{(z_i/T)}}{\Sigma_{j}e^{(z_j/T)}}
\end{equation}

\paragraph{\textbf{FitNets:}} This work \cite{Romero2014FitNetsHF} introduces a hint-based training process, which produces a student model that is deeper and thinner than the teacher. The approach makes use of both the student's ultimate performance and the teacher's intermediate representations as cues to enhance the training process. Additionally, a fully-connected trainable regressor is added to the guided layer to match the output size of the hint layer since the student intermediate hidden layer will typically be smaller than the teacher's intermediate hidden layer. This provides for a trade-off between training deeper students who can generalize better or train faster, depending on the specified student capacity. 

\paragraph{\textbf{Attention Transfer:}} The spatial attention map of a convolutional network shows where the network focuses in order to classify the given image. The authors of this paper \cite{Zagoruyko2017} utilized this type of information to boost the performance of a student CNN network by requiring it to mimic the attention maps of a strong teacher network. Both activation-based and gradient-based spatial attention maps are used for knowledge transfer. It has been presented that activation-based attention transfer can be combined with knowledge distillation, and this method improves performance compared to full activation transfer.  

\paragraph{\textbf{Correlation Congruence:}} The weakness of instance-level congruence is that it’s challenging for the student to learn a mapping function identical to the teacher due to the gap of network capacity between teacher and student. In this work \cite{Peng2019}, it is claimed that beyond instance congruence, the correlation between instances is also valuable knowledge for promoting the performance of the student. The authors proposed the Correlation-Congruence Knowledge Distillation (CCKD) \cite{Peng2019} approach, which enables transferring both correlation information between instances in addition to instance-level information. To better capture the correlation between instances, a generalized kernel-based method using gaussian RBF is suggested. The majority of teacher-student frameworks, including soft targets and hint-based learning techniques, can readily use the CCKD. 

% In recent years, knowledge distillation has become a hot topic in the field of machine learning. For example, in year 2022, more than 30 papers published in top CS conferences focus on research advance in knowledge distillation. We implemented two recent KD work and tested their performance on remote sensing datasets. 

\paragraph{\textbf{SimKD (Simple KD):}}In order to simplify the model development and interpretation of knowledge distillation, SimKD \cite{Chen2022KnowledgeDW} directly reuses the discriminative classifier from the pre-trained teacher model for student inference and trains a student encoder using feature alignment with a single L2 loss in order to reduce the performance gap between teachers and students. It is shown that as long as their extracted features are completely aligned, the student model can accomplish the exact same performance as the teacher model. This method can be used to a variety of teacher and student architectures by utilizing an additional projector that aids in the student encoder's matching with the teacher's classifier.

\paragraph{\textbf{CA-MKD (Confidence-Aware Multi-teacher KD):}}Another format of knowledge distillation is to take advantage of various knowledge sources from multiple teachers in order to improve student performance. Existing research, however, typically combine knowledge from different sources by averaging over many teacher predictions or by integrating them using other different label-free methodologies, which may lead student network sensitive to low-quality teacher predictions. With the use of ground-truth labels, CA-MKD \cite{Zhang2021ConfidenceAwareMK} adaptively assigns sample-wise reliability for each teacher prediction, giving large weights to those teacher predictions that are close to the ground-truth labels. Using CAM-KD, both the output and intermediate layers of the teacher model are utilized to boost the student performance. 

\paragraph{\textbf{Collaborative Learning with Intermediate-Level Representation (CL-ILR):}}To increase generalization and robustness to label noise without additional cost, the authors in \cite{Song2018cl} proposed a framework, CL-ILR, to enable collaborative learning that simultaneously trains several classifier heads of the same network. The output from multiple classifier heads on the same example gives each classifier more information and regularization, which enhances model generalization. Additionally, Intermediate-Level Representation (ILR) is introduced by sharing into co-distillation, which improves training time and memory efficiency and generalization error. Empirical findings using the CIFAR and ImageNet datasets show that deep neural networks that are collaboratively learned as a group significantly reduce generalization errors and improve robustness to label noise.

\paragraph{\textbf{Deep Mutual Learning (DML):}}Unlike the one-way manner of knowledge transfer from a static, predefined teacher network, DML \cite{zhang2018deep} allows a group of students study collaboratively and impart knowledge to one another during the training process. Specifically, using this mutual learning approach, we compute the predictions of the two models and update the parameters of both networks in accordance with one another's predictions at each iteration. According to the studies done in this paper, collaborative learning by small peers even produces better results when compared to distillation by a static, pre-trained big network.

\paragraph{\textbf{On-the-fly Native Ensemble (ONE):}}In this work \cite{Zhu2018knowledge}, the authors propose a one-stage online distillation learning approach called On-the-fly Native Ensemble (ONE). Unlike offline KD methods that rely on a strong pre-trained teacher network, online methods often allow networks to learn from each other during training. In order to improve the learning of the student network, ONE specifically trains one multi-branch network while simultaneously establishing a powerful teacher. The target network and the multiple auxiliary branches share the low-level layers. The teacher model is created using the collective models created by each branch with shared layers. During inference time, auxiliary branches may be either preserved or eliminated depending on the necessity for deployment efficiency. On the four image classification benchmarks of CIFAR-10, CIFAR-100, SVHN, and ImageNet, experiments demonstrate that a variety of deep networks may profit from the ONE strategy.

\paragraph{\textbf{Online Knowledge Distillation with Diverse Peers (OKDDip):}}Although using group-based targets is effective for distillation without teacher engagement, it may diminish diversity among group members using basic aggregation operations, resulting in locally optimal solutions. Another study, OKDDip \cite{chen2020online}, conducts two-tier distillation during training with several secondary peers and one group leader. In the first-tier distillation, each secondary peer possesses a distinct set of aggregation weights generated using an attention-based mechanism to create their own targets from the predictions of other secondary peers. The second-tier distillation is then used to transfer the knowledge from the ensemble of secondary peers to the group leader, which is the model employed for making predictions. It is shown that peer diversity is maintained to a relatively large extent throughout group learning by distilling from distinct target distributions derived with weights from an attentionbased component.

\paragraph{\textbf{Teacher-Free Knowledge Distillation (TF-KD):}}One early work of self-distillation is teacher-free knowledge distillation \cite{yuan2020revisiting}, where a strong teacher model is not available. In TF-KD, a manually designed virtual teacher with 100\% accuracy is used to guide the learning process of a student network. The authors propose to combine KD and LSR (Label Smoothing Regularization) to build a simple teacher model that will output distribution for classes as the following:

\begin{equation}
    p^{d}(k)=
    \begin{cases}
    a, & \text{if $k=c$}.\\
    \frac{1-a}{K-1}, & \text{if $k\neq{c}$}.
    \end{cases}
\end{equation}

where $K$ is the total number of classes, $c$ is the correct label and $a$ is the correct probability for the correct class. The manually designed teacher model outputs soft targets with 100\% classification accuracy, and also has the smoothing property of label smoothing. Experiment results show this Tf-KD method can achieve comparable results with Normal KD in image classification, even though a strong teacher model is not available.

\paragraph{\textbf{Class-wise Self-Knowledge Distillation (CS-KD):}}CS-KD \cite{yun2020regularizing} is another self-distillation method where a novel regularization strategy is proposed to penalize the predicted distribution of comparable samples. Specifically, during training, CS-KD distills the predictive distribution between various samples of the same label. This leads to the regularization of a single network's dark knowledge by guiding it to deliver more meaningful and consistent predictions in a class-wise way. As a result, it lowers intra-class variability and mitigates overconfident predictions. Their results on numerous image classification tasks show that this simple strategy may improve not only the generalization ability but also the calibration performance of CNNs.

\paragraph{\textbf{Progressive Refinement of Targets (PS-KD):}}PS-KD \cite{kim2021self} is another simple yet effective regularization approach, which progressively distills a model's knowledge to soften hard targets during training. An example of hard target is one-hot vectors. This method enables a student to become their own teacher by interpreting their knowledge within a framework of knowledge distillation. Specifically, targets are adjusted adaptively during each training epoch by merging the ground truth and past predictions from the model itself. It has been claimed that PS-KD provides an effect of hard example mining by rescaling gradients according to difficulty in classifying examples. This method has been evaluated on various tasks, including image classification, object detection, and machine translation, which indicates it improves the generalization of neural networks.

\paragraph{\textbf{Data-Distortion Guided Self-Distillation (DDGSD):}}The authors of DDGSD \cite{Xutingbing2019} offer a data-focused self-distillation approach to transfer knowledge between multiple distorted versions of the same training data without relying on supporting models. The benefit of this method is that the capacity of a single network can be enlarged through learning consistent global feature distributions and class probabilities across various distorted data. In comparison to previous knowledge distillation methods (i.e., teacher-to-student and student-to-student), the proposed method efficiently employs data-to-data knowledge distillation to thoroughly discard the dependence of supporting models, effectively lowering the training cost of obtaining a small network of high accuracy. Experiments on numerous datasets, including CIFAR-10, CIFAR-100, and ImageNet, demonstrate that the suggested technique may significantly increase the generalization performance of several network architectures.

\subsubsection{Knowledge distillation in remote sensing}

In recent years, knowledge distillation has been used to create smaller models that can be deployed on low-power devices or in environments with limited computational resources for remote sensing applications, such as hyperspectral image classification, object detection, and semantic segmentation.

Knowledge distillation has also been widely explored to compress deep learning models while maintaining performance in remote sensing tasks. Chen et al. \cite{Chen2018} apply a knowledge distillation framework for remote sensing scene classification, where a smaller student network is trained to mimic the softmax outputs of a larger teacher model. Experiments on four benchmark datasets—AID, UCMerced, NWPU-RESISC, and EuroSAT—demonstrate that this approach significantly improves the performance of compact models, particularly for datasets with lower spatial resolution, many categories, or limited training samples. Chi et al. \cite{Chi2022} propose a self-supervised HSI classification framework that combines knowledge distillation with adaptive soft label generation, where soft labels produced by a teacher model guide the training of a student model through metric-based learning; experiments on Indian Pines, University of Pavia, Kennedy Space Center, and Botswana datasets show notable improvements in classification accuracy. Xu et al. \cite{Xu2022} further employ knowledge distillation in a class-incremental learning framework for hyperspectral image classification, enabling the model to learn new land-cover classes without forgetting previously learned ones, with validation on the University of Pavia, Salinas, and Houston datasets. Min et al. \cite{Min2019} introduce a micro CNN (MCNN) for real-time synthetic aperture radar (SAR) target detection, where an 18-layer deep CNN is compressed into a two-layer network using a gradual distillation strategy that trains teacher and student models simultaneously; experiments on the MSTAR dataset show that the resulting model achieves a 177$\times$ reduction in memory usage and 12.8$\times$ lower computational cost. Similarly, Ma et al. \cite{Ma2021} propose Light-YOLOv4, an edge-device-oriented object detection framework for remote sensing images that combines pruning, quantization, and knowledge distillation during training; experiments on the SAR ship detection dataset (SSDD) demonstrate that the compressed model reduces model size, parameter count, and FLOPs by over 90\% while maintaining comparable detection accuracy to the original YOLOv4 model.

\begin{table*}[h!]
\centering
\footnotesize
\begin{tabular}{|
>{\columncolor[HTML]{FFFFFF}}c |
>{\columncolor[HTML]{FFFFFF}}c |
>{\columncolor[HTML]{FFFFFF}}c 
>{\columncolor[HTML]{FFFFFF}}c |
>{\columncolor[HTML]{FFFFFF}}c 
>{\columncolor[HTML]{FFFFFF}}c |}
\hline
\cellcolor[HTML]{FFFFFF}{\color[HTML]{000000} } &
  \cellcolor[HTML]{FFFFFF}{\color[HTML]{000000} } &
  \multicolumn{2}{c|}{\cellcolor[HTML]{FFFFFF}{\color[HTML]{000000} INDIAN PINES}} &
  \multicolumn{2}{c|}{\cellcolor[HTML]{FFFFFF}{\color[HTML]{000000} PAVIA UNIVERSITY}} \\ \cline{3-6} 
\multirow{-2}{*}{\cellcolor[HTML]{FFFFFF}{\color[HTML]{000000} }} &
  \multirow{-2}{*}{\cellcolor[HTML]{FFFFFF}{\color[HTML]{000000} \# trainable parameters}} &
  \multicolumn{1}{c|}{\cellcolor[HTML]{FFFFFF}{\color[HTML]{000000} Disjoint}} &
  {\color[HTML]{000000} Random} &
  \multicolumn{1}{c|}{\cellcolor[HTML]{FFFFFF}{\color[HTML]{000000} Disjoint}} &
  {\color[HTML]{000000} Random} \\ \hline
{\color[HTML]{000000} RF} &
  {\color[HTML]{000000} -} &
  \multicolumn{1}{c|}{\cellcolor[HTML]{FFFFFF}{\color[HTML]{000000} 66.1}} &
  {\color[HTML]{000000} 79.4} &
  \multicolumn{1}{c|}{\cellcolor[HTML]{FFFFFF}{\color[HTML]{000000} 68.5}} &
  {\color[HTML]{000000} 88.8} \\ \hline
{\color[HTML]{000000} MLR} &
  {\color[HTML]{000000} -} &
  \multicolumn{1}{c|}{\cellcolor[HTML]{FFFFFF}{\color[HTML]{000000} 78.5}} &
  {\color[HTML]{000000} 83.5} &
  \multicolumn{1}{c|}{\cellcolor[HTML]{FFFFFF}{\color[HTML]{000000} 74.6}} &
  {\color[HTML]{000000} 90.5} \\ \hline
{\color[HTML]{000000} SVM} &
  {\color[HTML]{000000} -} &
  \multicolumn{1}{c|}{\cellcolor[HTML]{FFFFFF}{\color[HTML]{000000} 69.9}} &
  {\color[HTML]{000000} 81.0} &
  \multicolumn{1}{c|}{\cellcolor[HTML]{FFFFFF}{\color[HTML]{000000} 79.8}} &
  {\color[HTML]{000000} 95.8} \\ \hline
{\color[HTML]{000000} MLP} &
  {\color[HTML]{000000} 31,047} &
  \multicolumn{1}{c|}{\cellcolor[HTML]{FFFFFF}{\color[HTML]{000000} 82.4}} &
  {\color[HTML]{000000} 88.9} &
  \multicolumn{1}{c|}{\cellcolor[HTML]{FFFFFF}{\color[HTML]{000000} 82.5}} &
  {\color[HTML]{000000} 92.5} \\ \hline
{\color[HTML]{000000} LSTM} &
  {\color[HTML]{000000} 102,416} &
  \multicolumn{1}{c|}{\cellcolor[HTML]{FFFFFF}{\color[HTML]{000000} 80.7}} &
  {\color[HTML]{000000} 88.4} &
  \multicolumn{1}{c|}{\cellcolor[HTML]{FFFFFF}{\color[HTML]{000000} 78.1}} &
  {\color[HTML]{000000} 96.1} \\ \hline
{\color[HTML]{000000} CNN1D} &
  {\color[HTML]{000000} 72,416} &
  \multicolumn{1}{c|}{\cellcolor[HTML]{FFFFFF}{\color[HTML]{000000} 82.5}} &
  {\color[HTML]{000000} 89.4} &
  \multicolumn{1}{c|}{\cellcolor[HTML]{FFFFFF}{\color[HTML]{000000} 82.3}} &
  {\color[HTML]{000000} 93.5} \\ \hline
{\color[HTML]{000000} CNN2D   (c=1)} &
  {\color[HTML]{000000} 378,116} &
  \multicolumn{1}{c|}{\cellcolor[HTML]{FFFFFF}{\color[HTML]{000000} 55.4}} &
  {\color[HTML]{000000} 99.0} &
  \multicolumn{1}{c|}{\cellcolor[HTML]{FFFFFF}{\color[HTML]{000000} 70.5}} &
  {\color[HTML]{000000} 99.3} \\ \hline
{\color[HTML]{000000} CNN2D   (c=40)} &
  {\color[HTML]{000000} 426,866} &
  \multicolumn{1}{c|}{\cellcolor[HTML]{FFFFFF}{\color[HTML]{000000} \textbf{86.3}}} &
  {\color[HTML]{000000} \textbf{99.4}} &
  \multicolumn{1}{c|}{\cellcolor[HTML]{FFFFFF}{\color[HTML]{000000} \textbf{83.2}}} &
  {\color[HTML]{000000} \textbf{99.3}} \\ \hline
{\color[HTML]{000000} CNN3D} &
  {\color[HTML]{000000} 1.8M} &
  \multicolumn{1}{c|}{\cellcolor[HTML]{FFFFFF}{\color[HTML]{000000} 60.9}} &
  {\color[HTML]{000000} 99.3} &
  \multicolumn{1}{c|}{\cellcolor[HTML]{FFFFFF}{\color[HTML]{000000} 58.6}} &
  {\color[HTML]{000000} 99.1} \\ \hline
\end{tabular}
\caption{Top-1 accuracy for baseline models (c: number of channels after PCA)}
\label{tab:baselines}
\end{table*}

\begin{table*}[h!]
\centering
\footnotesize
\begin{tabular}{|
>{\columncolor[HTML]{FFFFFF}}c |
>{\columncolor[HTML]{FFFFFF}}c |
>{\columncolor[HTML]{FFFFFF}}c |
>{\columncolor[HTML]{FFFFFF}}c |
>{\columncolor[HTML]{FFFFFF}}c |
>{\columncolor[HTML]{FFFFFF}}c |
>{\columncolor[HTML]{FFFFFF}}c |}
\hline
{\color[HTML]{000000} } &
  {\color[HTML]{000000} Conv1} &
  {\color[HTML]{000000} Conv2} &
  {\color[HTML]{000000} fc1} &
  {\color[HTML]{000000} fc2} &
  {\color[HTML]{000000} Total} &
  {\color[HTML]{000000} Memory (MB)} \\ \hline
{\color[HTML]{000000} MLP} &
  {\color[HTML]{000000} -} &
  {\color[HTML]{000000} -} &
  {\color[HTML]{000000} -} &
  {\color[HTML]{000000} -} &
  {\color[HTML]{000000} 31,047} &
  {\color[HTML]{000000} 0.13} \\ \hline
{\color[HTML]{000000} CNN1D} &
  {\color[HTML]{000000} -} &
  {\color[HTML]{000000} -} &
  {\color[HTML]{000000} -} &
  {\color[HTML]{000000} -} &
  {\color[HTML]{000000} 72,416} &
  {\color[HTML]{000000} 0.29} \\ \hline
{\color[HTML]{000000} CNN2D} &
  {\color[HTML]{000000} 50,050} &
  {\color[HTML]{000000} 125,100} &
  {\color[HTML]{000000} 250,100} &
  {\color[HTML]{000000} 909} &
  {\color[HTML]{000000} 426,866} &
  {\color[HTML]{000000} 1.71} \\ \hline
{\color[HTML]{000000} Prune   (90\%)} &
  {\color[HTML]{000000} 15,015} &
  {\color[HTML]{000000} 11,280} &
  {\color[HTML]{000000} 22,530} &
  {\color[HTML]{000000} 496} &
  {\color[HTML]{000000} 49,321} &
  {\color[HTML]{000000} 0.60} \\ \hline
{\color[HTML]{000000} Prune   (95\%)} &
  {\color[HTML]{000000} 10,010} &
  {\color[HTML]{000000} 5,020} &
  {\color[HTML]{000000} 10,020} &
  {\color[HTML]{000000} 336} &
  {\color[HTML]{000000} 25,386} &
  {\color[HTML]{000000} 0.31} \\ \hline
{\color[HTML]{000000} Prune   (98\%)} &
  {\color[HTML]{000000} 5,005} &
  {\color[HTML]{000000} 1,260} &
  {\color[HTML]{000000} 2,510} &
  {\color[HTML]{000000} 176} &
  {\color[HTML]{000000} 8,951} &
  {\color[HTML]{000000} 0.12} \\ \hline
\end{tabular}
\caption{Number of parameters in each layer and memory consumption for original and pruned models}
\label{tab:layers}
\end{table*}

\section{Experiment and Result}

In this section, we conducted the aforementioned model compression approaches on land cover classification application using two representative dataset: Indian Pines and Pavia University. The compression techniques are applied on a CNN2D neural network with 90\%, 95\%, and 98\% compression ratio, respectively. The evaluation metrics reported are Top-1 and Top-5 accuracy. Top-5 accuracy is computed based on any of the model’s top 5 highest probability answers that match the expected answer. It considers a classification correct if any of the five predictions match the target label.

\subsection{Baselines}

Various baseline algorithms are implemented in this work, including Random Forest (RF), Multivariate Logistic Regression (MLR), Support Vector Machine (SVM), Multi-layer Perceptron (MLP), LSTM, CNN1D, CNN2D, and CNN3D. To make the results comparable, we keep the same data splitting ratio for both random and disjoint splitting. Specifically, IP has 55\% for train and 45\% for test, while UP has 7\% for train and 93\% for test. As is mentioned in Section 3.1, spectral models are implemented on RF, MLP, SVM, MLP, LSTM, and CNN1D, whereas spatial-spectral models are implemented on CNN2D and CNN3D. 

Table \ref{tab:baselines} shows the Top-1 accuracy of baseline models for both random and disjoint data splitting of Indian Pines (IP) and University of Pavia (UP) dataset. The results demonstrate that the classification accuracy using disjoint data splitting is worse than random splitting for all baseline algorithms, which indicates that it is a more challenging task than random data splitting. We are more interested in disjoint data splitting since it is a more realistic land cover classification scenario, as mentioned in Section 3.2. It can be seen that CNN2D, with 40 channels, achieves the best performance by utilizing both spectral and spatial information. If we reduce the number of channels to one, significant information loss will hurt the model performance. Additionally, CNN3D does not work well in this case because it assumes existing spatial correlations between different channels, which may not be valid after applying Principle Component Analysis (PCA). We focus on CNN2D (c=40) for all the model compression techniques discussed in later sections. 

\begin{table}[h!]
\centering
\footnotesize
\begin{tabular}{|
>{\columncolor[HTML]{FFFFFF}}c |
>{\columncolor[HTML]{FFFFFF}}c 
>{\columncolor[HTML]{FFFFFF}}c |
>{\columncolor[HTML]{FFFFFF}}c 
>{\columncolor[HTML]{FFFFFF}}c |}
\hline
{\color[HTML]{000000} } &
  \multicolumn{1}{c|}{\cellcolor[HTML]{FFFFFF}{\color[HTML]{000000} Conv1}} &
  {\color[HTML]{000000} Conv2} &
  \multicolumn{1}{c|}{\cellcolor[HTML]{FFFFFF}{\color[HTML]{000000} fc1}} &
  {\color[HTML]{000000} fc2} \\ \hline
{\color[HTML]{000000} } &
  \multicolumn{2}{c|}{\cellcolor[HTML]{FFFFFF}{\color[HTML]{000000} \#   of filters}} &
  \multicolumn{2}{c|}{\cellcolor[HTML]{FFFFFF}{\color[HTML]{000000} \#   of neurons}} \\ \hline
{\color[HTML]{000000} CNN2D} &
  \multicolumn{1}{c|}{\cellcolor[HTML]{FFFFFF}{\color[HTML]{000000} 50}} &
  {\color[HTML]{000000} 100} &
  \multicolumn{1}{c|}{\cellcolor[HTML]{FFFFFF}{\color[HTML]{000000} 2,500}} &
  {\color[HTML]{000000} 100} \\ \hline
{\color[HTML]{000000} Prune   (90\%)} &
  \multicolumn{1}{c|}{\cellcolor[HTML]{FFFFFF}{\color[HTML]{000000} 15}} &
  {\color[HTML]{000000} 30} &
  \multicolumn{1}{c|}{\cellcolor[HTML]{FFFFFF}{\color[HTML]{000000} 750}} &
  {\color[HTML]{000000} 30} \\ \hline
{\color[HTML]{000000} Prune   (95\%)} &
  \multicolumn{1}{c|}{\cellcolor[HTML]{FFFFFF}{\color[HTML]{000000} 10}} &
  {\color[HTML]{000000} 20} &
  \multicolumn{1}{c|}{\cellcolor[HTML]{FFFFFF}{\color[HTML]{000000} 500}} &
  {\color[HTML]{000000} 20} \\ \hline
{\color[HTML]{000000} Prune   (98\%)} &
  \multicolumn{1}{c|}{\cellcolor[HTML]{FFFFFF}{\color[HTML]{000000} 5}} &
  {\color[HTML]{000000} 10} &
  \multicolumn{1}{c|}{\cellcolor[HTML]{FFFFFF}{\color[HTML]{000000} 250}} &
  {\color[HTML]{000000} 10} \\ \hline
\end{tabular}
\caption{Number of remaining filters in Conv layers and neurons in FC layers after pruning}
\label{tab:filters}
\end{table}

\begin{table*}[!h]
\centering
\footnotesize
\begin{tabular}{|
>{\columncolor[HTML]{FFFFFF}}c 
>{\columncolor[HTML]{FFFFFF}}c |
>{\columncolor[HTML]{FFFFFF}}c 
>{\columncolor[HTML]{FFFFFF}}c 
>{\columncolor[HTML]{FFFFFF}}c 
>{\columncolor[HTML]{FFFFFF}}c |
>{\columncolor[HTML]{FFFFFF}}c 
>{\columncolor[HTML]{FFFFFF}}c 
>{\columncolor[HTML]{FFFFFF}}c 
>{\columncolor[HTML]{FFFFFF}}c |}
\hline
\multicolumn{2}{|c|}{\cellcolor[HTML]{FFFFFF}{\color[HTML]{000000} }} &
  \multicolumn{4}{c|}{\cellcolor[HTML]{FFFFFF}{\color[HTML]{000000} \textbf{INDIAN PINES}}} &
  \multicolumn{4}{c|}{\cellcolor[HTML]{FFFFFF}{\color[HTML]{000000} \textbf{PAVIA UNIVERSITY}}} \\ \cline{3-10} 
\multicolumn{2}{|c|}{\cellcolor[HTML]{FFFFFF}{\color[HTML]{000000} }} &
  \multicolumn{2}{c|}{\cellcolor[HTML]{FFFFFF}{\color[HTML]{000000} Disjoint}} &
  \multicolumn{2}{c|}{\cellcolor[HTML]{FFFFFF}{\color[HTML]{000000} Random}} &
  \multicolumn{2}{c|}{\cellcolor[HTML]{FFFFFF}{\color[HTML]{000000} Disjoint}} &
  \multicolumn{2}{c|}{\cellcolor[HTML]{FFFFFF}{\color[HTML]{000000} Random}} \\ \cline{3-10} 
\multicolumn{2}{|c|}{\multirow{-3}{*}{\cellcolor[HTML]{FFFFFF}{\color[HTML]{000000} \textbf{\begin{tabular}[c]{@{}c@{}}Methods\\    \\ (Pruning ratio: 90\%)\end{tabular}}}}} &
  \multicolumn{1}{c|}{\cellcolor[HTML]{FFFFFF}{\color[HTML]{000000} Top-1}} &
  \multicolumn{1}{c|}{\cellcolor[HTML]{FFFFFF}{\color[HTML]{000000} Top-5}} &
  \multicolumn{1}{c|}{\cellcolor[HTML]{FFFFFF}{\color[HTML]{000000} Top-1}} &
  {\color[HTML]{000000} Top-5} &
  \multicolumn{1}{c|}{\cellcolor[HTML]{FFFFFF}{\color[HTML]{000000} Top-1}} &
  \multicolumn{1}{l|}{\cellcolor[HTML]{FFFFFF}{\color[HTML]{000000} Top-5}} &
  \multicolumn{1}{c|}{\cellcolor[HTML]{FFFFFF}{\color[HTML]{000000} Top-1}} &
  {\color[HTML]{000000} Top-5} \\ \hline
\multicolumn{1}{|c|}{\cellcolor[HTML]{FFFFFF}{\color[HTML]{000000} }} &
  {\color[HTML]{000000} \textbf{MLP}} &
  \multicolumn{1}{c|}{\cellcolor[HTML]{FFFFFF}{\color[HTML]{000000} 82.4}} &
  \multicolumn{1}{c|}{\cellcolor[HTML]{FFFFFF}{\color[HTML]{000000} 98.6}} &
  \multicolumn{1}{c|}{\cellcolor[HTML]{FFFFFF}{\color[HTML]{000000} 88.9}} &
  {\color[HTML]{000000} 99.7} &
  \multicolumn{1}{c|}{\cellcolor[HTML]{FFFFFF}{\color[HTML]{000000} 82.5}} &
  \multicolumn{1}{c|}{\cellcolor[HTML]{FFFFFF}{\color[HTML]{000000} 99.5}} &
  \multicolumn{1}{c|}{\cellcolor[HTML]{FFFFFF}{\color[HTML]{000000} 92.5}} &
  {\color[HTML]{000000} 99.9} \\ \cline{2-10} 
\multicolumn{1}{|c|}{\cellcolor[HTML]{FFFFFF}{\color[HTML]{000000} }} &
  {\color[HTML]{000000} \textbf{CNN-1D}} &
  \multicolumn{1}{c|}{\cellcolor[HTML]{FFFFFF}{\color[HTML]{000000} 82.5}} &
  \multicolumn{1}{c|}{\cellcolor[HTML]{FFFFFF}{\color[HTML]{000000} 97.9}} &
  \multicolumn{1}{c|}{\cellcolor[HTML]{FFFFFF}{\color[HTML]{000000} 89.4}} &
  {\color[HTML]{000000} 99.5} &
  \multicolumn{1}{c|}{\cellcolor[HTML]{FFFFFF}{\color[HTML]{000000} 82.3}} &
  \multicolumn{1}{c|}{\cellcolor[HTML]{FFFFFF}{\color[HTML]{000000} 99.7}} &
  \multicolumn{1}{c|}{\cellcolor[HTML]{FFFFFF}{\color[HTML]{000000} 93.5}} &
  {\color[HTML]{000000} 99.9} \\ \cline{2-10} 
\multicolumn{1}{|c|}{\multirow{-3}{*}{\cellcolor[HTML]{FFFFFF}{\color[HTML]{000000} \textbf{Baselines}}}} &
  {\color[HTML]{000000} \textbf{CNN-2D}} &
  \multicolumn{1}{c|}{\cellcolor[HTML]{FFFFFF}{\color[HTML]{000000} 86.3}} &
  \multicolumn{1}{c|}{\cellcolor[HTML]{FFFFFF}{\color[HTML]{000000} 98.7}} &
  \multicolumn{1}{c|}{\cellcolor[HTML]{FFFFFF}{\color[HTML]{000000} 99.5}} &
  {\color[HTML]{000000} 99.9} &
  \multicolumn{1}{c|}{\cellcolor[HTML]{FFFFFF}{\color[HTML]{000000} 83.2}} &
  \multicolumn{1}{c|}{\cellcolor[HTML]{FFFFFF}{\color[HTML]{000000} 99.9}} &
  \multicolumn{1}{c|}{\cellcolor[HTML]{FFFFFF}{\color[HTML]{000000} 99.3}} &
  {\color[HTML]{000000} 100} \\ \hline
\multicolumn{1}{|c|}{\cellcolor[HTML]{FFFFFF}{\color[HTML]{000000} \textbf{Scratch}}} &
  {\color[HTML]{000000} \textbf{Pruned CNN2D}} &
  \multicolumn{1}{c|}{\cellcolor[HTML]{FFFFFF}{\color[HTML]{000000} 82.8}} &
  \multicolumn{1}{c|}{\cellcolor[HTML]{FFFFFF}{\color[HTML]{000000} 97.6}} &
  \multicolumn{1}{c|}{\cellcolor[HTML]{FFFFFF}{\color[HTML]{000000} 99.2}} &
  {\color[HTML]{000000} 99.9} &
  \multicolumn{1}{c|}{\cellcolor[HTML]{FFFFFF}{\color[HTML]{000000} 77.7}} &
  \multicolumn{1}{c|}{\cellcolor[HTML]{FFFFFF}{\color[HTML]{000000} 99.7}} &
  \multicolumn{1}{c|}{\cellcolor[HTML]{FFFFFF}{\color[HTML]{000000} 99.6}} &
  {\color[HTML]{000000} 99.9} \\ \hline
\multicolumn{1}{|c|}{\cellcolor[HTML]{FFFFFF}{\color[HTML]{000000} }} &
  {\color[HTML]{000000} \textbf{L1-norm pruning}} &
  \multicolumn{1}{c|}{\cellcolor[HTML]{FFFFFF}{\color[HTML]{000000} 84.6}} &
  \multicolumn{1}{c|}{\cellcolor[HTML]{FFFFFF}{\color[HTML]{000000} 98.1}} &
  \multicolumn{1}{c|}{\cellcolor[HTML]{FFFFFF}{\color[HTML]{000000} 99.3}} &
  {\color[HTML]{000000} 99.9} &
  \multicolumn{1}{c|}{\cellcolor[HTML]{FFFFFF}{\color[HTML]{000000} 82.0}} &
  \multicolumn{1}{c|}{\cellcolor[HTML]{FFFFFF}{\color[HTML]{000000} 99.6}} &
  \multicolumn{1}{c|}{\cellcolor[HTML]{FFFFFF}{\color[HTML]{000000} 99.4}} &
  {\color[HTML]{000000} 100} \\ \cline{2-10} 
\multicolumn{1}{|c|}{\cellcolor[HTML]{FFFFFF}{\color[HTML]{000000} }} &
  {\color[HTML]{000000} \textbf{ThiNet}} &
  \multicolumn{1}{c|}{\cellcolor[HTML]{FFFFFF}{\color[HTML]{000000} 82.5}} &
  \multicolumn{1}{c|}{\cellcolor[HTML]{FFFFFF}{\color[HTML]{000000} 97.9}} &
  \multicolumn{1}{c|}{\cellcolor[HTML]{FFFFFF}{\color[HTML]{000000} 98.2}} &
  {\color[HTML]{000000} 99.9} &
  \multicolumn{1}{c|}{\cellcolor[HTML]{FFFFFF}{\color[HTML]{000000} 80.4}} &
  \multicolumn{1}{c|}{\cellcolor[HTML]{FFFFFF}{\color[HTML]{000000} 99.5}} &
  \multicolumn{1}{c|}{\cellcolor[HTML]{FFFFFF}{\color[HTML]{000000} 99.3}} &
  {\color[HTML]{000000} 99.9} \\ \cline{2-10} 
\multicolumn{1}{|c|}{\cellcolor[HTML]{FFFFFF}{\color[HTML]{000000} }} &
  {\color[HTML]{000000} \textbf{Network Slimming}} &
  \multicolumn{1}{c|}{\cellcolor[HTML]{FFFFFF}{\color[HTML]{000000} 87.0}} &
  \multicolumn{1}{c|}{\cellcolor[HTML]{FFFFFF}{\color[HTML]{000000} 98.2}} &
  \multicolumn{1}{c|}{\cellcolor[HTML]{FFFFFF}{\color[HTML]{000000} 99.2}} &
  {\color[HTML]{000000} 99.9} &
  \multicolumn{1}{c|}{\cellcolor[HTML]{FFFFFF}{\color[HTML]{000000} 80.6}} &
  \multicolumn{1}{c|}{\cellcolor[HTML]{FFFFFF}{\color[HTML]{000000} 99.9}} &
  \multicolumn{1}{c|}{\cellcolor[HTML]{FFFFFF}{\color[HTML]{000000} 99.3}} &
  {\color[HTML]{000000} 99.9} \\ \cline{2-10} 
\multicolumn{1}{|c|}{\multirow{-4}{*}{\cellcolor[HTML]{FFFFFF}{\color[HTML]{000000} \textbf{\begin{tabular}[c]{@{}c@{}}Pruning \\    \\   Methods\end{tabular}}}}} &
  {\color[HTML]{000000} \textbf{SFP}} &
  \multicolumn{1}{c|}{\cellcolor[HTML]{FFFFFF}{\color[HTML]{000000} 84.5}} &
  \multicolumn{1}{c|}{\cellcolor[HTML]{FFFFFF}{\color[HTML]{000000} 98.1}} &
  \multicolumn{1}{c|}{\cellcolor[HTML]{FFFFFF}{\color[HTML]{000000} 99.5}} &
  {\color[HTML]{000000} 99.9} &
  \multicolumn{1}{c|}{\cellcolor[HTML]{FFFFFF}{\color[HTML]{000000} 77.6}} &
  \multicolumn{1}{c|}{\cellcolor[HTML]{FFFFFF}{\color[HTML]{000000} 96.2}} &
  \multicolumn{1}{c|}{\cellcolor[HTML]{FFFFFF}{\color[HTML]{000000} 99.3}} &
  {\color[HTML]{000000} 99.9} \\ \hline
\end{tabular}
\caption{Top-1 and Top-5 accuracy of model using pruning methods (90\% reduction)}
\label{tab:prune90}
\end{table*}

% \usepackage{multirow}
\begin{table*}[!h]
\centering
\footnotesize
\begin{tabular}{|
>{\columncolor[HTML]{FFFFFF}}c 
>{\columncolor[HTML]{FFFFFF}}c |
>{\columncolor[HTML]{FFFFFF}}c 
>{\columncolor[HTML]{FFFFFF}}c 
>{\columncolor[HTML]{FFFFFF}}c 
>{\columncolor[HTML]{FFFFFF}}c |
>{\columncolor[HTML]{FFFFFF}}c 
>{\columncolor[HTML]{FFFFFF}}c 
>{\columncolor[HTML]{FFFFFF}}c 
>{\columncolor[HTML]{FFFFFF}}c |}
\hline
\multicolumn{2}{|c|}{\cellcolor[HTML]{FFFFFF}{\color[HTML]{000000} }} &
  \multicolumn{4}{c|}{\cellcolor[HTML]{FFFFFF}{\color[HTML]{000000} \textbf{INDIAN PINES}}} &
  \multicolumn{4}{c|}{\cellcolor[HTML]{FFFFFF}{\color[HTML]{000000} \textbf{PAVIA UNIVERSITY}}} \\ \cline{3-10} 
\multicolumn{2}{|c|}{\cellcolor[HTML]{FFFFFF}{\color[HTML]{000000} }} &
  \multicolumn{2}{c|}{\cellcolor[HTML]{FFFFFF}{\color[HTML]{000000} Disjoint}} &
  \multicolumn{2}{c|}{\cellcolor[HTML]{FFFFFF}{\color[HTML]{000000} Random}} &
  \multicolumn{2}{c|}{\cellcolor[HTML]{FFFFFF}{\color[HTML]{000000} Disjoint}} &
  \multicolumn{2}{c|}{\cellcolor[HTML]{FFFFFF}{\color[HTML]{000000} Random}} \\ \cline{3-10} 
\multicolumn{2}{|c|}{\multirow{-3}{*}{\cellcolor[HTML]{FFFFFF}{\color[HTML]{000000} \textbf{\begin{tabular}[c]{@{}c@{}}Methods\\    \\ (Pruning ratio: 95\%)\end{tabular}}}}} &
  \multicolumn{1}{c|}{\cellcolor[HTML]{FFFFFF}{\color[HTML]{000000} Top-1}} &
  \multicolumn{1}{c|}{\cellcolor[HTML]{FFFFFF}{\color[HTML]{000000} Top-5}} &
  \multicolumn{1}{c|}{\cellcolor[HTML]{FFFFFF}{\color[HTML]{000000} Top-1}} &
  {\color[HTML]{000000} Top-5} &
  \multicolumn{1}{c|}{\cellcolor[HTML]{FFFFFF}{\color[HTML]{000000} Top-1}} &
  \multicolumn{1}{l|}{\cellcolor[HTML]{FFFFFF}{\color[HTML]{000000} Top-5}} &
  \multicolumn{1}{c|}{\cellcolor[HTML]{FFFFFF}{\color[HTML]{000000} Top-1}} &
  {\color[HTML]{000000} Top-5} \\ \hline
\multicolumn{1}{|c|}{\cellcolor[HTML]{FFFFFF}{\color[HTML]{000000} }} &
  {\color[HTML]{000000} \textbf{MLP}} &
  \multicolumn{1}{c|}{\cellcolor[HTML]{FFFFFF}{\color[HTML]{000000} 82.4}} &
  \multicolumn{1}{c|}{\cellcolor[HTML]{FFFFFF}{\color[HTML]{000000} 98.6}} &
  \multicolumn{1}{c|}{\cellcolor[HTML]{FFFFFF}{\color[HTML]{000000} 88.9}} &
  {\color[HTML]{000000} 99.7} &
  \multicolumn{1}{c|}{\cellcolor[HTML]{FFFFFF}{\color[HTML]{000000} 82.5}} &
  \multicolumn{1}{c|}{\cellcolor[HTML]{FFFFFF}{\color[HTML]{000000} 99.5}} &
  \multicolumn{1}{c|}{\cellcolor[HTML]{FFFFFF}{\color[HTML]{000000} 92.5}} &
  {\color[HTML]{000000} 99.9} \\ \cline{2-10} 
\multicolumn{1}{|c|}{\cellcolor[HTML]{FFFFFF}{\color[HTML]{000000} }} &
  {\color[HTML]{000000} \textbf{CNN-1D}} &
  \multicolumn{1}{c|}{\cellcolor[HTML]{FFFFFF}{\color[HTML]{000000} 82.5}} &
  \multicolumn{1}{c|}{\cellcolor[HTML]{FFFFFF}{\color[HTML]{000000} 97.9}} &
  \multicolumn{1}{c|}{\cellcolor[HTML]{FFFFFF}{\color[HTML]{000000} 89.4}} &
  {\color[HTML]{000000} 99.5} &
  \multicolumn{1}{c|}{\cellcolor[HTML]{FFFFFF}{\color[HTML]{000000} 82.3}} &
  \multicolumn{1}{c|}{\cellcolor[HTML]{FFFFFF}{\color[HTML]{000000} 99.7}} &
  \multicolumn{1}{c|}{\cellcolor[HTML]{FFFFFF}{\color[HTML]{000000} 93.5}} &
  {\color[HTML]{000000} 99.9} \\ \cline{2-10} 
\multicolumn{1}{|c|}{\multirow{-3}{*}{\cellcolor[HTML]{FFFFFF}{\color[HTML]{000000} \textbf{Baselines}}}} &
  {\color[HTML]{000000} \textbf{CNN-2D}} &
  \multicolumn{1}{c|}{\cellcolor[HTML]{FFFFFF}{\color[HTML]{000000} 86.3}} &
  \multicolumn{1}{c|}{\cellcolor[HTML]{FFFFFF}{\color[HTML]{000000} 98.7}} &
  \multicolumn{1}{c|}{\cellcolor[HTML]{FFFFFF}{\color[HTML]{000000} 99.4}} &
  {\color[HTML]{000000} 99.9} &
  \multicolumn{1}{c|}{\cellcolor[HTML]{FFFFFF}{\color[HTML]{000000} 83.2}} &
  \multicolumn{1}{c|}{\cellcolor[HTML]{FFFFFF}{\color[HTML]{000000} 99.9}} &
  \multicolumn{1}{c|}{\cellcolor[HTML]{FFFFFF}{\color[HTML]{000000} 99.3}} &
  {\color[HTML]{000000} 100} \\ \hline
\multicolumn{1}{|c|}{\cellcolor[HTML]{FFFFFF}{\color[HTML]{000000} \textbf{Scratch}}} &
  {\color[HTML]{000000} \textbf{Pruned CNN2D}} &
  \multicolumn{1}{c|}{\cellcolor[HTML]{FFFFFF}{\color[HTML]{000000} 82.1}} &
  \multicolumn{1}{c|}{\cellcolor[HTML]{FFFFFF}{\color[HTML]{000000} 97.1}} &
  \multicolumn{1}{c|}{\cellcolor[HTML]{FFFFFF}{\color[HTML]{000000} 99.3}} &
  {\color[HTML]{000000} 100} &
  \multicolumn{1}{c|}{\cellcolor[HTML]{FFFFFF}{\color[HTML]{000000} 74.1}} &
  \multicolumn{1}{c|}{\cellcolor[HTML]{FFFFFF}{\color[HTML]{000000} 99.6}} &
  \multicolumn{1}{c|}{\cellcolor[HTML]{FFFFFF}{\color[HTML]{000000} 99.0}} &
  {\color[HTML]{000000} 100} \\ \hline
\multicolumn{1}{|c|}{\cellcolor[HTML]{FFFFFF}{\color[HTML]{000000} }} &
  {\color[HTML]{000000} \textbf{L1-norm pruning}} &
  \multicolumn{1}{c|}{\cellcolor[HTML]{FFFFFF}{\color[HTML]{000000} 86.1}} &
  \multicolumn{1}{c|}{\cellcolor[HTML]{FFFFFF}{\color[HTML]{000000} 98.4}} &
  \multicolumn{1}{c|}{\cellcolor[HTML]{FFFFFF}{\color[HTML]{000000} 99.4}} &
  {\color[HTML]{000000} 99.9} &
  \multicolumn{1}{c|}{\cellcolor[HTML]{FFFFFF}{\color[HTML]{000000} 83.0}} &
  \multicolumn{1}{c|}{\cellcolor[HTML]{FFFFFF}{\color[HTML]{000000} 99.3}} &
  \multicolumn{1}{c|}{\cellcolor[HTML]{FFFFFF}{\color[HTML]{000000} 99.3}} &
  {\color[HTML]{000000} 100} \\ \cline{2-10} 
\multicolumn{1}{|c|}{\cellcolor[HTML]{FFFFFF}{\color[HTML]{000000} }} &
  {\color[HTML]{000000} \textbf{ThiNet}} &
  \multicolumn{1}{c|}{\cellcolor[HTML]{FFFFFF}{\color[HTML]{000000} 83.2}} &
  \multicolumn{1}{c|}{\cellcolor[HTML]{FFFFFF}{\color[HTML]{000000} 98.1}} &
  \multicolumn{1}{c|}{\cellcolor[HTML]{FFFFFF}{\color[HTML]{000000} 97.8}} &
  {\color[HTML]{000000} 99.9} &
  \multicolumn{1}{c|}{\cellcolor[HTML]{FFFFFF}{\color[HTML]{000000} 81.4}} &
  \multicolumn{1}{c|}{\cellcolor[HTML]{FFFFFF}{\color[HTML]{000000} 99.3}} &
  \multicolumn{1}{c|}{\cellcolor[HTML]{FFFFFF}{\color[HTML]{000000} 99.3}} &
  {\color[HTML]{000000} 99.9} \\ \cline{2-10} 
\multicolumn{1}{|c|}{\cellcolor[HTML]{FFFFFF}{\color[HTML]{000000} }} &
  {\color[HTML]{000000} \textbf{Network Slimming}} &
  \multicolumn{1}{c|}{\cellcolor[HTML]{FFFFFF}{\color[HTML]{000000} 79.8}} &
  \multicolumn{1}{c|}{\cellcolor[HTML]{FFFFFF}{\color[HTML]{000000} 95.6}} &
  \multicolumn{1}{c|}{\cellcolor[HTML]{FFFFFF}{\color[HTML]{000000} 99.2}} &
  {\color[HTML]{000000} 99.9} &
  \multicolumn{1}{c|}{\cellcolor[HTML]{FFFFFF}{\color[HTML]{000000} 84.2}} &
  \multicolumn{1}{c|}{\cellcolor[HTML]{FFFFFF}{\color[HTML]{000000} 99.9}} &
  \multicolumn{1}{c|}{\cellcolor[HTML]{FFFFFF}{\color[HTML]{000000} 98.9}} &
  {\color[HTML]{000000} 99.9} \\ \cline{2-10} 
\multicolumn{1}{|c|}{\multirow{-4}{*}{\cellcolor[HTML]{FFFFFF}{\color[HTML]{000000} \textbf{\begin{tabular}[c]{@{}c@{}}Pruning \\    \\   Methods\end{tabular}}}}} &
  {\color[HTML]{000000} \textbf{SFP}} &
  \multicolumn{1}{c|}{\cellcolor[HTML]{FFFFFF}{\color[HTML]{000000} 85.0}} &
  \multicolumn{1}{c|}{\cellcolor[HTML]{FFFFFF}{\color[HTML]{000000} 97.3}} &
  \multicolumn{1}{c|}{\cellcolor[HTML]{FFFFFF}{\color[HTML]{000000} 99.7}} &
  {\color[HTML]{000000} 99.9} &
  \multicolumn{1}{c|}{\cellcolor[HTML]{FFFFFF}{\color[HTML]{000000} 76.5}} &
  \multicolumn{1}{c|}{\cellcolor[HTML]{FFFFFF}{\color[HTML]{000000} 98.5}} &
  \multicolumn{1}{c|}{\cellcolor[HTML]{FFFFFF}{\color[HTML]{000000} 99.1}} &
  {\color[HTML]{000000} 99.9} \\ \hline
\end{tabular}
\caption{Top-1 and Top-5 accuracy of model using pruning methods (95\% reduction)}
\label{tab:prune95}
\end{table*}

\begin{table*}[!ht]
\centering
\footnotesize
\begin{tabular}{|
>{\columncolor[HTML]{FFFFFF}}c 
>{\columncolor[HTML]{FFFFFF}}c |
>{\columncolor[HTML]{FFFFFF}}c 
>{\columncolor[HTML]{FFFFFF}}c 
>{\columncolor[HTML]{FFFFFF}}c 
>{\columncolor[HTML]{FFFFFF}}c |
>{\columncolor[HTML]{FFFFFF}}c 
>{\columncolor[HTML]{FFFFFF}}c 
>{\columncolor[HTML]{FFFFFF}}c 
>{\columncolor[HTML]{FFFFFF}}c |}
\hline
\multicolumn{2}{|c|}{\cellcolor[HTML]{FFFFFF}{\color[HTML]{000000} }} &
  \multicolumn{4}{c|}{\cellcolor[HTML]{FFFFFF}{\color[HTML]{000000} \textbf{INDIAN PINES}}} &
  \multicolumn{4}{c|}{\cellcolor[HTML]{FFFFFF}{\color[HTML]{000000} \textbf{PAVIA UNIVERSITY}}} \\ \cline{3-10} 
\multicolumn{2}{|c|}{\cellcolor[HTML]{FFFFFF}{\color[HTML]{000000} }} &
  \multicolumn{2}{c|}{\cellcolor[HTML]{FFFFFF}{\color[HTML]{000000} Disjoint}} &
  \multicolumn{2}{c|}{\cellcolor[HTML]{FFFFFF}{\color[HTML]{000000} Random}} &
  \multicolumn{2}{c|}{\cellcolor[HTML]{FFFFFF}{\color[HTML]{000000} Disjoint}} &
  \multicolumn{2}{c|}{\cellcolor[HTML]{FFFFFF}{\color[HTML]{000000} Random}} \\ \cline{3-10} 
\multicolumn{2}{|c|}{\multirow{-3}{*}{\cellcolor[HTML]{FFFFFF}{\color[HTML]{000000} \textbf{\begin{tabular}[c]{@{}c@{}}Methods\\    \\ (Pruning ratio: 98\%)\end{tabular}}}}} &
  \multicolumn{1}{c|}{\cellcolor[HTML]{FFFFFF}{\color[HTML]{000000} Top-1}} &
  \multicolumn{1}{c|}{\cellcolor[HTML]{FFFFFF}{\color[HTML]{000000} Top-5}} &
  \multicolumn{1}{c|}{\cellcolor[HTML]{FFFFFF}{\color[HTML]{000000} Top-1}} &
  {\color[HTML]{000000} Top-5} &
  \multicolumn{1}{c|}{\cellcolor[HTML]{FFFFFF}{\color[HTML]{000000} Top-1}} &
  \multicolumn{1}{l|}{\cellcolor[HTML]{FFFFFF}{\color[HTML]{000000} Top-5}} &
  \multicolumn{1}{c|}{\cellcolor[HTML]{FFFFFF}{\color[HTML]{000000} Top-1}} &
  {\color[HTML]{000000} Top-5} \\ \hline
\multicolumn{1}{|c|}{\cellcolor[HTML]{FFFFFF}{\color[HTML]{000000} }} &
  {\color[HTML]{000000} \textbf{MLP}} &
  \multicolumn{1}{c|}{\cellcolor[HTML]{FFFFFF}{\color[HTML]{000000} 82.4}} &
  \multicolumn{1}{c|}{\cellcolor[HTML]{FFFFFF}{\color[HTML]{000000} 98.6}} &
  \multicolumn{1}{c|}{\cellcolor[HTML]{FFFFFF}{\color[HTML]{000000} 88.9}} &
  {\color[HTML]{000000} 99.7} &
  \multicolumn{1}{c|}{\cellcolor[HTML]{FFFFFF}{\color[HTML]{000000} 82.5}} &
  \multicolumn{1}{c|}{\cellcolor[HTML]{FFFFFF}{\color[HTML]{000000} 99.5}} &
  \multicolumn{1}{c|}{\cellcolor[HTML]{FFFFFF}{\color[HTML]{000000} 92.5}} &
  {\color[HTML]{000000} 99.9} \\ \cline{2-10} 
\multicolumn{1}{|c|}{\cellcolor[HTML]{FFFFFF}{\color[HTML]{000000} }} &
  {\color[HTML]{000000} \textbf{CNN-1D}} &
  \multicolumn{1}{c|}{\cellcolor[HTML]{FFFFFF}{\color[HTML]{000000} 82.5}} &
  \multicolumn{1}{c|}{\cellcolor[HTML]{FFFFFF}{\color[HTML]{000000} 97.9}} &
  \multicolumn{1}{c|}{\cellcolor[HTML]{FFFFFF}{\color[HTML]{000000} 89.4}} &
  {\color[HTML]{000000} 99.5} &
  \multicolumn{1}{c|}{\cellcolor[HTML]{FFFFFF}{\color[HTML]{000000} 82.3}} &
  \multicolumn{1}{c|}{\cellcolor[HTML]{FFFFFF}{\color[HTML]{000000} 99.7}} &
  \multicolumn{1}{c|}{\cellcolor[HTML]{FFFFFF}{\color[HTML]{000000} 93.5}} &
  {\color[HTML]{000000} 99.9} \\ \cline{2-10} 
\multicolumn{1}{|c|}{\multirow{-3}{*}{\cellcolor[HTML]{FFFFFF}{\color[HTML]{000000} \textbf{Baselines}}}} &
  {\color[HTML]{000000} \textbf{CNN-2D}} &
  \multicolumn{1}{c|}{\cellcolor[HTML]{FFFFFF}{\color[HTML]{000000} 86.3}} &
  \multicolumn{1}{c|}{\cellcolor[HTML]{FFFFFF}{\color[HTML]{000000} 98.7}} &
  \multicolumn{1}{c|}{\cellcolor[HTML]{FFFFFF}{\color[HTML]{000000} 99.4}} &
  {\color[HTML]{000000} 99.9} &
  \multicolumn{1}{c|}{\cellcolor[HTML]{FFFFFF}{\color[HTML]{000000} 83.2}} &
  \multicolumn{1}{c|}{\cellcolor[HTML]{FFFFFF}{\color[HTML]{000000} 99.9}} &
  \multicolumn{1}{c|}{\cellcolor[HTML]{FFFFFF}{\color[HTML]{000000} 99.3}} &
  {\color[HTML]{000000} 100} \\ \hline
\multicolumn{1}{|c|}{\cellcolor[HTML]{FFFFFF}{\color[HTML]{000000} \textbf{Scratch}}} &
  {\color[HTML]{000000} \textbf{Pruned CNN2D}} &
  \multicolumn{1}{c|}{\cellcolor[HTML]{FFFFFF}{\color[HTML]{000000} 78.2}} &
  \multicolumn{1}{c|}{\cellcolor[HTML]{FFFFFF}{\color[HTML]{000000} 96.4}} &
  \multicolumn{1}{c|}{\cellcolor[HTML]{FFFFFF}{\color[HTML]{000000} 98.9}} &
  {\color[HTML]{000000} 100} &
  \multicolumn{1}{c|}{\cellcolor[HTML]{FFFFFF}{\color[HTML]{000000} 77.2}} &
  \multicolumn{1}{c|}{\cellcolor[HTML]{FFFFFF}{\color[HTML]{000000} 99.9}} &
  \multicolumn{1}{c|}{\cellcolor[HTML]{FFFFFF}{\color[HTML]{000000} 98.1}} &
  {\color[HTML]{000000} 100} \\ \hline
\multicolumn{1}{|c|}{\cellcolor[HTML]{FFFFFF}{\color[HTML]{000000} }} &
  {\color[HTML]{000000} \textbf{L1-norm pruning}} &
  \multicolumn{1}{c|}{\cellcolor[HTML]{FFFFFF}{\color[HTML]{000000} 79.6}} &
  \multicolumn{1}{c|}{\cellcolor[HTML]{FFFFFF}{\color[HTML]{000000} 97.4}} &
  \multicolumn{1}{c|}{\cellcolor[HTML]{FFFFFF}{\color[HTML]{000000} 99.2}} &
  {\color[HTML]{000000} 99.9} &
  \multicolumn{1}{c|}{\cellcolor[HTML]{FFFFFF}{\color[HTML]{000000} 80.1}} &
  \multicolumn{1}{c|}{\cellcolor[HTML]{FFFFFF}{\color[HTML]{000000} 99.7}} &
  \multicolumn{1}{c|}{\cellcolor[HTML]{FFFFFF}{\color[HTML]{000000} 98.3}} &
  {\color[HTML]{000000} 99.9} \\ \cline{2-10} 
\multicolumn{1}{|c|}{\cellcolor[HTML]{FFFFFF}{\color[HTML]{000000} }} &
  {\color[HTML]{000000} \textbf{ThiNet}} &
  \multicolumn{1}{c|}{\cellcolor[HTML]{FFFFFF}{\color[HTML]{000000} 71.1}} &
  \multicolumn{1}{c|}{\cellcolor[HTML]{FFFFFF}{\color[HTML]{000000} 95.0}} &
  \multicolumn{1}{c|}{\cellcolor[HTML]{FFFFFF}{\color[HTML]{000000} 96.4}} &
  {\color[HTML]{000000} 99.7} &
  \multicolumn{1}{c|}{\cellcolor[HTML]{FFFFFF}{\color[HTML]{000000} 81.7}} &
  \multicolumn{1}{c|}{\cellcolor[HTML]{FFFFFF}{\color[HTML]{000000} 99.7}} &
  \multicolumn{1}{c|}{\cellcolor[HTML]{FFFFFF}{\color[HTML]{000000} 99.1}} &
  {\color[HTML]{000000} 99.9} \\ \cline{2-10} 
\multicolumn{1}{|c|}{\cellcolor[HTML]{FFFFFF}{\color[HTML]{000000} }} &
  {\color[HTML]{000000} \textbf{Network Slimming}} &
  \multicolumn{1}{c|}{\cellcolor[HTML]{FFFFFF}{\color[HTML]{000000} 81.3}} &
  \multicolumn{1}{c|}{\cellcolor[HTML]{FFFFFF}{\color[HTML]{000000} 95.6}} &
  \multicolumn{1}{c|}{\cellcolor[HTML]{FFFFFF}{\color[HTML]{000000} 99.1}} &
  {\color[HTML]{000000} 99.9} &
  \multicolumn{1}{c|}{\cellcolor[HTML]{FFFFFF}{\color[HTML]{000000} 75.7}} &
  \multicolumn{1}{c|}{\cellcolor[HTML]{FFFFFF}{\color[HTML]{000000} 99.8}} &
  \multicolumn{1}{c|}{\cellcolor[HTML]{FFFFFF}{\color[HTML]{000000} 98.6}} &
  {\color[HTML]{000000} 99.9} \\ \cline{2-10} 
\multicolumn{1}{|c|}{\multirow{-4}{*}{\cellcolor[HTML]{FFFFFF}{\color[HTML]{000000} \textbf{\begin{tabular}[c]{@{}c@{}}Pruning \\    \\   Methods\end{tabular}}}}} &
  {\color[HTML]{000000} \textbf{SFP}} &
  \multicolumn{1}{c|}{\cellcolor[HTML]{FFFFFF}{\color[HTML]{000000} 86.3}} &
  \multicolumn{1}{c|}{\cellcolor[HTML]{FFFFFF}{\color[HTML]{000000} 97.7}} &
  \multicolumn{1}{c|}{\cellcolor[HTML]{FFFFFF}{\color[HTML]{000000} 98.8}} &
  {\color[HTML]{000000} 99.9} &
  \multicolumn{1}{c|}{\cellcolor[HTML]{FFFFFF}{\color[HTML]{000000} 74.1}} &
  \multicolumn{1}{c|}{\cellcolor[HTML]{FFFFFF}{\color[HTML]{000000} 96.5}} &
  \multicolumn{1}{c|}{\cellcolor[HTML]{FFFFFF}{\color[HTML]{000000} 98.9}} &
  {\color[HTML]{000000} 99.9} \\ \hline
\end{tabular}
\caption{Top-1 and Top-5 accuracy of model using pruning methods (98\% reduction)}
\label{tab:prune98}
\end{table*}

\begin{table*}[!ht]
\centering
\footnotesize
\begin{tabular}{|
>{\columncolor[HTML]{FFFFFF}}c |
>{\columncolor[HTML]{FFFFFF}}c |
>{\columncolor[HTML]{FFFFFF}}c 
>{\columncolor[HTML]{FFFFFF}}c 
>{\columncolor[HTML]{FFFFFF}}c 
>{\columncolor[HTML]{FFFFFF}}c |
>{\columncolor[HTML]{FFFFFF}}c 
>{\columncolor[HTML]{FFFFFF}}c 
>{\columncolor[HTML]{FFFFFF}}c 
>{\columncolor[HTML]{FFFFFF}}c |}
\hline
\cellcolor[HTML]{FFFFFF}{\color[HTML]{000000} } &
  \cellcolor[HTML]{FFFFFF}{\color[HTML]{000000} } &
  \multicolumn{4}{c|}{\cellcolor[HTML]{FFFFFF}{\color[HTML]{000000} \textbf{INDIAN PINES}}} &
  \multicolumn{4}{c|}{\cellcolor[HTML]{FFFFFF}{\color[HTML]{000000} \textbf{PAVIA UNIVERSITY}}} \\ \cline{3-10} 
\cellcolor[HTML]{FFFFFF}{\color[HTML]{000000} } &
  \cellcolor[HTML]{FFFFFF}{\color[HTML]{000000} } &
  \multicolumn{2}{c|}{\cellcolor[HTML]{FFFFFF}{\color[HTML]{000000} Disjoint}} &
  \multicolumn{2}{c|}{\cellcolor[HTML]{FFFFFF}{\color[HTML]{000000} Random}} &
  \multicolumn{2}{c|}{\cellcolor[HTML]{FFFFFF}{\color[HTML]{000000} Disjoint}} &
  \multicolumn{2}{c|}{\cellcolor[HTML]{FFFFFF}{\color[HTML]{000000} Random}} \\ \cline{3-10} 
\multirow{-3}{*}{\cellcolor[HTML]{FFFFFF}{\color[HTML]{000000} \textbf{\begin{tabular}[c]{@{}c@{}}Pruning\\    \\ Ratio\end{tabular}}}} &
  \multirow{-3}{*}{\cellcolor[HTML]{FFFFFF}{\color[HTML]{000000} \textbf{\begin{tabular}[c]{@{}c@{}}Fine-tuning\\    \\ Strategy\end{tabular}}}} &
  \multicolumn{1}{c|}{\cellcolor[HTML]{FFFFFF}{\color[HTML]{000000} Top-1}} &
  \multicolumn{1}{c|}{\cellcolor[HTML]{FFFFFF}{\color[HTML]{000000} Top-5}} &
  \multicolumn{1}{c|}{\cellcolor[HTML]{FFFFFF}{\color[HTML]{000000} Top-1}} &
  {\color[HTML]{000000} Top-5} &
  \multicolumn{1}{c|}{\cellcolor[HTML]{FFFFFF}{\color[HTML]{000000} Top-1}} &
  \multicolumn{1}{l|}{\cellcolor[HTML]{FFFFFF}{\color[HTML]{000000} Top-5}} &
  \multicolumn{1}{c|}{\cellcolor[HTML]{FFFFFF}{\color[HTML]{000000} Top-1}} &
  {\color[HTML]{000000} Top-5} \\ \hline
\cellcolor[HTML]{FFFFFF}{\color[HTML]{000000} } &
  {\color[HTML]{000000} \textbf{I}} &
  \multicolumn{1}{c|}{\cellcolor[HTML]{FFFFFF}{\color[HTML]{000000} 84.6}} &
  \multicolumn{1}{c|}{\cellcolor[HTML]{FFFFFF}{\color[HTML]{000000} 98.1}} &
  \multicolumn{1}{c|}{\cellcolor[HTML]{FFFFFF}{\color[HTML]{000000} 99.3}} &
  {\color[HTML]{000000} 99.9} &
  \multicolumn{1}{c|}{\cellcolor[HTML]{FFFFFF}{\color[HTML]{000000} 82.0}} &
  \multicolumn{1}{c|}{\cellcolor[HTML]{FFFFFF}{\color[HTML]{000000} 99.6}} &
  \multicolumn{1}{c|}{\cellcolor[HTML]{FFFFFF}{\color[HTML]{000000} 99.4}} &
  {\color[HTML]{000000} 100} \\ \cline{2-10} 
\cellcolor[HTML]{FFFFFF}{\color[HTML]{000000} } &
  {\color[HTML]{000000} \textbf{II}} &
  \multicolumn{1}{c|}{\cellcolor[HTML]{FFFFFF}{\color[HTML]{000000} 86.1}} &
  \multicolumn{1}{c|}{\cellcolor[HTML]{FFFFFF}{\color[HTML]{000000} 97.5}} &
  \multicolumn{1}{c|}{\cellcolor[HTML]{FFFFFF}{\color[HTML]{000000} 99.6}} &
  {\color[HTML]{000000} 99.9} &
  \multicolumn{1}{c|}{\cellcolor[HTML]{FFFFFF}{\color[HTML]{000000} 81.7}} &
  \multicolumn{1}{c|}{\cellcolor[HTML]{FFFFFF}{\color[HTML]{000000} 99.7}} &
  \multicolumn{1}{c|}{\cellcolor[HTML]{FFFFFF}{\color[HTML]{000000} 99.5}} &
  {\color[HTML]{000000} 100} \\ \cline{2-10} 
\multirow{-3}{*}{\cellcolor[HTML]{FFFFFF}{\color[HTML]{000000} \textbf{90\%}}} &
  {\color[HTML]{000000} \textbf{III}} &
  \multicolumn{1}{c|}{\cellcolor[HTML]{FFFFFF}{\color[HTML]{000000} 81.5}} &
  \multicolumn{1}{c|}{\cellcolor[HTML]{FFFFFF}{\color[HTML]{000000} 97.9}} &
  \multicolumn{1}{c|}{\cellcolor[HTML]{FFFFFF}{\color[HTML]{000000} 99.2}} &
  {\color[HTML]{000000} 99.9} &
  \multicolumn{1}{c|}{\cellcolor[HTML]{FFFFFF}{\color[HTML]{000000} 81.3}} &
  \multicolumn{1}{c|}{\cellcolor[HTML]{FFFFFF}{\color[HTML]{000000} 98.9}} &
  \multicolumn{1}{c|}{\cellcolor[HTML]{FFFFFF}{\color[HTML]{000000} 98.7}} &
  {\color[HTML]{000000} 99.9} \\ \hline
\cellcolor[HTML]{FFFFFF}{\color[HTML]{000000} } &
  {\color[HTML]{000000} \textbf{I}} &
  \multicolumn{1}{c|}{\cellcolor[HTML]{FFFFFF}{\color[HTML]{000000} 86.1}} &
  \multicolumn{1}{c|}{\cellcolor[HTML]{FFFFFF}{\color[HTML]{000000} 98.4}} &
  \multicolumn{1}{c|}{\cellcolor[HTML]{FFFFFF}{\color[HTML]{000000} 99.4}} &
  {\color[HTML]{000000} 99.9} &
  \multicolumn{1}{c|}{\cellcolor[HTML]{FFFFFF}{\color[HTML]{000000} 83.0}} &
  \multicolumn{1}{c|}{\cellcolor[HTML]{FFFFFF}{\color[HTML]{000000} 99.3}} &
  \multicolumn{1}{c|}{\cellcolor[HTML]{FFFFFF}{\color[HTML]{000000} 99.3}} &
  {\color[HTML]{000000} 100} \\ \cline{2-10} 
\cellcolor[HTML]{FFFFFF}{\color[HTML]{000000} } &
  {\color[HTML]{000000} \textbf{II}} &
  \multicolumn{1}{c|}{\cellcolor[HTML]{FFFFFF}{\color[HTML]{000000} 87.3}} &
  \multicolumn{1}{c|}{\cellcolor[HTML]{FFFFFF}{\color[HTML]{000000} 97.7}} &
  \multicolumn{1}{c|}{\cellcolor[HTML]{FFFFFF}{\color[HTML]{000000} 99.4}} &
  {\color[HTML]{000000} 99.9} &
  \multicolumn{1}{c|}{\cellcolor[HTML]{FFFFFF}{\color[HTML]{000000} 82.1}} &
  \multicolumn{1}{c|}{\cellcolor[HTML]{FFFFFF}{\color[HTML]{000000} 99.0}} &
  \multicolumn{1}{c|}{\cellcolor[HTML]{FFFFFF}{\color[HTML]{000000} 99.4}} &
  {\color[HTML]{000000} 100} \\ \cline{2-10} 
\multirow{-3}{*}{\cellcolor[HTML]{FFFFFF}{\color[HTML]{000000} \textbf{95\%}}} &
  {\color[HTML]{000000} \textbf{III}} &
  \multicolumn{1}{c|}{\cellcolor[HTML]{FFFFFF}{\color[HTML]{000000} 84.9}} &
  \multicolumn{1}{c|}{\cellcolor[HTML]{FFFFFF}{\color[HTML]{000000} 97.3}} &
  \multicolumn{1}{c|}{\cellcolor[HTML]{FFFFFF}{\color[HTML]{000000} 99.5}} &
  {\color[HTML]{000000} 99.9} &
  \multicolumn{1}{c|}{\cellcolor[HTML]{FFFFFF}{\color[HTML]{000000} 83.9}} &
  \multicolumn{1}{c|}{\cellcolor[HTML]{FFFFFF}{\color[HTML]{000000} 99.8}} &
  \multicolumn{1}{c|}{\cellcolor[HTML]{FFFFFF}{\color[HTML]{000000} 99.6}} &
  {\color[HTML]{000000} 100} \\ \hline
\cellcolor[HTML]{FFFFFF}{\color[HTML]{000000} } &
  {\color[HTML]{000000} \textbf{I}} &
  \multicolumn{1}{c|}{\cellcolor[HTML]{FFFFFF}{\color[HTML]{000000} 79.6}} &
  \multicolumn{1}{c|}{\cellcolor[HTML]{FFFFFF}{\color[HTML]{000000} 97.4}} &
  \multicolumn{1}{c|}{\cellcolor[HTML]{FFFFFF}{\color[HTML]{000000} 99.2}} &
  {\color[HTML]{000000} 99.9} &
  \multicolumn{1}{c|}{\cellcolor[HTML]{FFFFFF}{\color[HTML]{000000} 80.1}} &
  \multicolumn{1}{c|}{\cellcolor[HTML]{FFFFFF}{\color[HTML]{000000} 99.7}} &
  \multicolumn{1}{c|}{\cellcolor[HTML]{FFFFFF}{\color[HTML]{000000} 98.3}} &
  {\color[HTML]{000000} 99.9} \\ \cline{2-10} 
\cellcolor[HTML]{FFFFFF}{\color[HTML]{000000} } &
  {\color[HTML]{000000} \textbf{II}} &
  \multicolumn{1}{c|}{\cellcolor[HTML]{FFFFFF}{\color[HTML]{000000} 84.6}} &
  \multicolumn{1}{c|}{\cellcolor[HTML]{FFFFFF}{\color[HTML]{000000} 96.3}} &
  \multicolumn{1}{c|}{\cellcolor[HTML]{FFFFFF}{\color[HTML]{000000} 99.2}} &
  {\color[HTML]{000000} 99.9} &
  \multicolumn{1}{c|}{\cellcolor[HTML]{FFFFFF}{\color[HTML]{000000} 80.8}} &
  \multicolumn{1}{c|}{\cellcolor[HTML]{FFFFFF}{\color[HTML]{000000} 96.6}} &
  \multicolumn{1}{c|}{\cellcolor[HTML]{FFFFFF}{\color[HTML]{000000} 99.1}} &
  {\color[HTML]{000000} 100} \\ \cline{2-10} 
\multirow{-3}{*}{\cellcolor[HTML]{FFFFFF}{\color[HTML]{000000} \textbf{98\%}}} &
  {\color[HTML]{000000} \textbf{III}} &
  \multicolumn{1}{c|}{\cellcolor[HTML]{FFFFFF}{\color[HTML]{000000} 81.5}} &
  \multicolumn{1}{c|}{\cellcolor[HTML]{FFFFFF}{\color[HTML]{000000} 97.9}} &
  \multicolumn{1}{c|}{\cellcolor[HTML]{FFFFFF}{\color[HTML]{000000} 99.2}} &
  {\color[HTML]{000000} 99.9} &
  \multicolumn{1}{c|}{\cellcolor[HTML]{FFFFFF}{\color[HTML]{000000} 81.3}} &
  \multicolumn{1}{c|}{\cellcolor[HTML]{FFFFFF}{\color[HTML]{000000} 98.9}} &
  \multicolumn{1}{c|}{\cellcolor[HTML]{FFFFFF}{\color[HTML]{000000} 98.7}} &
  {\color[HTML]{000000} 99.9} \\ \hline
\end{tabular}
\caption{Different fine-tuning strategy using L1-norm pruning method (Strategy I: prune once and retrain; Strategy II: prune and retrain iteratively; Strategy III: Multi-pass scheme)}
\label{tab:strategy}
\end{table*}

\begin{table*}[h!]
\centering
\footnotesize
\begin{tabular}{|
>{\columncolor[HTML]{FFFFFF}}c |
>{\columncolor[HTML]{FFFFFF}}c |
>{\columncolor[HTML]{FFFFFF}}c |
>{\columncolor[HTML]{FFFFFF}}c 
>{\columncolor[HTML]{FFFFFF}}c |
>{\columncolor[HTML]{FFFFFF}}c 
>{\columncolor[HTML]{FFFFFF}}c |}
\hline
\cellcolor[HTML]{FFFFFF}{\color[HTML]{000000} } &
  \cellcolor[HTML]{FFFFFF}{\color[HTML]{000000} } &
  \cellcolor[HTML]{FFFFFF}{\color[HTML]{000000} } &
  \multicolumn{2}{c|}{\cellcolor[HTML]{FFFFFF}{\color[HTML]{000000} \textbf{INDIAN PINES}}} &
  \multicolumn{2}{c|}{\cellcolor[HTML]{FFFFFF}{\color[HTML]{000000} \textbf{PAVIA UNIVERSITY}}} \\ \cline{4-7} 
\multirow{-2}{*}{\cellcolor[HTML]{FFFFFF}{\color[HTML]{000000} }} &
  \multirow{-2}{*}{\cellcolor[HTML]{FFFFFF}{\color[HTML]{000000} \textbf{\begin{tabular}[c]{@{}c@{}}Memory\\   (MB)\end{tabular}}}} &
  \multirow{-2}{*}{\cellcolor[HTML]{FFFFFF}{\color[HTML]{000000} \textbf{\begin{tabular}[c]{@{}c@{}}Inference Latency \\  (ms/sample)\end{tabular}}}} &
  \multicolumn{1}{c|}{\cellcolor[HTML]{FFFFFF}{\color[HTML]{000000} Disjoint}} &
  {\color[HTML]{000000} Random} &
  \multicolumn{1}{c|}{\cellcolor[HTML]{FFFFFF}{\color[HTML]{000000} Disjoint}} &
  {\color[HTML]{000000} Random} \\ \hline
{\color[HTML]{000000} \textbf{MLP}} &
  {\color[HTML]{000000} 0.13} &
  {\color[HTML]{000000} 0.34} &
  \multicolumn{1}{c|}{\cellcolor[HTML]{FFFFFF}{\color[HTML]{000000} 82.4}} &
  {\color[HTML]{000000} 88.9} &
  \multicolumn{1}{c|}{\cellcolor[HTML]{FFFFFF}{\color[HTML]{000000} 82.5}} &
  {\color[HTML]{000000} 92.5} \\ \hline
{\color[HTML]{000000} \textbf{CNN1D}} &
  {\color[HTML]{000000} 0.29} &
  {\color[HTML]{000000} 0.62} &
  \multicolumn{1}{c|}{\cellcolor[HTML]{FFFFFF}{\color[HTML]{000000} 82.5}} &
  {\color[HTML]{000000} 89.4} &
  \multicolumn{1}{c|}{\cellcolor[HTML]{FFFFFF}{\color[HTML]{000000} 82.3}} &
  {\color[HTML]{000000} 93.5} \\ \hline
{\color[HTML]{000000} \textbf{CNN2D}} &
  {\color[HTML]{000000} 1.71} &
  {\color[HTML]{000000} 3.28} &
  \multicolumn{1}{c|}{\cellcolor[HTML]{FFFFFF}{\color[HTML]{000000} 86.3}} &
  {\color[HTML]{000000} 99.4} &
  \multicolumn{1}{c|}{\cellcolor[HTML]{FFFFFF}{\color[HTML]{000000} 83.2}} &
  {\color[HTML]{000000} 99.3} \\ \hline
{\color[HTML]{000000} \textbf{Dynamic Quantization}} &
  {\color[HTML]{000000} 0.96} &
  {\color[HTML]{000000} 2.58} &
  \multicolumn{1}{c|}{\cellcolor[HTML]{FFFFFF}{\color[HTML]{000000} 86.5}} &
  {\color[HTML]{000000} 99.4} &
  \multicolumn{1}{c|}{\cellcolor[HTML]{FFFFFF}{\color[HTML]{000000} 83.2}} &
  {\color[HTML]{000000} 99.6} \\ \hline
{\color[HTML]{000000} \textbf{Static Quantization}} &
  {\color[HTML]{000000} 0.44} &
  {\color[HTML]{000000} 0.70} &
  \multicolumn{1}{c|}{\cellcolor[HTML]{FFFFFF}{\color[HTML]{000000} 86.7}} &
  {\color[HTML]{000000} 99.4} &
  \multicolumn{1}{c|}{\cellcolor[HTML]{FFFFFF}{\color[HTML]{000000} 83.3}} &
  {\color[HTML]{000000} 99.6} \\ \hline
{\color[HTML]{000000} \textbf{Quantization Aware Training}} &
  {\color[HTML]{000000} 0.44} &
  {\color[HTML]{000000} 1.00} &
  \multicolumn{1}{c|}{\cellcolor[HTML]{FFFFFF}{\color[HTML]{000000} 87.5}} &
  {\color[HTML]{000000} 99.5} &
  \multicolumn{1}{c|}{\cellcolor[HTML]{FFFFFF}{\color[HTML]{000000} 84.1}} &
  {\color[HTML]{000000} 99.6} \\ \hline
\end{tabular}
\caption{Neural network performance using different quantization modes (converted from float32 into int8)}
\label{tab:qresults}
\end{table*}

\begin{table*}[h!]
\centering
\footnotesize
\begin{tabular}{|
>{\columncolor[HTML]{FFFFFF}}c 
>{\columncolor[HTML]{FFFFFF}}c |
>{\columncolor[HTML]{FFFFFF}}c 
>{\columncolor[HTML]{FFFFFF}}c 
>{\columncolor[HTML]{FFFFFF}}c 
>{\columncolor[HTML]{FFFFFF}}c |
>{\columncolor[HTML]{FFFFFF}}c 
>{\columncolor[HTML]{FFFFFF}}c 
>{\columncolor[HTML]{FFFFFF}}c 
>{\columncolor[HTML]{FFFFFF}}c |}
\hline
\multicolumn{2}{|c|}{\cellcolor[HTML]{FFFFFF}{\color[HTML]{000000} }} &
  \multicolumn{4}{c|}{\cellcolor[HTML]{FFFFFF}{\color[HTML]{000000} \textbf{INDIAN PINES}}} &
  \multicolumn{4}{c|}{\cellcolor[HTML]{FFFFFF}{\color[HTML]{000000} \textbf{PAVIA UNIVERSITY}}} \\ \cline{3-10} 
\multicolumn{2}{|c|}{\cellcolor[HTML]{FFFFFF}{\color[HTML]{000000} }} &
  \multicolumn{2}{c|}{\cellcolor[HTML]{FFFFFF}{\color[HTML]{000000} Disjoint}} &
  \multicolumn{2}{c|}{\cellcolor[HTML]{FFFFFF}{\color[HTML]{000000} Random}} &
  \multicolumn{2}{c|}{\cellcolor[HTML]{FFFFFF}{\color[HTML]{000000} Disjoint}} &
  \multicolumn{2}{c|}{\cellcolor[HTML]{FFFFFF}{\color[HTML]{000000} Random}} \\ \cline{3-10} 
\multicolumn{2}{|c|}{\multirow{-3}{*}{\cellcolor[HTML]{FFFFFF}{\color[HTML]{000000} \textbf{\begin{tabular}[c]{@{}c@{}}Methods\\    \\ (student network: 90\%)\end{tabular}}}}} &
  \multicolumn{1}{c|}{\cellcolor[HTML]{FFFFFF}{\color[HTML]{000000} Top-1}} &
  \multicolumn{1}{c|}{\cellcolor[HTML]{FFFFFF}{\color[HTML]{000000} Top-5}} &
  \multicolumn{1}{c|}{\cellcolor[HTML]{FFFFFF}{\color[HTML]{000000} Top-1}} &
  {\color[HTML]{000000} Top-5} &
  \multicolumn{1}{c|}{\cellcolor[HTML]{FFFFFF}{\color[HTML]{000000} Top-1}} &
  \multicolumn{1}{l|}{\cellcolor[HTML]{FFFFFF}{\color[HTML]{000000} Top-5}} &
  \multicolumn{1}{c|}{\cellcolor[HTML]{FFFFFF}{\color[HTML]{000000} Top-1}} &
  {\color[HTML]{000000} Top-5} \\ \hline
\multicolumn{1}{|c|}{\cellcolor[HTML]{FFFFFF}{\color[HTML]{000000} }} &
  {\color[HTML]{000000} \textbf{MLP}} &
  \multicolumn{1}{c|}{\cellcolor[HTML]{FFFFFF}{\color[HTML]{000000} 82.4}} &
  \multicolumn{1}{c|}{\cellcolor[HTML]{FFFFFF}{\color[HTML]{000000} 98.6}} &
  \multicolumn{1}{c|}{\cellcolor[HTML]{FFFFFF}{\color[HTML]{000000} 88.9}} &
  {\color[HTML]{000000} 99.7} &
  \multicolumn{1}{c|}{\cellcolor[HTML]{FFFFFF}{\color[HTML]{000000} 82.5}} &
  \multicolumn{1}{c|}{\cellcolor[HTML]{FFFFFF}{\color[HTML]{000000} 99.5}} &
  \multicolumn{1}{c|}{\cellcolor[HTML]{FFFFFF}{\color[HTML]{000000} 92.5}} &
  {\color[HTML]{000000} 99.9} \\ \cline{2-10} 
\multicolumn{1}{|c|}{\cellcolor[HTML]{FFFFFF}{\color[HTML]{000000} }} &
  {\color[HTML]{000000} \textbf{CNN-1D}} &
  \multicolumn{1}{c|}{\cellcolor[HTML]{FFFFFF}{\color[HTML]{000000} 82.5}} &
  \multicolumn{1}{c|}{\cellcolor[HTML]{FFFFFF}{\color[HTML]{000000} 97.9}} &
  \multicolumn{1}{c|}{\cellcolor[HTML]{FFFFFF}{\color[HTML]{000000} 89.4}} &
  {\color[HTML]{000000} 99.5} &
  \multicolumn{1}{c|}{\cellcolor[HTML]{FFFFFF}{\color[HTML]{000000} 82.3}} &
  \multicolumn{1}{c|}{\cellcolor[HTML]{FFFFFF}{\color[HTML]{000000} 99.7}} &
  \multicolumn{1}{c|}{\cellcolor[HTML]{FFFFFF}{\color[HTML]{000000} 93.5}} &
  {\color[HTML]{000000} 99.9} \\ \cline{2-10} 
\multicolumn{1}{|c|}{\multirow{-3}{*}{\cellcolor[HTML]{FFFFFF}{\color[HTML]{000000} \textbf{Baselines}}}} &
  {\color[HTML]{000000} \textbf{CNN-2D}} &
  \multicolumn{1}{c|}{\cellcolor[HTML]{FFFFFF}{\color[HTML]{000000} 86.3}} &
  \multicolumn{1}{c|}{\cellcolor[HTML]{FFFFFF}{\color[HTML]{000000} 98.7}} &
  \multicolumn{1}{c|}{\cellcolor[HTML]{FFFFFF}{\color[HTML]{000000} 99.5}} &
  {\color[HTML]{000000} 99.9} &
  \multicolumn{1}{c|}{\cellcolor[HTML]{FFFFFF}{\color[HTML]{000000} 83.2}} &
  \multicolumn{1}{c|}{\cellcolor[HTML]{FFFFFF}{\color[HTML]{000000} 99.9}} &
  \multicolumn{1}{c|}{\cellcolor[HTML]{FFFFFF}{\color[HTML]{000000} 99.3}} &
  {\color[HTML]{000000} 100} \\ \hline
\multicolumn{1}{|c|}{\cellcolor[HTML]{FFFFFF}{\color[HTML]{000000} \textbf{Scratch}}} &
  {\color[HTML]{000000} \textbf{Pruned CNN2D}} &
  \multicolumn{1}{c|}{\cellcolor[HTML]{FFFFFF}{\color[HTML]{000000} 82.8}} &
  \multicolumn{1}{c|}{\cellcolor[HTML]{FFFFFF}{\color[HTML]{000000} 97.6}} &
  \multicolumn{1}{c|}{\cellcolor[HTML]{FFFFFF}{\color[HTML]{000000} 99.2}} &
  {\color[HTML]{000000} 99.9} &
  \multicolumn{1}{c|}{\cellcolor[HTML]{FFFFFF}{\color[HTML]{000000} 77.7}} &
  \multicolumn{1}{c|}{\cellcolor[HTML]{FFFFFF}{\color[HTML]{000000} 99.7}} &
  \multicolumn{1}{c|}{\cellcolor[HTML]{FFFFFF}{\color[HTML]{000000} 99.6}} &
  {\color[HTML]{000000} 99.9} \\ \hline
\multicolumn{1}{|c|}{\cellcolor[HTML]{FFFFFF}{\color[HTML]{000000} }} &
  {\color[HTML]{000000} \textbf{Soft targets}} &
  \multicolumn{1}{c|}{\cellcolor[HTML]{FFFFFF}{\color[HTML]{000000} 83.0}} &
  \multicolumn{1}{c|}{\cellcolor[HTML]{FFFFFF}{\color[HTML]{000000} 97.8}} &
  \multicolumn{1}{c|}{\cellcolor[HTML]{FFFFFF}{\color[HTML]{000000} 99.4}} &
  {\color[HTML]{000000} 99.9} &
  \multicolumn{1}{c|}{\cellcolor[HTML]{FFFFFF}{\color[HTML]{000000} 80.1}} &
  \multicolumn{1}{c|}{\cellcolor[HTML]{FFFFFF}{\color[HTML]{000000} 99.9}} &
  \multicolumn{1}{c|}{\cellcolor[HTML]{FFFFFF}{\color[HTML]{000000} 99.1}} &
  {\color[HTML]{000000} 100} \\ \cline{2-10} 
\multicolumn{1}{|c|}{\cellcolor[HTML]{FFFFFF}{\color[HTML]{000000} }} &
  {\color[HTML]{000000} \textbf{FitNets}} &
  \multicolumn{1}{c|}{\cellcolor[HTML]{FFFFFF}{\color[HTML]{000000} 86.3}} &
  \multicolumn{1}{c|}{\cellcolor[HTML]{FFFFFF}{\color[HTML]{000000} 98.8}} &
  \multicolumn{1}{c|}{\cellcolor[HTML]{FFFFFF}{\color[HTML]{000000} 99.4}} &
  {\color[HTML]{000000} 100} &
  \multicolumn{1}{c|}{\cellcolor[HTML]{FFFFFF}{\color[HTML]{000000} 81.4}} &
  \multicolumn{1}{c|}{\cellcolor[HTML]{FFFFFF}{\color[HTML]{000000} 99.9}} &
  \multicolumn{1}{c|}{\cellcolor[HTML]{FFFFFF}{\color[HTML]{000000} 99.2}} &
  {\color[HTML]{000000} 100} \\ \cline{2-10} 
\multicolumn{1}{|c|}{\cellcolor[HTML]{FFFFFF}{\color[HTML]{000000} }} &
  {\color[HTML]{000000} \textbf{Attention Transfer}} &
  \multicolumn{1}{c|}{\cellcolor[HTML]{FFFFFF}{\color[HTML]{000000} 87.1}} &
  \multicolumn{1}{c|}{\cellcolor[HTML]{FFFFFF}{\color[HTML]{000000} 99.2}} &
  \multicolumn{1}{c|}{\cellcolor[HTML]{FFFFFF}{\color[HTML]{000000} 99.3}} &
  {\color[HTML]{000000} 100} &
  \multicolumn{1}{c|}{\cellcolor[HTML]{FFFFFF}{\color[HTML]{000000} 81.5}} &
  \multicolumn{1}{c|}{\cellcolor[HTML]{FFFFFF}{\color[HTML]{000000} 99.9}} &
  \multicolumn{1}{c|}{\cellcolor[HTML]{FFFFFF}{\color[HTML]{000000} 99.4}} &
  {\color[HTML]{000000} 100} \\ \cline{2-10} 
\multicolumn{1}{|c|}{\cellcolor[HTML]{FFFFFF}{\color[HTML]{000000} }} &
  {\color[HTML]{000000} \textbf{Correlation Congruence}} &
  \multicolumn{1}{c|}{\cellcolor[HTML]{FFFFFF}{\color[HTML]{000000} 85.9}} &
  \multicolumn{1}{c|}{\cellcolor[HTML]{FFFFFF}{\color[HTML]{000000} 98.8}} &
  \multicolumn{1}{c|}{\cellcolor[HTML]{FFFFFF}{\color[HTML]{000000} 99.5}} &
  {\color[HTML]{000000} 100} &
  \multicolumn{1}{c|}{\cellcolor[HTML]{FFFFFF}{\color[HTML]{000000} 79.8}} &
  \multicolumn{1}{c|}{\cellcolor[HTML]{FFFFFF}{\color[HTML]{000000} 99.9}} &
  \multicolumn{1}{c|}{\cellcolor[HTML]{FFFFFF}{\color[HTML]{000000} 99.4}} &
  {\color[HTML]{000000} 100} \\ \cline{2-10} 
\multicolumn{1}{|c|}{\cellcolor[HTML]{FFFFFF}{\color[HTML]{000000} }} &
  {\color[HTML]{000000} \textbf{SimKD}} &
  \multicolumn{1}{c|}{\cellcolor[HTML]{FFFFFF}{\color[HTML]{000000} 86.1}} &
  \multicolumn{1}{c|}{\cellcolor[HTML]{FFFFFF}{\color[HTML]{000000} 99.0}} &
  \multicolumn{1}{c|}{\cellcolor[HTML]{FFFFFF}{\color[HTML]{000000} 99.5}} &
  {\color[HTML]{000000} 100} &
  \multicolumn{1}{c|}{\cellcolor[HTML]{FFFFFF}{\color[HTML]{000000} 80.9}} &
  \multicolumn{1}{c|}{\cellcolor[HTML]{FFFFFF}{\color[HTML]{000000} 99.9}} &
  \multicolumn{1}{c|}{\cellcolor[HTML]{FFFFFF}{\color[HTML]{000000} 99.3}} &
  {\color[HTML]{000000} 99.9} \\ \cline{2-10} 
\multicolumn{1}{|c|}{\multirow{-6}{*}{\cellcolor[HTML]{FFFFFF}{\color[HTML]{000000} \textbf{\begin{tabular}[c]{@{}c@{}}Offline \\    Distillation\end{tabular}}}}} &
  {\color[HTML]{000000} \textbf{CA-MKD}} &
  \multicolumn{1}{c|}{\cellcolor[HTML]{FFFFFF}{\color[HTML]{000000} 86.4}} &
  \multicolumn{1}{c|}{\cellcolor[HTML]{FFFFFF}{\color[HTML]{000000} 97.8}} &
  \multicolumn{1}{c|}{\cellcolor[HTML]{FFFFFF}{\color[HTML]{000000} 99.5}} &
  {\color[HTML]{000000} 100} &
  \multicolumn{1}{c|}{\cellcolor[HTML]{FFFFFF}{\color[HTML]{000000} 80.0}} &
  \multicolumn{1}{c|}{\cellcolor[HTML]{FFFFFF}{\color[HTML]{000000} 99.9}} &
  \multicolumn{1}{c|}{\cellcolor[HTML]{FFFFFF}{\color[HTML]{000000} 99.3}} &
  {\color[HTML]{000000} 99.9} \\ \hline
\multicolumn{1}{|c|}{\cellcolor[HTML]{FFFFFF}{\color[HTML]{000000} }} &
  {\color[HTML]{000000} \textbf{DML}} &
  \multicolumn{1}{c|}{\cellcolor[HTML]{FFFFFF}{\color[HTML]{000000} 87.0}} &
  \multicolumn{1}{c|}{\cellcolor[HTML]{FFFFFF}{\color[HTML]{000000} 97.5}} &
  \multicolumn{1}{c|}{\cellcolor[HTML]{FFFFFF}{\color[HTML]{000000} 99.3}} &
  {\color[HTML]{000000} 100} &
  \multicolumn{1}{c|}{\cellcolor[HTML]{FFFFFF}{\color[HTML]{000000} 83.1}} &
  \multicolumn{1}{c|}{\cellcolor[HTML]{FFFFFF}{\color[HTML]{000000} 99.9}} &
  \multicolumn{1}{c|}{\cellcolor[HTML]{FFFFFF}{\color[HTML]{000000} 99.2}} &
  {\color[HTML]{000000} 100} \\ \cline{2-10} 
\multicolumn{1}{|c|}{\cellcolor[HTML]{FFFFFF}{\color[HTML]{000000} }} &
  {\color[HTML]{000000} \textbf{ONE}} &
  \multicolumn{1}{c|}{\cellcolor[HTML]{FFFFFF}{\color[HTML]{000000} 87.4}} &
  \multicolumn{1}{c|}{\cellcolor[HTML]{FFFFFF}{\color[HTML]{000000} 97.8}} &
  \multicolumn{1}{c|}{\cellcolor[HTML]{FFFFFF}{\color[HTML]{000000} 99.2}} &
  {\color[HTML]{000000} 100} &
  \multicolumn{1}{c|}{\cellcolor[HTML]{FFFFFF}{\color[HTML]{000000} 81.6}} &
  \multicolumn{1}{c|}{\cellcolor[HTML]{FFFFFF}{\color[HTML]{000000} 99.9}} &
  \multicolumn{1}{c|}{\cellcolor[HTML]{FFFFFF}{\color[HTML]{000000} 99.5}} &
  {\color[HTML]{000000} 100} \\ \cline{2-10} 
\multicolumn{1}{|c|}{\cellcolor[HTML]{FFFFFF}{\color[HTML]{000000} }} &
  {\color[HTML]{000000} \textbf{CL-ILR}} &
  \multicolumn{1}{c|}{\cellcolor[HTML]{FFFFFF}{\color[HTML]{000000} 84.2}} &
  \multicolumn{1}{c|}{\cellcolor[HTML]{FFFFFF}{\color[HTML]{000000} 97.2}} &
  \multicolumn{1}{c|}{\cellcolor[HTML]{FFFFFF}{\color[HTML]{000000} 99.4}} &
  {\color[HTML]{000000} 100} &
  \multicolumn{1}{c|}{\cellcolor[HTML]{FFFFFF}{\color[HTML]{000000} 81.1}} &
  \multicolumn{1}{c|}{\cellcolor[HTML]{FFFFFF}{\color[HTML]{000000} 96.7}} &
  \multicolumn{1}{c|}{\cellcolor[HTML]{FFFFFF}{\color[HTML]{000000} 99.6}} &
  {\color[HTML]{000000} 100} \\ \cline{2-10} 
\multicolumn{1}{|c|}{\multirow{-4}{*}{\cellcolor[HTML]{FFFFFF}{\color[HTML]{000000} \textbf{\begin{tabular}[c]{@{}c@{}}Online\\     Distillation\end{tabular}}}}} &
  {\color[HTML]{000000} \textbf{OKDDip}} &
  \multicolumn{1}{c|}{\cellcolor[HTML]{FFFFFF}{\color[HTML]{000000} 86.4}} &
  \multicolumn{1}{c|}{\cellcolor[HTML]{FFFFFF}{\color[HTML]{000000} 97.9}} &
  \multicolumn{1}{c|}{\cellcolor[HTML]{FFFFFF}{\color[HTML]{000000} 99.5}} &
  {\color[HTML]{000000} 100} &
  \multicolumn{1}{c|}{\cellcolor[HTML]{FFFFFF}{\color[HTML]{000000} 82.6}} &
  \multicolumn{1}{c|}{\cellcolor[HTML]{FFFFFF}{\color[HTML]{000000} 99.9}} &
  \multicolumn{1}{c|}{\cellcolor[HTML]{FFFFFF}{\color[HTML]{000000} 99.6}} &
  {\color[HTML]{000000} 100} \\ \hline
\multicolumn{1}{|c|}{\cellcolor[HTML]{FFFFFF}{\color[HTML]{000000} }} &
  {\color[HTML]{000000} \textbf{TF-KD}} &
  \multicolumn{1}{c|}{\cellcolor[HTML]{FFFFFF}{\color[HTML]{000000} 81.8}} &
  \multicolumn{1}{c|}{\cellcolor[HTML]{FFFFFF}{\color[HTML]{000000} 96.3}} &
  \multicolumn{1}{c|}{\cellcolor[HTML]{FFFFFF}{\color[HTML]{000000} 98.0}} &
  {\color[HTML]{000000} 99.2} &
  \multicolumn{1}{c|}{\cellcolor[HTML]{FFFFFF}{\color[HTML]{000000} 82.7}} &
  \multicolumn{1}{c|}{\cellcolor[HTML]{FFFFFF}{\color[HTML]{000000} 99.9}} &
  \multicolumn{1}{c|}{\cellcolor[HTML]{FFFFFF}{\color[HTML]{000000} 99.3}} &
  {\color[HTML]{000000} 100} \\ \cline{2-10} 
\multicolumn{1}{|c|}{\cellcolor[HTML]{FFFFFF}{\color[HTML]{000000} }} &
  {\color[HTML]{000000} \textbf{CS-KD}} &
  \multicolumn{1}{c|}{\cellcolor[HTML]{FFFFFF}{\color[HTML]{000000} 83.4}} &
  \multicolumn{1}{c|}{\cellcolor[HTML]{FFFFFF}{\color[HTML]{000000} 98.5}} &
  \multicolumn{1}{c|}{\cellcolor[HTML]{FFFFFF}{\color[HTML]{000000} 98.8}} &
  {\color[HTML]{000000} 99.8} &
  \multicolumn{1}{c|}{\cellcolor[HTML]{FFFFFF}{\color[HTML]{000000} 81.4}} &
  \multicolumn{1}{c|}{\cellcolor[HTML]{FFFFFF}{\color[HTML]{000000} 99.7}} &
  \multicolumn{1}{c|}{\cellcolor[HTML]{FFFFFF}{\color[HTML]{000000} 99.1}} &
  {\color[HTML]{000000} 100} \\ \cline{2-10} 
\multicolumn{1}{|c|}{\cellcolor[HTML]{FFFFFF}{\color[HTML]{000000} }} &
  {\color[HTML]{000000} \textbf{PS-KD}} &
  \multicolumn{1}{c|}{\cellcolor[HTML]{FFFFFF}{\color[HTML]{000000} 86.5}} &
  \multicolumn{1}{c|}{\cellcolor[HTML]{FFFFFF}{\color[HTML]{000000} 97.5}} &
  \multicolumn{1}{c|}{\cellcolor[HTML]{FFFFFF}{\color[HTML]{000000} 99.4}} &
  {\color[HTML]{000000} 100} &
  \multicolumn{1}{c|}{\cellcolor[HTML]{FFFFFF}{\color[HTML]{000000} 80.0}} &
  \multicolumn{1}{c|}{\cellcolor[HTML]{FFFFFF}{\color[HTML]{000000} 100}} &
  \multicolumn{1}{c|}{\cellcolor[HTML]{FFFFFF}{\color[HTML]{000000} 99.4}} &
  {\color[HTML]{000000} 100} \\ \cline{2-10} 
\multicolumn{1}{|c|}{\multirow{-4}{*}{\cellcolor[HTML]{FFFFFF}{\color[HTML]{000000} \textbf{\begin{tabular}[c]{@{}c@{}}Self \\     Distillation\end{tabular}}}}} &
  {\color[HTML]{000000} \textbf{DDGSD}} &
  \multicolumn{1}{c|}{\cellcolor[HTML]{FFFFFF}{\color[HTML]{000000} 87.6}} &
  \multicolumn{1}{c|}{\cellcolor[HTML]{FFFFFF}{\color[HTML]{000000} 97.8}} &
  \multicolumn{1}{c|}{\cellcolor[HTML]{FFFFFF}{\color[HTML]{000000} 99.3}} &
  {\color[HTML]{000000} 100} &
  \multicolumn{1}{c|}{\cellcolor[HTML]{FFFFFF}{\color[HTML]{000000} 86.1}} &
  \multicolumn{1}{c|}{\cellcolor[HTML]{FFFFFF}{\color[HTML]{000000} 99.9}} &
  \multicolumn{1}{c|}{\cellcolor[HTML]{FFFFFF}{\color[HTML]{000000} 99.4}} &
  {\color[HTML]{000000} 100} \\ \hline
\end{tabular}
\caption{Top-1 and Top-5 accuracy of student model using knowledge distillation methods (90\% reduction)}
\label{tab:kd90}
\end{table*}

\begin{table*}[!h]
\centering
\footnotesize
\begin{tabular}{|
>{\columncolor[HTML]{FFFFFF}}c 
>{\columncolor[HTML]{FFFFFF}}c |
>{\columncolor[HTML]{FFFFFF}}c 
>{\columncolor[HTML]{FFFFFF}}c 
>{\columncolor[HTML]{FFFFFF}}c 
>{\columncolor[HTML]{FFFFFF}}c |
>{\columncolor[HTML]{FFFFFF}}c 
>{\columncolor[HTML]{FFFFFF}}c 
>{\columncolor[HTML]{FFFFFF}}c 
>{\columncolor[HTML]{FFFFFF}}c |}
\hline
\multicolumn{2}{|c|}{\cellcolor[HTML]{FFFFFF}{\color[HTML]{000000} }} &
  \multicolumn{4}{c|}{\cellcolor[HTML]{FFFFFF}{\color[HTML]{000000} \textbf{INDIAN PINES}}} &
  \multicolumn{4}{c|}{\cellcolor[HTML]{FFFFFF}{\color[HTML]{000000} \textbf{PAVIA UNIVERSITY}}} \\ \cline{3-10} 
\multicolumn{2}{|c|}{\cellcolor[HTML]{FFFFFF}{\color[HTML]{000000} }} &
  \multicolumn{2}{c|}{\cellcolor[HTML]{FFFFFF}{\color[HTML]{000000} Disjoint}} &
  \multicolumn{2}{c|}{\cellcolor[HTML]{FFFFFF}{\color[HTML]{000000} Random}} &
  \multicolumn{2}{c|}{\cellcolor[HTML]{FFFFFF}{\color[HTML]{000000} Disjoint}} &
  \multicolumn{2}{c|}{\cellcolor[HTML]{FFFFFF}{\color[HTML]{000000} Random}} \\ \cline{3-10} 
\multicolumn{2}{|c|}{\multirow{-3}{*}{\cellcolor[HTML]{FFFFFF}{\color[HTML]{000000} \textbf{\begin{tabular}[c]{@{}c@{}}Methods\\    \\ (student network: 95\%)\end{tabular}}}}} &
  \multicolumn{1}{c|}{\cellcolor[HTML]{FFFFFF}{\color[HTML]{000000} Top-1}} &
  \multicolumn{1}{c|}{\cellcolor[HTML]{FFFFFF}{\color[HTML]{000000} Top-5}} &
  \multicolumn{1}{c|}{\cellcolor[HTML]{FFFFFF}{\color[HTML]{000000} Top-1}} &
  {\color[HTML]{000000} Top-5} &
  \multicolumn{1}{c|}{\cellcolor[HTML]{FFFFFF}{\color[HTML]{000000} Top-1}} &
  \multicolumn{1}{l|}{\cellcolor[HTML]{FFFFFF}{\color[HTML]{000000} Top-5}} &
  \multicolumn{1}{c|}{\cellcolor[HTML]{FFFFFF}{\color[HTML]{000000} Top-1}} &
  {\color[HTML]{000000} Top-5} \\ \hline
\multicolumn{1}{|c|}{\cellcolor[HTML]{FFFFFF}{\color[HTML]{000000} }} &
  {\color[HTML]{000000} \textbf{MLP}} &
  \multicolumn{1}{c|}{\cellcolor[HTML]{FFFFFF}{\color[HTML]{000000} 82.4}} &
  \multicolumn{1}{c|}{\cellcolor[HTML]{FFFFFF}{\color[HTML]{000000} 98.6}} &
  \multicolumn{1}{c|}{\cellcolor[HTML]{FFFFFF}{\color[HTML]{000000} 88.9}} &
  {\color[HTML]{000000} 99.7} &
  \multicolumn{1}{c|}{\cellcolor[HTML]{FFFFFF}{\color[HTML]{000000} 82.5}} &
  \multicolumn{1}{c|}{\cellcolor[HTML]{FFFFFF}{\color[HTML]{000000} 99.5}} &
  \multicolumn{1}{c|}{\cellcolor[HTML]{FFFFFF}{\color[HTML]{000000} 92.5}} &
  {\color[HTML]{000000} 99.9} \\ \cline{2-10} 
\multicolumn{1}{|c|}{\cellcolor[HTML]{FFFFFF}{\color[HTML]{000000} }} &
  {\color[HTML]{000000} \textbf{CNN-1D}} &
  \multicolumn{1}{c|}{\cellcolor[HTML]{FFFFFF}{\color[HTML]{000000} 82.5}} &
  \multicolumn{1}{c|}{\cellcolor[HTML]{FFFFFF}{\color[HTML]{000000} 97.9}} &
  \multicolumn{1}{c|}{\cellcolor[HTML]{FFFFFF}{\color[HTML]{000000} 89.4}} &
  {\color[HTML]{000000} 99.5} &
  \multicolumn{1}{c|}{\cellcolor[HTML]{FFFFFF}{\color[HTML]{000000} 82.3}} &
  \multicolumn{1}{c|}{\cellcolor[HTML]{FFFFFF}{\color[HTML]{000000} 99.7}} &
  \multicolumn{1}{c|}{\cellcolor[HTML]{FFFFFF}{\color[HTML]{000000} 93.5}} &
  {\color[HTML]{000000} 99.9} \\ \cline{2-10} 
\multicolumn{1}{|c|}{\multirow{-3}{*}{\cellcolor[HTML]{FFFFFF}{\color[HTML]{000000} \textbf{Baselines}}}} &
  {\color[HTML]{000000} \textbf{CNN-2D}} &
  \multicolumn{1}{c|}{\cellcolor[HTML]{FFFFFF}{\color[HTML]{000000} 86.3}} &
  \multicolumn{1}{c|}{\cellcolor[HTML]{FFFFFF}{\color[HTML]{000000} 98.7}} &
  \multicolumn{1}{c|}{\cellcolor[HTML]{FFFFFF}{\color[HTML]{000000} 99.5}} &
  {\color[HTML]{000000} 99.9} &
  \multicolumn{1}{c|}{\cellcolor[HTML]{FFFFFF}{\color[HTML]{000000} 83.2}} &
  \multicolumn{1}{c|}{\cellcolor[HTML]{FFFFFF}{\color[HTML]{000000} 99.9}} &
  \multicolumn{1}{c|}{\cellcolor[HTML]{FFFFFF}{\color[HTML]{000000} 99.3}} &
  {\color[HTML]{000000} 100} \\ \hline
\multicolumn{1}{|c|}{\cellcolor[HTML]{FFFFFF}{\color[HTML]{000000} \textbf{Scratch}}} &
  {\color[HTML]{000000} \textbf{Pruned CNN2D}} &
  \multicolumn{1}{c|}{\cellcolor[HTML]{FFFFFF}{\color[HTML]{000000} 82.1}} &
  \multicolumn{1}{c|}{\cellcolor[HTML]{FFFFFF}{\color[HTML]{000000} 97.1}} &
  \multicolumn{1}{c|}{\cellcolor[HTML]{FFFFFF}{\color[HTML]{000000} 99.3}} &
  {\color[HTML]{000000} 100} &
  \multicolumn{1}{c|}{\cellcolor[HTML]{FFFFFF}{\color[HTML]{000000} 74.1}} &
  \multicolumn{1}{c|}{\cellcolor[HTML]{FFFFFF}{\color[HTML]{000000} 99.6}} &
  \multicolumn{1}{c|}{\cellcolor[HTML]{FFFFFF}{\color[HTML]{000000} 99.0}} &
  {\color[HTML]{000000} 100} \\ \hline
\multicolumn{1}{|c|}{\cellcolor[HTML]{FFFFFF}{\color[HTML]{000000} }} &
  {\color[HTML]{000000} \textbf{Soft targets}} &
  \multicolumn{1}{c|}{\cellcolor[HTML]{FFFFFF}{\color[HTML]{000000} 83.9}} &
  \multicolumn{1}{c|}{\cellcolor[HTML]{FFFFFF}{\color[HTML]{000000} 97.6}} &
  \multicolumn{1}{c|}{\cellcolor[HTML]{FFFFFF}{\color[HTML]{000000} 99.5}} &
  {\color[HTML]{000000} 99.9} &
  \multicolumn{1}{c|}{\cellcolor[HTML]{FFFFFF}{\color[HTML]{000000} 80.3}} &
  \multicolumn{1}{c|}{\cellcolor[HTML]{FFFFFF}{\color[HTML]{000000} 99.8}} &
  \multicolumn{1}{c|}{\cellcolor[HTML]{FFFFFF}{\color[HTML]{000000} 99.3}} &
  {\color[HTML]{000000} 100} \\ \cline{2-10} 
\multicolumn{1}{|c|}{\cellcolor[HTML]{FFFFFF}{\color[HTML]{000000} }} &
  {\color[HTML]{000000} \textbf{FitNets}} &
  \multicolumn{1}{c|}{\cellcolor[HTML]{FFFFFF}{\color[HTML]{000000} 87.3}} &
  \multicolumn{1}{c|}{\cellcolor[HTML]{FFFFFF}{\color[HTML]{000000} 98.4}} &
  \multicolumn{1}{c|}{\cellcolor[HTML]{FFFFFF}{\color[HTML]{000000} 99.3}} &
  {\color[HTML]{000000} 99.9} &
  \multicolumn{1}{c|}{\cellcolor[HTML]{FFFFFF}{\color[HTML]{000000} 80.8}} &
  \multicolumn{1}{c|}{\cellcolor[HTML]{FFFFFF}{\color[HTML]{000000} 99.9}} &
  \multicolumn{1}{c|}{\cellcolor[HTML]{FFFFFF}{\color[HTML]{000000} 99.3}} &
  {\color[HTML]{000000} 100} \\ \cline{2-10} 
\multicolumn{1}{|c|}{\cellcolor[HTML]{FFFFFF}{\color[HTML]{000000} }} &
  {\color[HTML]{000000} \textbf{Attention Transfer}} &
  \multicolumn{1}{c|}{\cellcolor[HTML]{FFFFFF}{\color[HTML]{000000} 85.7}} &
  \multicolumn{1}{c|}{\cellcolor[HTML]{FFFFFF}{\color[HTML]{000000} 98.3}} &
  \multicolumn{1}{c|}{\cellcolor[HTML]{FFFFFF}{\color[HTML]{000000} 99.4}} &
  {\color[HTML]{000000} 100} &
  \multicolumn{1}{c|}{\cellcolor[HTML]{FFFFFF}{\color[HTML]{000000} 82.3}} &
  \multicolumn{1}{c|}{\cellcolor[HTML]{FFFFFF}{\color[HTML]{000000} 99.9}} &
  \multicolumn{1}{c|}{\cellcolor[HTML]{FFFFFF}{\color[HTML]{000000} 99.3}} &
  {\color[HTML]{000000} 100} \\ \cline{2-10} 
\multicolumn{1}{|c|}{\cellcolor[HTML]{FFFFFF}{\color[HTML]{000000} }} &
  {\color[HTML]{000000} \textbf{Correlation Congruence}} &
  \multicolumn{1}{c|}{\cellcolor[HTML]{FFFFFF}{\color[HTML]{000000} 82.5}} &
  \multicolumn{1}{c|}{\cellcolor[HTML]{FFFFFF}{\color[HTML]{000000} 98.5}} &
  \multicolumn{1}{c|}{\cellcolor[HTML]{FFFFFF}{\color[HTML]{000000} 99.4}} &
  {\color[HTML]{000000} 100} &
  \multicolumn{1}{c|}{\cellcolor[HTML]{FFFFFF}{\color[HTML]{000000} 78.5}} &
  \multicolumn{1}{c|}{\cellcolor[HTML]{FFFFFF}{\color[HTML]{000000} 99.9}} &
  \multicolumn{1}{c|}{\cellcolor[HTML]{FFFFFF}{\color[HTML]{000000} 99.2}} &
  {\color[HTML]{000000} 100} \\ \cline{2-10} 
\multicolumn{1}{|c|}{\cellcolor[HTML]{FFFFFF}{\color[HTML]{000000} }} &
  {\color[HTML]{000000} \textbf{SimKD}} &
  \multicolumn{1}{c|}{\cellcolor[HTML]{FFFFFF}{\color[HTML]{000000} 86.9}} &
  \multicolumn{1}{c|}{\cellcolor[HTML]{FFFFFF}{\color[HTML]{000000} 98.4}} &
  \multicolumn{1}{c|}{\cellcolor[HTML]{FFFFFF}{\color[HTML]{000000} 99.4}} &
  {\color[HTML]{000000} 100} &
  \multicolumn{1}{c|}{\cellcolor[HTML]{FFFFFF}{\color[HTML]{000000} 81.0}} &
  \multicolumn{1}{c|}{\cellcolor[HTML]{FFFFFF}{\color[HTML]{000000} 99.9}} &
  \multicolumn{1}{c|}{\cellcolor[HTML]{FFFFFF}{\color[HTML]{000000} 99.0}} &
  {\color[HTML]{000000} 99.9} \\ \cline{2-10} 
\multicolumn{1}{|c|}{\multirow{-6}{*}{\cellcolor[HTML]{FFFFFF}{\color[HTML]{000000} \textbf{\begin{tabular}[c]{@{}c@{}}Offline \\    Distillation\end{tabular}}}}} &
  {\color[HTML]{000000} \textbf{CA-MKD}} &
  \multicolumn{1}{c|}{\cellcolor[HTML]{FFFFFF}{\color[HTML]{000000} 86.4}} &
  \multicolumn{1}{c|}{\cellcolor[HTML]{FFFFFF}{\color[HTML]{000000} 97.9}} &
  \multicolumn{1}{c|}{\cellcolor[HTML]{FFFFFF}{\color[HTML]{000000} 99.4}} &
  {\color[HTML]{000000} 100} &
  \multicolumn{1}{c|}{\cellcolor[HTML]{FFFFFF}{\color[HTML]{000000} 78.4}} &
  \multicolumn{1}{c|}{\cellcolor[HTML]{FFFFFF}{\color[HTML]{000000} 99.9}} &
  \multicolumn{1}{c|}{\cellcolor[HTML]{FFFFFF}{\color[HTML]{000000} 99.3}} &
  {\color[HTML]{000000} 99.9} \\ \hline
\multicolumn{1}{|c|}{\cellcolor[HTML]{FFFFFF}{\color[HTML]{000000} }} &
  {\color[HTML]{000000} \textbf{DML}} &
  \multicolumn{1}{c|}{\cellcolor[HTML]{FFFFFF}{\color[HTML]{000000} 83.5}} &
  \multicolumn{1}{c|}{\cellcolor[HTML]{FFFFFF}{\color[HTML]{000000} 97.0}} &
  \multicolumn{1}{c|}{\cellcolor[HTML]{FFFFFF}{\color[HTML]{000000} 99.4}} &
  {\color[HTML]{000000} 100} &
  \multicolumn{1}{c|}{\cellcolor[HTML]{FFFFFF}{\color[HTML]{000000} 81.3}} &
  \multicolumn{1}{c|}{\cellcolor[HTML]{FFFFFF}{\color[HTML]{000000} 100}} &
  \multicolumn{1}{c|}{\cellcolor[HTML]{FFFFFF}{\color[HTML]{000000} 99.4}} &
  {\color[HTML]{000000} 100} \\ \cline{2-10} 
\multicolumn{1}{|c|}{\cellcolor[HTML]{FFFFFF}{\color[HTML]{000000} }} &
  {\color[HTML]{000000} \textbf{ONE}} &
  \multicolumn{1}{c|}{\cellcolor[HTML]{FFFFFF}{\color[HTML]{000000} 83.7}} &
  \multicolumn{1}{c|}{\cellcolor[HTML]{FFFFFF}{\color[HTML]{000000} 97.9}} &
  \multicolumn{1}{c|}{\cellcolor[HTML]{FFFFFF}{\color[HTML]{000000} 99.5}} &
  {\color[HTML]{000000} 100} &
  \multicolumn{1}{c|}{\cellcolor[HTML]{FFFFFF}{\color[HTML]{000000} 78.7}} &
  \multicolumn{1}{c|}{\cellcolor[HTML]{FFFFFF}{\color[HTML]{000000} 99.5}} &
  \multicolumn{1}{c|}{\cellcolor[HTML]{FFFFFF}{\color[HTML]{000000} 99.6}} &
  {\color[HTML]{000000} 100} \\ \cline{2-10} 
\multicolumn{1}{|c|}{\cellcolor[HTML]{FFFFFF}{\color[HTML]{000000} }} &
  {\color[HTML]{000000} \textbf{CL-ILR}} &
  \multicolumn{1}{c|}{\cellcolor[HTML]{FFFFFF}{\color[HTML]{000000} 83.2}} &
  \multicolumn{1}{c|}{\cellcolor[HTML]{FFFFFF}{\color[HTML]{000000} 97.7}} &
  \multicolumn{1}{c|}{\cellcolor[HTML]{FFFFFF}{\color[HTML]{000000} 99.0}} &
  {\color[HTML]{000000} 100} &
  \multicolumn{1}{c|}{\cellcolor[HTML]{FFFFFF}{\color[HTML]{000000} 81.5}} &
  \multicolumn{1}{c|}{\cellcolor[HTML]{FFFFFF}{\color[HTML]{000000} 99.0}} &
  \multicolumn{1}{c|}{\cellcolor[HTML]{FFFFFF}{\color[HTML]{000000} 99.3}} &
  {\color[HTML]{000000} 100} \\ \cline{2-10} 
\multicolumn{1}{|c|}{\multirow{-4}{*}{\cellcolor[HTML]{FFFFFF}{\color[HTML]{000000} \textbf{\begin{tabular}[c]{@{}c@{}}Online\\     Distillation\end{tabular}}}}} &
  {\color[HTML]{000000} \textbf{OKDDip}} &
  \multicolumn{1}{c|}{\cellcolor[HTML]{FFFFFF}{\color[HTML]{000000} 89.2}} &
  \multicolumn{1}{c|}{\cellcolor[HTML]{FFFFFF}{\color[HTML]{000000} 98.0}} &
  \multicolumn{1}{c|}{\cellcolor[HTML]{FFFFFF}{\color[HTML]{000000} 99.4}} &
  {\color[HTML]{000000} 100} &
  \multicolumn{1}{c|}{\cellcolor[HTML]{FFFFFF}{\color[HTML]{000000} 85.5}} &
  \multicolumn{1}{c|}{\cellcolor[HTML]{FFFFFF}{\color[HTML]{000000} 99.9}} &
  \multicolumn{1}{c|}{\cellcolor[HTML]{FFFFFF}{\color[HTML]{000000} 99.5}} &
  {\color[HTML]{000000} 100} \\ \hline
\multicolumn{1}{|c|}{\cellcolor[HTML]{FFFFFF}{\color[HTML]{000000} }} &
  {\color[HTML]{000000} \textbf{TF-KD}} &
  \multicolumn{1}{c|}{\cellcolor[HTML]{FFFFFF}{\color[HTML]{000000} 85.4}} &
  \multicolumn{1}{c|}{\cellcolor[HTML]{FFFFFF}{\color[HTML]{000000} 97.6}} &
  \multicolumn{1}{c|}{\cellcolor[HTML]{FFFFFF}{\color[HTML]{000000} 97.4}} &
  {\color[HTML]{000000} 99.2} &
  \multicolumn{1}{c|}{\cellcolor[HTML]{FFFFFF}{\color[HTML]{000000} 77.6}} &
  \multicolumn{1}{c|}{\cellcolor[HTML]{FFFFFF}{\color[HTML]{000000} 99.4}} &
  \multicolumn{1}{c|}{\cellcolor[HTML]{FFFFFF}{\color[HTML]{000000} 99.3}} &
  {\color[HTML]{000000} 100} \\ \cline{2-10} 
\multicolumn{1}{|c|}{\cellcolor[HTML]{FFFFFF}{\color[HTML]{000000} }} &
  {\color[HTML]{000000} \textbf{CS-KD}} &
  \multicolumn{1}{c|}{\cellcolor[HTML]{FFFFFF}{\color[HTML]{000000} 77.2}} &
  \multicolumn{1}{c|}{\cellcolor[HTML]{FFFFFF}{\color[HTML]{000000} 95.8}} &
  \multicolumn{1}{c|}{\cellcolor[HTML]{FFFFFF}{\color[HTML]{000000} 98.5}} &
  {\color[HTML]{000000} 99.7} &
  \multicolumn{1}{c|}{\cellcolor[HTML]{FFFFFF}{\color[HTML]{000000} 79.9}} &
  \multicolumn{1}{c|}{\cellcolor[HTML]{FFFFFF}{\color[HTML]{000000} 99.6}} &
  \multicolumn{1}{c|}{\cellcolor[HTML]{FFFFFF}{\color[HTML]{000000} 98.3}} &
  {\color[HTML]{000000} 100} \\ \cline{2-10} 
\multicolumn{1}{|c|}{\cellcolor[HTML]{FFFFFF}{\color[HTML]{000000} }} &
  {\color[HTML]{000000} \textbf{PS-KD}} &
  \multicolumn{1}{c|}{\cellcolor[HTML]{FFFFFF}{\color[HTML]{000000} 82.0}} &
  \multicolumn{1}{c|}{\cellcolor[HTML]{FFFFFF}{\color[HTML]{000000} 98.2}} &
  \multicolumn{1}{c|}{\cellcolor[HTML]{FFFFFF}{\color[HTML]{000000} 99.3}} &
  {\color[HTML]{000000} 100} &
  \multicolumn{1}{c|}{\cellcolor[HTML]{FFFFFF}{\color[HTML]{000000} 88.0}} &
  \multicolumn{1}{c|}{\cellcolor[HTML]{FFFFFF}{\color[HTML]{000000} 99.7}} &
  \multicolumn{1}{c|}{\cellcolor[HTML]{FFFFFF}{\color[HTML]{000000} 99.3}} &
  {\color[HTML]{000000} 100} \\ \cline{2-10} 
\multicolumn{1}{|c|}{\multirow{-4}{*}{\cellcolor[HTML]{FFFFFF}{\color[HTML]{000000} \textbf{\begin{tabular}[c]{@{}c@{}}Self \\     Distillation\end{tabular}}}}} &
  {\color[HTML]{000000} \textbf{DDGSD}} &
  \multicolumn{1}{c|}{\cellcolor[HTML]{FFFFFF}{\color[HTML]{000000} 84.1}} &
  \multicolumn{1}{c|}{\cellcolor[HTML]{FFFFFF}{\color[HTML]{000000} 97.9}} &
  \multicolumn{1}{c|}{\cellcolor[HTML]{FFFFFF}{\color[HTML]{000000} 99.0}} &
  {\color[HTML]{000000} 100} &
  \multicolumn{1}{c|}{\cellcolor[HTML]{FFFFFF}{\color[HTML]{000000} 81.5}} &
  \multicolumn{1}{c|}{\cellcolor[HTML]{FFFFFF}{\color[HTML]{000000} 100}} &
  \multicolumn{1}{c|}{\cellcolor[HTML]{FFFFFF}{\color[HTML]{000000} 99.2}} &
  {\color[HTML]{000000} 100} \\ \hline
\end{tabular}
\caption{Top-1 and Top-5 accuracy of student model using knowledge distillation methods (95\% reduction)}
\label{tab:kd95}
\end{table*}

\begin{table*}[!h]
\centering
\footnotesize
\begin{tabular}{|
>{\columncolor[HTML]{FFFFFF}}c 
>{\columncolor[HTML]{FFFFFF}}c |
>{\columncolor[HTML]{FFFFFF}}c 
>{\columncolor[HTML]{FFFFFF}}c 
>{\columncolor[HTML]{FFFFFF}}c 
>{\columncolor[HTML]{FFFFFF}}c |
>{\columncolor[HTML]{FFFFFF}}c 
>{\columncolor[HTML]{FFFFFF}}c 
>{\columncolor[HTML]{FFFFFF}}c 
>{\columncolor[HTML]{FFFFFF}}c |}
\hline
\multicolumn{2}{|c|}{\cellcolor[HTML]{FFFFFF}{\color[HTML]{000000} }} &
  \multicolumn{4}{c|}{\cellcolor[HTML]{FFFFFF}{\color[HTML]{000000} \textbf{INDIAN PINES}}} &
  \multicolumn{4}{c|}{\cellcolor[HTML]{FFFFFF}{\color[HTML]{000000} \textbf{PAVIA UNIVERSITY}}} \\ \cline{3-10} 
\multicolumn{2}{|c|}{\cellcolor[HTML]{FFFFFF}{\color[HTML]{000000} }} &
  \multicolumn{2}{c|}{\cellcolor[HTML]{FFFFFF}{\color[HTML]{000000} Disjoint}} &
  \multicolumn{2}{c|}{\cellcolor[HTML]{FFFFFF}{\color[HTML]{000000} Random}} &
  \multicolumn{2}{c|}{\cellcolor[HTML]{FFFFFF}{\color[HTML]{000000} Disjoint}} &
  \multicolumn{2}{c|}{\cellcolor[HTML]{FFFFFF}{\color[HTML]{000000} Random}} \\ \cline{3-10} 
\multicolumn{2}{|c|}{\multirow{-3}{*}{\cellcolor[HTML]{FFFFFF}{\color[HTML]{000000} \textbf{\begin{tabular}[c]{@{}c@{}}Methods\\    \\ (student network: 98\%)\end{tabular}}}}} &
  \multicolumn{1}{c|}{\cellcolor[HTML]{FFFFFF}{\color[HTML]{000000} Top-1}} &
  \multicolumn{1}{c|}{\cellcolor[HTML]{FFFFFF}{\color[HTML]{000000} Top-5}} &
  \multicolumn{1}{c|}{\cellcolor[HTML]{FFFFFF}{\color[HTML]{000000} Top-1}} &
  {\color[HTML]{000000} Top-5} &
  \multicolumn{1}{c|}{\cellcolor[HTML]{FFFFFF}{\color[HTML]{000000} Top-1}} &
  \multicolumn{1}{l|}{\cellcolor[HTML]{FFFFFF}{\color[HTML]{000000} Top-5}} &
  \multicolumn{1}{c|}{\cellcolor[HTML]{FFFFFF}{\color[HTML]{000000} Top-1}} &
  {\color[HTML]{000000} Top-5} \\ \hline
\multicolumn{1}{|c|}{\cellcolor[HTML]{FFFFFF}{\color[HTML]{000000} }} &
  {\color[HTML]{000000} \textbf{MLP}} &
  \multicolumn{1}{c|}{\cellcolor[HTML]{FFFFFF}{\color[HTML]{000000} 82.4}} &
  \multicolumn{1}{c|}{\cellcolor[HTML]{FFFFFF}{\color[HTML]{000000} 98.6}} &
  \multicolumn{1}{c|}{\cellcolor[HTML]{FFFFFF}{\color[HTML]{000000} 88.9}} &
  {\color[HTML]{000000} 99.7} &
  \multicolumn{1}{c|}{\cellcolor[HTML]{FFFFFF}{\color[HTML]{000000} 82.5}} &
  \multicolumn{1}{c|}{\cellcolor[HTML]{FFFFFF}{\color[HTML]{000000} 99.5}} &
  \multicolumn{1}{c|}{\cellcolor[HTML]{FFFFFF}{\color[HTML]{000000} 92.5}} &
  {\color[HTML]{000000} 99.9} \\ \cline{2-10} 
\multicolumn{1}{|c|}{\cellcolor[HTML]{FFFFFF}{\color[HTML]{000000} }} &
  {\color[HTML]{000000} \textbf{CNN-1D}} &
  \multicolumn{1}{c|}{\cellcolor[HTML]{FFFFFF}{\color[HTML]{000000} 82.5}} &
  \multicolumn{1}{c|}{\cellcolor[HTML]{FFFFFF}{\color[HTML]{000000} 97.9}} &
  \multicolumn{1}{c|}{\cellcolor[HTML]{FFFFFF}{\color[HTML]{000000} 89.4}} &
  {\color[HTML]{000000} 99.5} &
  \multicolumn{1}{c|}{\cellcolor[HTML]{FFFFFF}{\color[HTML]{000000} 82.3}} &
  \multicolumn{1}{c|}{\cellcolor[HTML]{FFFFFF}{\color[HTML]{000000} 99.7}} &
  \multicolumn{1}{c|}{\cellcolor[HTML]{FFFFFF}{\color[HTML]{000000} 93.5}} &
  {\color[HTML]{000000} 99.9} \\ \cline{2-10} 
\multicolumn{1}{|c|}{\multirow{-3}{*}{\cellcolor[HTML]{FFFFFF}{\color[HTML]{000000} \textbf{Baselines}}}} &
  {\color[HTML]{000000} \textbf{CNN-2D}} &
  \multicolumn{1}{c|}{\cellcolor[HTML]{FFFFFF}{\color[HTML]{000000} 86.3}} &
  \multicolumn{1}{c|}{\cellcolor[HTML]{FFFFFF}{\color[HTML]{000000} 98.7}} &
  \multicolumn{1}{c|}{\cellcolor[HTML]{FFFFFF}{\color[HTML]{000000} 99.5}} &
  {\color[HTML]{000000} 99.9} &
  \multicolumn{1}{c|}{\cellcolor[HTML]{FFFFFF}{\color[HTML]{000000} 83.2}} &
  \multicolumn{1}{c|}{\cellcolor[HTML]{FFFFFF}{\color[HTML]{000000} 99.9}} &
  \multicolumn{1}{c|}{\cellcolor[HTML]{FFFFFF}{\color[HTML]{000000} 99.3}} &
  {\color[HTML]{000000} 100} \\ \hline
\multicolumn{1}{|c|}{\cellcolor[HTML]{FFFFFF}{\color[HTML]{000000} \textbf{Scratch}}} &
  {\color[HTML]{000000} \textbf{Pruned CNN2D}} &
  \multicolumn{1}{c|}{\cellcolor[HTML]{FFFFFF}{\color[HTML]{000000} 78.2}} &
  \multicolumn{1}{c|}{\cellcolor[HTML]{FFFFFF}{\color[HTML]{000000} 96.4}} &
  \multicolumn{1}{c|}{\cellcolor[HTML]{FFFFFF}{\color[HTML]{000000} 98.9}} &
  {\color[HTML]{000000} 100} &
  \multicolumn{1}{c|}{\cellcolor[HTML]{FFFFFF}{\color[HTML]{000000} 77.2}} &
  \multicolumn{1}{c|}{\cellcolor[HTML]{FFFFFF}{\color[HTML]{000000} 99.9}} &
  \multicolumn{1}{c|}{\cellcolor[HTML]{FFFFFF}{\color[HTML]{000000} 98.1}} &
  {\color[HTML]{000000} 100} \\ \hline
\multicolumn{1}{|c|}{\cellcolor[HTML]{FFFFFF}{\color[HTML]{000000} }} &
  {\color[HTML]{000000} \textbf{Soft targets}} &
  \multicolumn{1}{c|}{\cellcolor[HTML]{FFFFFF}{\color[HTML]{000000} 82.1}} &
  \multicolumn{1}{c|}{\cellcolor[HTML]{FFFFFF}{\color[HTML]{000000} 97.1}} &
  \multicolumn{1}{c|}{\cellcolor[HTML]{FFFFFF}{\color[HTML]{000000} 99.3}} &
  {\color[HTML]{000000} 99.9} &
  \multicolumn{1}{c|}{\cellcolor[HTML]{FFFFFF}{\color[HTML]{000000} 77.3}} &
  \multicolumn{1}{c|}{\cellcolor[HTML]{FFFFFF}{\color[HTML]{000000} 99.9}} &
  \multicolumn{1}{c|}{\cellcolor[HTML]{FFFFFF}{\color[HTML]{000000} 98.4}} &
  {\color[HTML]{000000} 99.9} \\ \cline{2-10} 
\multicolumn{1}{|c|}{\cellcolor[HTML]{FFFFFF}{\color[HTML]{000000} }} &
  {\color[HTML]{000000} \textbf{FitNets}} &
  \multicolumn{1}{c|}{\cellcolor[HTML]{FFFFFF}{\color[HTML]{000000} 84.2}} &
  \multicolumn{1}{c|}{\cellcolor[HTML]{FFFFFF}{\color[HTML]{000000} 98.8}} &
  \multicolumn{1}{c|}{\cellcolor[HTML]{FFFFFF}{\color[HTML]{000000} 99.3}} &
  {\color[HTML]{000000} 100} &
  \multicolumn{1}{c|}{\cellcolor[HTML]{FFFFFF}{\color[HTML]{000000} 79.3}} &
  \multicolumn{1}{c|}{\cellcolor[HTML]{FFFFFF}{\color[HTML]{000000} 99.8}} &
  \multicolumn{1}{c|}{\cellcolor[HTML]{FFFFFF}{\color[HTML]{000000} 97.3}} &
  {\color[HTML]{000000} 99.9} \\ \cline{2-10} 
\multicolumn{1}{|c|}{\cellcolor[HTML]{FFFFFF}{\color[HTML]{000000} }} &
  {\color[HTML]{000000} \textbf{Attention Transfer}} &
  \multicolumn{1}{c|}{\cellcolor[HTML]{FFFFFF}{\color[HTML]{000000} 84.9}} &
  \multicolumn{1}{c|}{\cellcolor[HTML]{FFFFFF}{\color[HTML]{000000} 98.6}} &
  \multicolumn{1}{c|}{\cellcolor[HTML]{FFFFFF}{\color[HTML]{000000} 99.3}} &
  {\color[HTML]{000000} 100} &
  \multicolumn{1}{c|}{\cellcolor[HTML]{FFFFFF}{\color[HTML]{000000} 75.4}} &
  \multicolumn{1}{c|}{\cellcolor[HTML]{FFFFFF}{\color[HTML]{000000} 99.8}} &
  \multicolumn{1}{c|}{\cellcolor[HTML]{FFFFFF}{\color[HTML]{000000} 98.1}} &
  {\color[HTML]{000000} 100} \\ \cline{2-10} 
\multicolumn{1}{|c|}{\cellcolor[HTML]{FFFFFF}{\color[HTML]{000000} }} &
  {\color[HTML]{000000} \textbf{Correlation Congruence}} &
  \multicolumn{1}{c|}{\cellcolor[HTML]{FFFFFF}{\color[HTML]{000000} 77.1}} &
  \multicolumn{1}{c|}{\cellcolor[HTML]{FFFFFF}{\color[HTML]{000000} 95.0}} &
  \multicolumn{1}{c|}{\cellcolor[HTML]{FFFFFF}{\color[HTML]{000000} 98.9}} &
  {\color[HTML]{000000} 100} &
  \multicolumn{1}{c|}{\cellcolor[HTML]{FFFFFF}{\color[HTML]{000000} 84.8}} &
  \multicolumn{1}{c|}{\cellcolor[HTML]{FFFFFF}{\color[HTML]{000000} 99.7}} &
  \multicolumn{1}{c|}{\cellcolor[HTML]{FFFFFF}{\color[HTML]{000000} 98.3}} &
  {\color[HTML]{000000} 100} \\ \cline{2-10} 
\multicolumn{1}{|c|}{\cellcolor[HTML]{FFFFFF}{\color[HTML]{000000} }} &
  {\color[HTML]{000000} \textbf{SimKD}} &
  \multicolumn{1}{c|}{\cellcolor[HTML]{FFFFFF}{\color[HTML]{000000} 82.8}} &
  \multicolumn{1}{c|}{\cellcolor[HTML]{FFFFFF}{\color[HTML]{000000} 97.2}} &
  \multicolumn{1}{c|}{\cellcolor[HTML]{FFFFFF}{\color[HTML]{000000} 98.6}} &
  {\color[HTML]{000000} 99.9} &
  \multicolumn{1}{c|}{\cellcolor[HTML]{FFFFFF}{\color[HTML]{000000} 81.1}} &
  \multicolumn{1}{c|}{\cellcolor[HTML]{FFFFFF}{\color[HTML]{000000} 99.9}} &
  \multicolumn{1}{c|}{\cellcolor[HTML]{FFFFFF}{\color[HTML]{000000} 97.2}} &
  {\color[HTML]{000000} 99.9} \\ \cline{2-10} 
\multicolumn{1}{|c|}{\multirow{-6}{*}{\cellcolor[HTML]{FFFFFF}{\color[HTML]{000000} \textbf{\begin{tabular}[c]{@{}c@{}}Offline \\    Distillation\end{tabular}}}}} &
  {\color[HTML]{000000} \textbf{CA-MKD}} &
  \multicolumn{1}{c|}{\cellcolor[HTML]{FFFFFF}{\color[HTML]{000000} 85.4}} &
  \multicolumn{1}{c|}{\cellcolor[HTML]{FFFFFF}{\color[HTML]{000000} 97.8}} &
  \multicolumn{1}{c|}{\cellcolor[HTML]{FFFFFF}{\color[HTML]{000000} 99.1}} &
  {\color[HTML]{000000} 100} &
  \multicolumn{1}{c|}{\cellcolor[HTML]{FFFFFF}{\color[HTML]{000000} 77.4}} &
  \multicolumn{1}{c|}{\cellcolor[HTML]{FFFFFF}{\color[HTML]{000000} 99.9}} &
  \multicolumn{1}{c|}{\cellcolor[HTML]{FFFFFF}{\color[HTML]{000000} 98.1}} &
  {\color[HTML]{000000} 99.9} \\ \hline
\multicolumn{1}{|c|}{\cellcolor[HTML]{FFFFFF}{\color[HTML]{000000} }} &
  {\color[HTML]{000000} \textbf{DML}} &
  \multicolumn{1}{c|}{\cellcolor[HTML]{FFFFFF}{\color[HTML]{000000} 73.7}} &
  \multicolumn{1}{c|}{\cellcolor[HTML]{FFFFFF}{\color[HTML]{000000} 95.2}} &
  \multicolumn{1}{c|}{\cellcolor[HTML]{FFFFFF}{\color[HTML]{000000} 98.9}} &
  {\color[HTML]{000000} 100} &
  \multicolumn{1}{c|}{\cellcolor[HTML]{FFFFFF}{\color[HTML]{000000} 77.1}} &
  \multicolumn{1}{c|}{\cellcolor[HTML]{FFFFFF}{\color[HTML]{000000} 99.9}} &
  \multicolumn{1}{c|}{\cellcolor[HTML]{FFFFFF}{\color[HTML]{000000} 97.9}} &
  {\color[HTML]{000000} 100} \\ \cline{2-10} 
\multicolumn{1}{|c|}{\cellcolor[HTML]{FFFFFF}{\color[HTML]{000000} }} &
  {\color[HTML]{000000} \textbf{ONE}} &
  \multicolumn{1}{c|}{\cellcolor[HTML]{FFFFFF}{\color[HTML]{000000} 76.5}} &
  \multicolumn{1}{c|}{\cellcolor[HTML]{FFFFFF}{\color[HTML]{000000} 96.7}} &
  \multicolumn{1}{c|}{\cellcolor[HTML]{FFFFFF}{\color[HTML]{000000} 99.4}} &
  {\color[HTML]{000000} 100} &
  \multicolumn{1}{c|}{\cellcolor[HTML]{FFFFFF}{\color[HTML]{000000} 76.1}} &
  \multicolumn{1}{c|}{\cellcolor[HTML]{FFFFFF}{\color[HTML]{000000} 99.1}} &
  \multicolumn{1}{c|}{\cellcolor[HTML]{FFFFFF}{\color[HTML]{000000} 98.7}} &
  {\color[HTML]{000000} 100} \\ \cline{2-10} 
\multicolumn{1}{|c|}{\cellcolor[HTML]{FFFFFF}{\color[HTML]{000000} }} &
  {\color[HTML]{000000} \textbf{CL-ILR}} &
  \multicolumn{1}{c|}{\cellcolor[HTML]{FFFFFF}{\color[HTML]{000000} 80.3}} &
  \multicolumn{1}{c|}{\cellcolor[HTML]{FFFFFF}{\color[HTML]{000000} 96.8}} &
  \multicolumn{1}{c|}{\cellcolor[HTML]{FFFFFF}{\color[HTML]{000000} 99.1}} &
  {\color[HTML]{000000} 100} &
  \multicolumn{1}{c|}{\cellcolor[HTML]{FFFFFF}{\color[HTML]{000000} 79.8}} &
  \multicolumn{1}{c|}{\cellcolor[HTML]{FFFFFF}{\color[HTML]{000000} 99.6}} &
  \multicolumn{1}{c|}{\cellcolor[HTML]{FFFFFF}{\color[HTML]{000000} 98.9}} &
  {\color[HTML]{000000} 100} \\ \cline{2-10} 
\multicolumn{1}{|c|}{\multirow{-4}{*}{\cellcolor[HTML]{FFFFFF}{\color[HTML]{000000} \textbf{\begin{tabular}[c]{@{}c@{}}Online\\     Distillation\end{tabular}}}}} &
  {\color[HTML]{000000} \textbf{OKDDip}} &
  \multicolumn{1}{c|}{\cellcolor[HTML]{FFFFFF}{\color[HTML]{000000} 85.1}} &
  \multicolumn{1}{c|}{\cellcolor[HTML]{FFFFFF}{\color[HTML]{000000} 97.3}} &
  \multicolumn{1}{c|}{\cellcolor[HTML]{FFFFFF}{\color[HTML]{000000} 99.4}} &
  {\color[HTML]{000000} 100} &
  \multicolumn{1}{c|}{\cellcolor[HTML]{FFFFFF}{\color[HTML]{000000} 76.8}} &
  \multicolumn{1}{c|}{\cellcolor[HTML]{FFFFFF}{\color[HTML]{000000} 100}} &
  \multicolumn{1}{c|}{\cellcolor[HTML]{FFFFFF}{\color[HTML]{000000} 99.2}} &
  {\color[HTML]{000000} 100} \\ \hline
\multicolumn{1}{|c|}{\cellcolor[HTML]{FFFFFF}{\color[HTML]{000000} }} &
  {\color[HTML]{000000} \textbf{TF-KD}} &
  \multicolumn{1}{c|}{\cellcolor[HTML]{FFFFFF}{\color[HTML]{000000} 77.5}} &
  \multicolumn{1}{c|}{\cellcolor[HTML]{FFFFFF}{\color[HTML]{000000} 93.8}} &
  \multicolumn{1}{c|}{\cellcolor[HTML]{FFFFFF}{\color[HTML]{000000} 95.4}} &
  {\color[HTML]{000000} 98.7} &
  \multicolumn{1}{c|}{\cellcolor[HTML]{FFFFFF}{\color[HTML]{000000} 78.1}} &
  \multicolumn{1}{c|}{\cellcolor[HTML]{FFFFFF}{\color[HTML]{000000} 98.1}} &
  \multicolumn{1}{c|}{\cellcolor[HTML]{FFFFFF}{\color[HTML]{000000} 99.1}} &
  {\color[HTML]{000000} 100} \\ \cline{2-10} 
\multicolumn{1}{|c|}{\cellcolor[HTML]{FFFFFF}{\color[HTML]{000000} }} &
  {\color[HTML]{000000} \textbf{CS-KD}} &
  \multicolumn{1}{c|}{\cellcolor[HTML]{FFFFFF}{\color[HTML]{000000} 76.5}} &
  \multicolumn{1}{c|}{\cellcolor[HTML]{FFFFFF}{\color[HTML]{000000} 94.1}} &
  \multicolumn{1}{c|}{\cellcolor[HTML]{FFFFFF}{\color[HTML]{000000} 95.9}} &
  {\color[HTML]{000000} 98.0} &
  \multicolumn{1}{c|}{\cellcolor[HTML]{FFFFFF}{\color[HTML]{000000} 81.4}} &
  \multicolumn{1}{c|}{\cellcolor[HTML]{FFFFFF}{\color[HTML]{000000} 98.1}} &
  \multicolumn{1}{c|}{\cellcolor[HTML]{FFFFFF}{\color[HTML]{000000} 96.0}} &
  {\color[HTML]{000000} 99.7} \\ \cline{2-10} 
\multicolumn{1}{|c|}{\cellcolor[HTML]{FFFFFF}{\color[HTML]{000000} }} &
  {\color[HTML]{000000} \textbf{PS-KD}} &
  \multicolumn{1}{c|}{\cellcolor[HTML]{FFFFFF}{\color[HTML]{000000} 82.6}} &
  \multicolumn{1}{c|}{\cellcolor[HTML]{FFFFFF}{\color[HTML]{000000} 98.2}} &
  \multicolumn{1}{c|}{\cellcolor[HTML]{FFFFFF}{\color[HTML]{000000} 98.9}} &
  {\color[HTML]{000000} 100} &
  \multicolumn{1}{c|}{\cellcolor[HTML]{FFFFFF}{\color[HTML]{000000} 75.5}} &
  \multicolumn{1}{c|}{\cellcolor[HTML]{FFFFFF}{\color[HTML]{000000} 99.7}} &
  \multicolumn{1}{c|}{\cellcolor[HTML]{FFFFFF}{\color[HTML]{000000} 98.6}} &
  {\color[HTML]{000000} 100} \\ \cline{2-10} 
\multicolumn{1}{|c|}{\multirow{-4}{*}{\cellcolor[HTML]{FFFFFF}{\color[HTML]{000000} \textbf{\begin{tabular}[c]{@{}c@{}}Self \\     Distillation\end{tabular}}}}} &
  {\color[HTML]{000000} \textbf{DDGSD}} &
  \multicolumn{1}{c|}{\cellcolor[HTML]{FFFFFF}{\color[HTML]{000000} 76.9}} &
  \multicolumn{1}{c|}{\cellcolor[HTML]{FFFFFF}{\color[HTML]{000000} 96.6}} &
  \multicolumn{1}{c|}{\cellcolor[HTML]{FFFFFF}{\color[HTML]{000000} 98.2}} &
  {\color[HTML]{000000} 100} &
  \multicolumn{1}{c|}{\cellcolor[HTML]{FFFFFF}{\color[HTML]{000000} 83.4}} &
  \multicolumn{1}{c|}{\cellcolor[HTML]{FFFFFF}{\color[HTML]{000000} 99.8}} &
  \multicolumn{1}{c|}{\cellcolor[HTML]{FFFFFF}{\color[HTML]{000000} 97.2}} &
  {\color[HTML]{000000} 100} \\ \hline
\end{tabular}
\caption{Top-1 and Top-5 accuracy of student model using knowledge distillation methods (98\% reduction)}
\label{tab:kd98}
\end{table*}

\subsection{Pruning}

We evaluated network pruning based on three pruning ratios: 90\%, 95\%, and 98\%. Table \ref{tab:layers} shows the number of parameters in each layer and memory consumption after pruning. It can be seen that by using network pruning, the memory size can be reduced by up to 15x. We can even build a smaller CNN2D network than MLP and CNN1D, with a pruning ratio of 98\%. The number of filters in Conv layers and neurons in FC layers for the pruned network is shown in Table \ref{tab:filters}. We keep the same student network architectures as the pruned model weights to compare the knowledge distillation methods with network pruning.

The top-1 and top-5 accuracy using different pruning ratios is shown in Table \ref{tab:prune90}, Table \ref{tab:prune95}, and Table \ref{tab:prune98}, respectively. We also conducted experiments training the same pruned architecture from random initialization (Scratch) and compared its performance with the pruning methods. It can be seen the network pruning results outperform the scratch training baseline, which indicates that the pruning methods can maintain the model capacity after removing less important filters and neurons. The model using a pruning ratio of 98\% performs better than MLP and CNN-1D, even though it consumes less memory than MLP and CNN-1D, as shown in Table \ref{tab:layers}. 

Additionally, different fine-tuning strategies (discussed in the method section) have been shown in Table \ref{tab:strategy}. It can be seen that if we prune the model layer by layer and retrain it iteratively, the model performance shows the best. However, this strategy is time-consuming when a deep model with a large number of layers is fine-tuned. For simplicity, we use the classic fine-tuning strategy I for all our pruning algorithms.

\subsection{Quantization}

We implemented the dynamic quantization, static quantization, and QAT quantization methods. The quantization result is shown in Table \ref{tab:qresults}. Inference latency represents how fast the model is during the inference time. It is demonstrated that quantized models consume less memory with lower inference latency because of less expansive computations without significant loss in model performance. QAT achieves better accuracy than static quantization by updating the quantization parameters during training. The memory can be reduced up to 4x, and the inference latency can also be optimized up to 4x. Quantization techniques can be easily combined with other network compression methods to reduce the computation and memory costs further.

\subsection{Knowledge distillation}

The results using various knowledge distillation methods given different pruning ratios are shown in Table \ref{tab:kd90}, \ref{tab:kd95}, and \ref{tab:kd98}, respectively. We also compare their performance with baseline performance and train from scratch with the same student network architecture. It can be observed that offline distillation methods successfully reduce the number of parameters while maintaining or even improving the model performance. It is shown that most online and self-KD methods outperform the baseline and scratch accuracy. Specifically, two more recent works, OKDDip and DDGSD show the best overall performance. Compared with the three offline KD tables, it can be seen that offline distillation approaches achieve an overall better performance than online and self-distillation. However, online and self-distillation methods consume less memory since no strong teacher network is required. 

\section{Discussions and Conclusions}

In this work, we implement and evaluate three network compression techniques in land cover classification: network pruning, quantization, and knowledge distillation. Four pruning, three quantization, and fourteen knowledge distillation methods have been implemented and tested. The results show that network compression successfully reduces the model size and memory without significant classification loss in the application of land cover classification. Specifically, knowledge distillation methods perform better than benchmark pruning methods. Quantization successfully reduces the bit width of network weights, which makes it both memory and computationally efficient. Offline KD methods perform better than online and self-distillation methods, given that they often require a large, well-trained teacher network and consume more memory. Future work may combine different network compression techniques to achieve better model performance. 

We are aware of several limitations in this work: 1) Not all network compression algorithms are covered since it is a broad topic and requires a more comprehensive study. For example, low-rank approximation (Principle Component Analysis, Singular Value Decomposition, etc.)is a technique used in linear algebra and numerical analysis to approximate a given matrix by another matrix with a lower rank. By finding a low-rank approximation of a given matrix, we can reduce the storage and computational requirements for manipulating the matrix and extract meaningful information about the underlying data. 2) We only focus on two land cover classification datasets to evaluate all algorithms. Tests on more datasets can further justify our observations. 3) We focus on simple CNN architectures to test the algorithms. Future work can be deeper networks such as VGG, ResNet, etc. 

We have not implemented the listed network compression techniques specifically in remote sensing because of two primary reasons: 1) the datasets used in those works are not the same as ours since different works focus on a specific remote sensing application, such as object detection, target recognition, etc.; 2) most of the work did not provide a published GitHub link for their implementation, and it might be challenging to reproduce their results.   

{
	\begin{spacing}{1.17}
		\normalsize
		\bibliography{network_compression} % Include your own bibliography (*.bib), style is given in isprs.cls
	\end{spacing}
}

% \section*{APPENDIX (Optional)}\label{APPENDIX}

% Any additional supporting data may be appended, provided the paper does not exceed the limits given above. 

\end{document}